\documentclass{article}

\usepackage{iclr2027_conference,times}
\usepackage{xcolor}
\usepackage[T1]{fontenc}

\usepackage{amsmath,amsfonts,bm}

\def\eqref#1{equation~\ref{#1}}

\def\1{\bm{1}}

\DeclareMathAlphabet{\mathsfit}{\encodingdefault}{\sfdefault}{m}{sl}
\SetMathAlphabet{\mathsfit}{bold}{\encodingdefault}{\sfdefault}{bx}{n}

\usepackage{booktabs}
\usepackage{CJKutf8}
\usepackage{graphicx}
\usepackage{subcaption}

\definecolor{APTint}{HTML}{E3ECF6}
\usepackage{soul}
\soulregister\cite7
\soulregister\citet7
\soulregister\citep7
\soulregister\ref7
\newcommand{\aphighlight}[1]{#1}
\usepackage{hyperref}
\usepackage{microtype}
\usepackage{placeins}
\usepackage{url}

\usepackage{tabularx}
\usepackage{array}

\usepackage[most]{tcolorbox}
\usepackage{xcolor}
\usepackage{fontawesome5}

\definecolor{rcblue}{HTML}{5B8DB8}
\definecolor{rclight}{HTML}{F7FAFD}
\definecolor{rcframe}{HTML}{B8CEDF}

\newtcolorbox{keyinsight}[1][Key Insight]{
  enhanced,
  breakable,
  colback=rclight,
  colframe=rcframe,
  boxrule=0.65pt,
  arc=2.8mm,
  left=3mm,
  right=3mm,
  top=4.5mm,
  bottom=2.8mm,
  drop fuzzy shadow=black!10,
  title={\faLightbulb\hspace{0.6em}\textbf{#1}},
  coltitle=white,
  fonttitle=\small\bfseries,
  attach boxed title to top left={
    xshift=3mm,
    yshift=-4mm
  },
  boxed title style={
    colback=rcblue,
    colframe=rcblue,
    boxrule=0pt,
    arc=2mm,
    left=2.2mm,
    right=2.2mm,
    top=0.8mm,
    bottom=0.8mm
  },
  before upper=\small
}

\usepackage{xspace}

\usepackage{xcolor}
\definecolor{cvprblue}{rgb}{0.21,0.49,0.74}
\hypersetup{
  colorlinks=true,
  linkcolor=cvprblue,
  citecolor=cvprblue,
  urlcolor=cvprblue,
  filecolor=cvprblue,
  linktoc=all,
  pdftitle={ICA Lens: Non-Gaussianity Reveals Interpretable Directions in Language Models},
  pdfauthor={Sida Liu, Feijiang Han, Sarah Wiegreffe, and Furong Huang},
  pdfsubject={Preprint}
}

\definecolor{rebuttal}{HTML}{2B6CB0}

\definecolor{PythonBlue}{HTML}{306998}
\definecolor{PythonGold}{HTML}{B48B20}
\definecolor{PythonLight}{HTML}{EDF4FA}
\definecolor{MetaGray}{HTML}{667085}

\title{ICA Lens: Non-Gaussianity Reveals Interpretable Directions in Language Models}

\author{%
  \begin{tabular*}{\textwidth}{@{\extracolsep{\fill}}cc@{}}
    \\[0em]
    \begin{tabular}[t]{c}
      \textbf{Sida Liu}\textsuperscript{*}\\
      Independent Researcher
    \end{tabular}
    &
    \begin{tabular}[t]{c}
      \textbf{Feijiang Han}\textsuperscript{*}\\
      University of Maryland, College Park
    \end{tabular}
    \\[2em]
    \begin{tabular}[t]{c}
      \textbf{Sarah Wiegreffe}\\
      University of Maryland, College Park
    \end{tabular}
    &
    \begin{tabular}[t]{c}
      \textbf{Furong Huang}\\
      University of Maryland, College Park
    \end{tabular}
  \end{tabular*}\\[0.9em]
  {\footnotesize
    \textcolor{MetaGray}{\textsuperscript{*}Equal contribution.}\\[0.4em]
    \href{mailto:me@liusida.com}{\textcolor{MetaGray}{\nolinkurl{me@liusida.com}}}
    \quad\textcolor{MetaGray}{\textbullet}\quad
    \href{mailto:feijhan@umd.edu}{\textcolor{MetaGray}{\nolinkurl{feijhan@umd.edu}}}\\[0.8em]
    \textcolor{PythonBlue}{\faIcon{python}}
    \hspace{0.3em}
    {\sffamily\bfseries\textcolor{PythonBlue}{Python Package}}
    \hspace{0.6em}
    \href{https://pypi.org/project/icalens/}{%
      {\ttfamily\bfseries\textcolor{PythonBlue}{pip install icalens}}%
    }
  }%
}

\iclrpreprintcopy

\begin{document}

\suppressfloats[t]
\maketitle

\addtocontents{toc}{\protect\setcounter{tocdepth}{-1}}

\begin{abstract}
  Finding human-interpretable directions in language-model activations is important for understanding and controlling model behavior.
  These directions are usually discovered through indirect learning objectives such as sparse reconstruction, but it remains unclear what existing statistical structure distinguishes interpretable directions and whether that structure can be used directly for feature discovery.
  We find that non-Gaussianity provides such a signature: projection distributions of natural-text residual-stream activations along sparse autoencoder (SAE) decoder directions are substantially more non-Gaussian than those along random directions, with most contexts forming a concentrated background and recurring patterns often occupying one tail.
  This raises a natural question: \textit{if interpretable directions tend to have non-Gaussian projection distributions, can we discover them by searching for non-Gaussian directions directly in activation space, without training another sparse dictionary?}
  To test this hypothesis, we introduce \textsc{ICA Lens}\footnote{Project page: \url{https://liusida.github.io/ica-lens-paper/}.}, a practical workflow for \textbf{efficient} discovery, \textbf{reliable} annotation, and \textbf{faithful} evaluation of non-Gaussian directions in LLM residual-stream activations.
  Across GPT-2 Small, Gemma~2~2B, and Qwen~3.5~9B Base, we find that (1) the recovered non-Gaussian directions capture coherent lexical, syntactic, and semantic patterns under both human and LLM-based evaluation; (2) non-Gaussianity is positively associated with both human-label confidence and automated interpretability; and (3) the resulting coordinates are competitive with SAEs across interpretability, sparse reconstruction, concept prediction, and causal interventions, while fitting in minutes per layer and using 7--43$\times$ less memory.
  These results show that non-Gaussianity is not merely a statistical property of learned features, but a direct unsupervised signal for discovering interpretable and useful structure in activations.

\end{abstract}
\section{Introduction}

Understanding the internal representations of large language models (LLMs) is a central goal of mechanistic interpretability~\citep{olah2020zoom,elhage2021mathematical,gurnee2023finding,park2023linear,han2025zerotuning,liu2024unraveling,han2026read}.
Sparse autoencoders (SAEs) have become a dominant approach to this problem by decomposing activations into sparse latents over an overcomplete dictionary of learned directions~\citep{huben2024sparse,bricken2023monosemanticity,templeton2024scaling,gao2025scaling}.
Their success raises a more fundamental question: what makes these directions special in the underlying activation space?
Do they share a measurable statistical property that distinguishes them from arbitrary directions?
If so, we may be able to discover interpretable directions by searching for this property directly, rather than relying solely on dictionary learning.

We start from a simple observation: interpretable directions are often selective. A direction associated with a particular lexical, syntactic, semantic, or discourse pattern tends to respond strongly when that pattern appears, while remaining relatively inactive in most other contexts.
Across natural-text activations, this selectivity can produce a concentrated background with a distinct tail, resulting in a non-Gaussian projection distribution.

\begin{figure}[ht!]
  \centering
  \includegraphics[width=\textwidth]{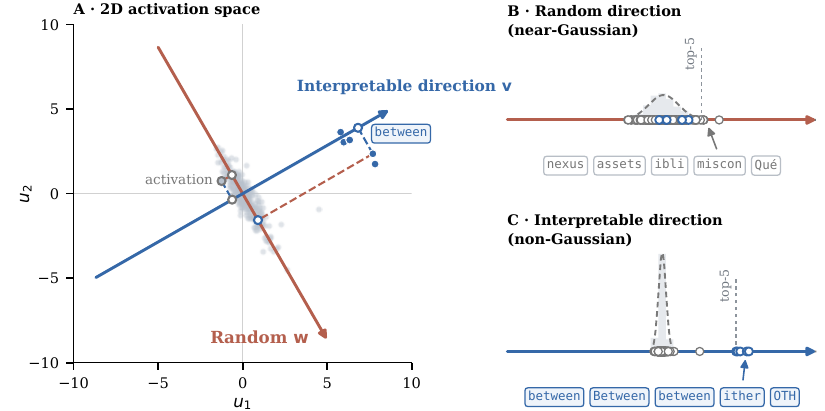}
  \caption{\textbf{Interpretable features leave a non-Gaussian signature.}
    (A) Real GPT-2 token activations projected onto a random direction $\mathbf{w}$ and an interpretable direction $\mathbf{v}$.
    (B) The random projection is close to Gaussian and its top-scoring tokens are semantically mixed.
  (C) The interpretable direction isolates a coherent set of related tokens in a non-Gaussian tail.}
  \label{fig:overview}
\end{figure}

We find this pattern in public SAE dictionaries (Section~\ref{sec:ica-discovery}), in directions constructed directly from coherent concepts (Appendix~\ref{app:concept-constructed-directions}), and in native MLP neurons (Appendix~\ref{app:mlp-nongaussianity}).
For example, projections onto SAE decoder directions are substantially more non-Gaussian than random projections, even though SAEs are trained for sparse reconstruction rather than explicit non-Gaussianity. Figure~\ref{fig:overview} illustrates this pattern. These observations motivate our central hypothesis: \aphighlight{\textbf{\emph{non-Gaussianity itself can serve as an unsupervised signal for discovering interpretable directions in language-model activations}.}}

\vspace{-0.5em}

The hypothesis also suggests a practical opportunity.
SAEs are powerful when a large, fine-grained feature dictionary is needed, but training a new dictionary for every model, layer, and activation site can be expensive~\citep{lieberum2024gemma}.
For early-stage analysis, however, a compact set of interpretable coordinates may already suffice to identify promising layers, inspect model states, probe concepts, and test lightweight interventions.
Direct feature discovery from activation statistics could therefore provide a useful first lens before larger-scale dictionary learning.

We test this idea with \textsc{ICA Lens}, a workflow based on independent component analysis (ICA) for discovering, annotating, evaluating, and using non-Gaussian directions in LLM residual-stream activations.
We develop an optimized GPU-parallel implementation of symmetric FastICA~\citep{hyvarinen2000independent}, with improved monitoring of fitting dynamics and more reliable convergence criteria for efficient feature discovery. We then interpret the recovered directions using top-activating contexts, vocabulary readouts, and controlled counterfactuals. Importantly, we find that coherent semantic patterns often concentrate in one tail of an ICA component's activation distribution. Because ICA assigns component signs arbitrarily, we use skewness to identify this tail before evaluation. This simple correction addresses a long-overlooked sign-sensitive bias that substantially understated ICA interpretability in the earliest direct ICA--SAE comparison~\citep{huben2024sparse}.

We investigate two questions: whether the recovered directions are interpretable, and whether they are useful.
Across GPT-2 Small, Gemma~2~2B, and Qwen~3.5~9B, our human audit shows that these directions capture coherent lexical, syntactic, semantic, positional, and contextual patterns (Section~\ref{sec:ica-interpretability}).
At scale, automated evaluation shows that skewness-selected ICA directions achieve interpretability comparable to public SAE features.
We further find that mean automated interpretability increases as FastICA recovers more non-Gaussian directions. Across components, greater non-Gaussianity is also associated with higher human-label confidence and automated interpretability scores.

The recovered coordinates are also useful beyond feature labeling (Section~\ref{sec:ica-evaluation}).
Their component mixtures can be used to inspect new activations and support generation-time steering.
Across held-out downstream datasets, they are competitive with public SAEs in sparse reconstruction and sparse concept probing, achieve higher targeted probe perturbation scores under small-to-medium feature budgets, and show better causal control on RAVEL.
These capabilities come without training a large sparse dictionary.
With one million activations per layer, our GPU implementation fits an ICA basis in under one minute for GPT-2 Small, about 5.5 minutes for Gemma~2~2B, and 12--13 minutes for Qwen~3.5~9B Base, while using 7--43$\times$ less parameter memory per layer than the matched SAEs.

Overall, these results suggest a different view of feature discovery in language models.
\aphighlight{\textbf{Interpretable directions have an intrinsic higher-order statistical signature. This signature can itself be optimized to recover interpretable, useful coordinates.}}
This reframes non-Gaussianity from a descriptive property of learned features into a direct signal for discovering them.
\section{From Non-Gaussianity to Feature Discovery}
\label{sec:icalens}

\vspace{-1em}
\begin{figure}[ht!]
  \centering
  \includegraphics[width=\textwidth]{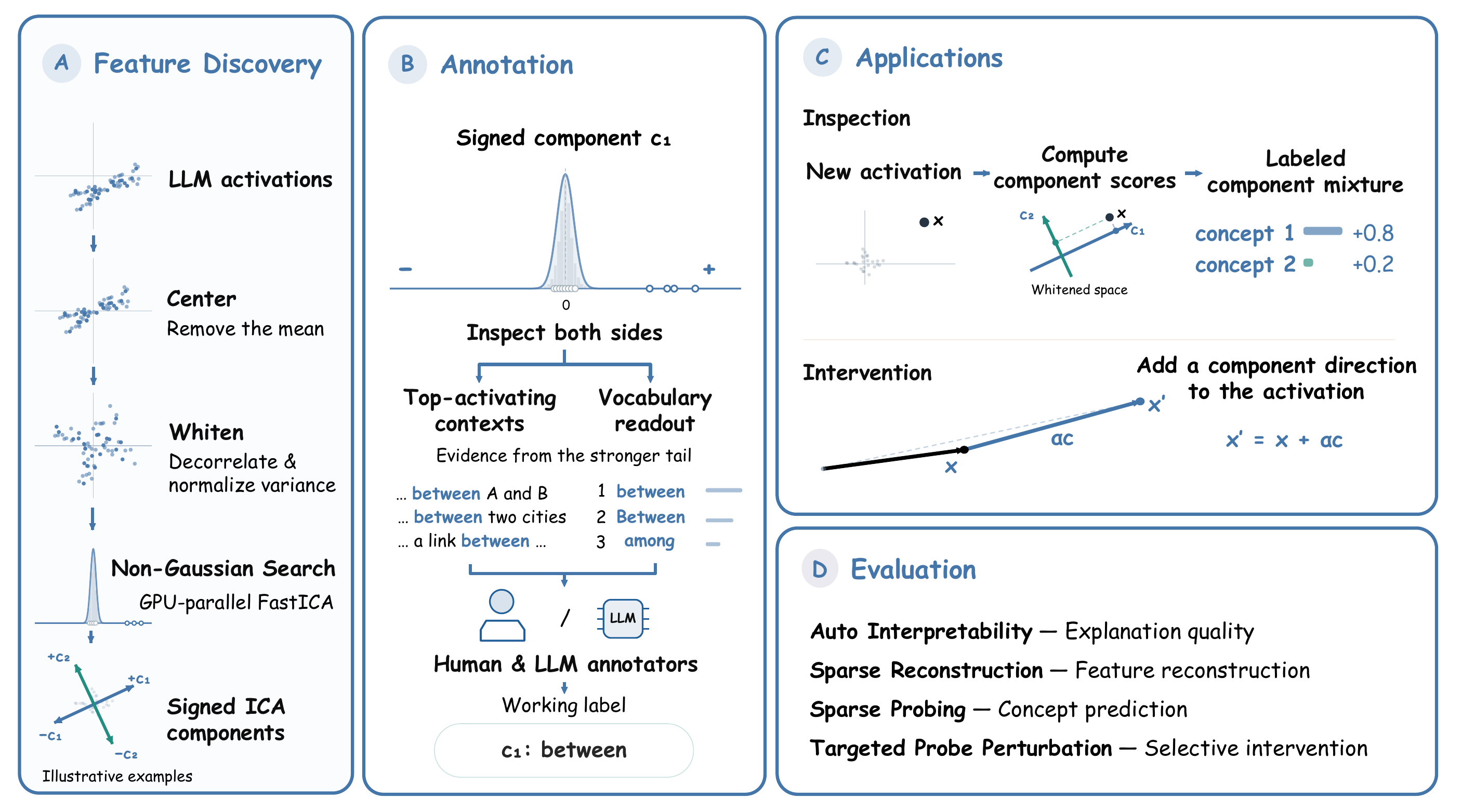}
  \caption{\textbf{ICA Lens workflow.}
    (A) GPU-accelerated symmetric FastICA discovers non-Gaussian directions
    from centered, whitened activations.
    (B) Top-activating contexts and vocabulary readouts support annotation.
    (C) Coordinates enable state inspection and editing.
  (D) Evaluation covers interpretability, sparse reconstruction, sparse probing, and targeted probe perturbation.}
  \label{fig:icalens-workflow}
\end{figure}

We investigate whether non-Gaussianity can guide feature discovery using ICA Lens, an end-to-end workflow illustrated in Figure~\ref{fig:icalens-workflow}. 
We first recover non-Gaussian directions directly from residual-stream activations (Section~\ref{sec:ica-discovery}), then examine their interpretability and its relationship to non-Gaussianity (Section~\ref{sec:ica-interpretability}), and finally evaluate their utility for reconstruction, concept readout, and targeted intervention (Section~\ref{sec:ica-evaluation}).

\subsection{Non-Gaussian Structure Can Be Recovered Directly}
\label{sec:ica-discovery}

We begin by testing whether the non-Gaussian structure observed in SAE decoder directions can be recovered directly from residual-stream activations. After centering and whitening the activations, we use FastICA to search for non-Gaussian directions without learning a sparse dictionary.

Given cached activations $X^\ell\in\mathbb{R}^{n\times d}$ from layer $\ell$, we subtract the empirical mean to obtain $X_c$. Assuming a full-rank empirical covariance, we write its eigendecomposition as $\frac{1}{n}X_c^\top X_c=U\Lambda U^\top$ and whiten the activations using $Z=X_cU\Lambda^{-1/2}$, so that $Z$ has zero empirical mean and identity empirical covariance. We then fit a full-dimensional basis with symmetric FastICA~\citep{hyvarinen1999fast}, which updates all components in parallel and jointly orthogonalizes them after each update.\footnote{The resulting axes are orthogonal in whitened space. Their corresponding reading and writing directions in the original residual-stream space need not be orthogonal.} The resulting orthogonal matrix $W=[w_1,\ldots,w_d]^\top\in\mathbb{R}^{d\times d}$ defines component scores $Y=ZW^\top$, with $y_j=Zw_j\in\mathbb{R}^n$ collecting the scores of component $j$ across tokens. We use the logcosh contrast as a surrogate objective for non-Gaussianity. Here, \emph{non-Gaussianity} refers to departures from a standard Gaussian distribution in each component's scores across natural-text token activations. Appendix~\ref{app:ica-fitting} gives the full activation-collection and FastICA pipeline.

\begin{figure}[ht!]
  \centering
  \includegraphics[width=\textwidth]{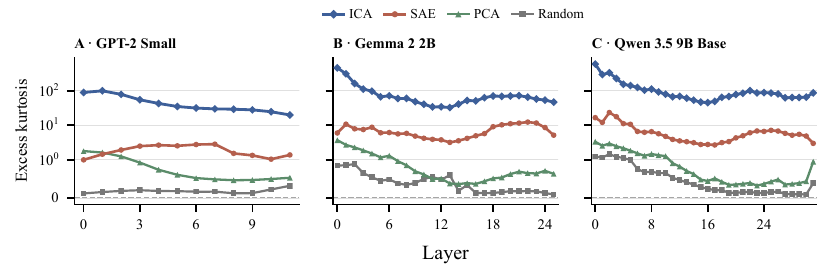}
  \caption{\textbf{ICA recovers the most heavy-tailed directions across models and layers.}
  Mean excess kurtosis of projections onto unit-norm ICA, SAE, PCA, and random directions, measured on one million residual-stream activations per layer. Random projections remain near the Gaussian reference, while ICA consistently produces the largest excess kurtosis.}
  \label{fig:direction-excess-kurtosis}
\end{figure}

\aphighlight{\textbf{LLM residual streams contain strong, recoverable non-Gaussian structure.}}
To compare this structure across ICA, SAE, PCA, and random directions, we measure the excess kurtosis of their projection scores as follows\footnote{Logcosh contrast and excess kurtosis follow closely aligned optimization trajectories and show similar convergence behavior (Appendix~\ref{app:ica-convergence}). We use excess kurtosis in the following analyses because it is easier for readers to understand, and its wider numerical range makes differences across directions easier to interpret.}. 
For each method, let $y_{ij}$ denote the projection score of activation $i$ along direction $j$. We compute
\begin{equation}
  \kappa_j
  =
  \frac{
    \frac{1}{n}\sum_{i=1}^{n}(y_{ij}-\bar y_j)^4
  }{
    \left[
      \frac{1}{n}\sum_{i=1}^{n}(y_{ij}-\bar y_j)^2
    \right]^2
  }
  -3,
  \qquad
  \bar y_j=\frac{1}{n}\sum_{i=1}^{n}y_{ij}.
\end{equation}
A Gaussian distribution has population excess kurtosis zero, while positive values indicate heavier tails in the fourth-moment sense.
Figure~\ref{fig:direction-excess-kurtosis} reveals a consistent pattern across models and layers.
Random projections remain close to the Gaussian reference, while PCA stays much closer to the random baseline and captures only weak heavy-tailed structure.
In contrast, SAE decoder directions are substantially more heavy-tailed despite being trained for sparse reconstruction rather than non-Gaussianity.
ICA, which searches for non-Gaussianity directly, recovers the strongest heavy-tailed projections across all three models.
These results show that modern LLM residual streams contain strong higher-order structure that is directly recoverable.
Appendix~\ref{app:concept-constructed-directions} constructs interpretable non-Gaussian directions from coherent concepts and compares them with ICA directions. Appendix~\ref{app:mlp-nongaussianity} extends the analysis to native MLP neurons, finding a positive association between excess kurtosis and lexical concentration. Appendix~\ref{app:ica-sae-relationship} further compares ICA and SAE directions through directional overlap and tokenwise responses.

\aphighlight{\textbf{This structure can be recovered efficiently at LLM scale.}}
Applying ICA to modern LLMs requires fitting thousands of component directions to a million token activations per layer, often across many layers. Prior work has largely relied on the scikit-learn implementation of FastICA.\footnote{\url{https://scikit-learn.org/stable/modules/generated/sklearn.decomposition.FastICA.html}} At this scale, CPU fitting becomes costly, while the default convergence criterion can demand continued optimization after most of the non-Gaussian structure has already been recovered. 

We first address the computational bottleneck with a PyTorch implementation of symmetric FastICA, running both component updates and joint orthogonalization on GPUs.
We also reconsider how convergence should be assessed for feature discovery. Scikit-learn uses maximum LIM, $\max_j \bigl(1-\lvert (w_j^{(t)})^\top w_j^{(t-1)}\rvert\bigr)$, where $w_j^{(t)}$ is the unit direction of component $j$ at iteration $t$. This statistic measures the largest directional change between successive iterations, up to sign. A few slowly stabilizing components can therefore keep maximum LIM high. More fundamentally, LIM measures whether directions have stopped moving, not whether the recovered non-Gaussian structure has saturated. We therefore monitor logcosh contrast and excess kurtosis across the component population alongside directional stability. Population non-Gaussianity largely plateaus by roughly 50 iterations, while mean and maximum LIM continue to evolve (Appendix~\ref{app:ica-convergence}). We use 50 iterations as the default operating point based on this statistical saturation. Extending to 200 iterations further improves cross-seed component matching when reproducible component identities are the priority (Appendix~\ref{app:ica-stability}).

Our sample-size ablation shows that FastICA can overfit small fitting sets, yielding high excess kurtosis on fitting tokens and much lower values on held-out documents (Appendix~\ref{app:ica-fitting-sample-size}). We find that 1M fitting activations are sufficient for three models we tested.
At this fitting budget, per-layer fitting on a single NVIDIA GB10 ($\approx 0.08\times$H100, $0.13\times$A100) takes under one minute for GPT-2 Small, $\approx$5.5 minutes for Gemma~2~2B, and 12--13 minutes for Qwen~3.5~9B Base.
The resulting bases use 7--43$\times$ less parameter memory per layer than the matched public SAEs. Appendix~\ref{app:ica-efficiency} gives the full memory, latency, and end-to-end fitting comparison, while Appendix~\ref{app:sae-baselines} lists the exact matched public SAE checkpoints. 

\subsection{Many Non-Gaussian Directions Are Interpretable}
\label{sec:ica-interpretability}

Non-Gaussianity alone does not guarantee interpretability. We therefore examine the recovered directions through human annotation and automated interpretability evaluation, and investigate how interpretability varies with non-Gaussianity.

\subsubsection{From Components to Working Interpretations}

A few coherent top examples are not sufficient to establish interpretability, since a label can be fit post hoc to a small set of extreme activations.
We therefore treat component labels as hypotheses that must be supported by multiple sources of evidence and survive controlled counterexamples.

\begin{table}[ht!]
  \centering
  \caption{\textbf{Representative ICA components.}
  Top-activating contexts and vocabulary readouts support the same working label. Underlining marks the scored token.}
  \label{tab:component-profile-examples}
  \scriptsize
  \begin{CJK*}{UTF8}{gbsn}
    {\setlength{\tabcolsep}{2pt}
      \begin{tabular}{@{}p{0.10\textwidth}p{0.16\textwidth}p{0.37\textwidth}p{0.15\textwidth}p{0.13\textwidth}@{}}
        \toprule
        Component & Working label & Top-activating contexts & Logit Lens & R-Lens \\
        \midrule
        \raggedright GPT-2\newline L8/C396
        &
        \raggedright Programming/API specification language
        &
        \raggedright
        \#1 in Java interface that \underline{returns} class or subclass type\newline
        \#2 require all modules ending \underline{in} ``.spec'' from\newline
        \#3 ``categories'' is not \underline{an} array, but object
        &
        \raggedright
        \#1 ``JSON''\newline
        \#2 ``syntax''\newline
        \#3 ``initialization''
        &
        \raggedright
        \#1 ``JSON''\newline
        \#2 ``syntax''\newline
        \#3 ``tuple''
        \tabularnewline

        \midrule

        \raggedright Gemma~2\newline 2B L20/C1154
        &
        \raggedright 1960s--1970s American culture/history
        &
        \raggedright
        \#1 76 by then \underline{President} Gerald Ford, and\newline
        \#2 at the height of \underline{The} Summer of Love.\newline
        \#3 started for the Yankees \underline{against} the Reds, whose
        &
        \raggedright
        \#1 ``Vietnam''\newline
        \#2 ``hippie''\newline
        \#3 ``seventies''
        &
        \raggedright
        \#1 ``Vietnam''\newline
        \#2 ``hippie''\newline
        \#3 ``Vietnam''
        \tabularnewline

        \midrule

        \raggedright Qwen~3.5\newline 9B Base L20/C352
        &
        \raggedright Both
        &
        \raggedright
        \#1 3 weeks. \underline{Both} the rats' behavior\newline
        \#2 creative nonfiction. \underline{Both} graduate and or flagship\newline
        \#3 copies, and that \underline{both} that copyright notice
        &
        \raggedright
        \#1 ``sides''\newline
        \#2 ``sexes''\newline
        \#3 ``/all''
        &
        \raggedright
        \#1 ``sides''\newline
        \#2 ``sexes''\newline
        \#3 ``这两种''
        \tabularnewline

        \bottomrule
    \end{tabular}}
  \end{CJK*}
\end{table}

\textbf{Component profiles combine contextual and vocabulary evidence.}
We use two sources of evidence to annotate each component.
(1) Activation contexts. For component $j$, FastICA returns an axis $w_j$ in whitened activation space, with
score $y_{ij}=w_j^\top z_i$ for token activation $z_i$. The highest-scoring contexts show \emph{when} a component is active. (2) Vocabulary readouts. Mapping the component back to the residual stream gives a writing direction $d_j$, which we decode with Logit Lens~\citep{nostalgebraist2020logitlens} and
R-Lens~\citep{blank2026rlens} to read \emph{what} it tends to write. Table~\ref{tab:component-profile-examples} shows some examples.

\textbf{Iterative counterfactual tests refine working labels.}
A human or LLM annotator first summarizes the evidence in a working label, then tests the label on controlled prompt pairs that differ in the proposed concept
while preserving other aspects of the input. The annotator revises the label
when the tests reveal a mismatch and repeats this process until they can assign
a label with confidence. For example, for a component labeled ``children's
educational content,'' we compare a picture \emph{book} for preschoolers with a
reference \emph{book} for engineers. If the label is informative, the component
should respond much more strongly to the first prompt despite the shared target
word.
Appendix~\ref{app:manual-audit} gives the complete annotation workflow, sentence-level traces, controlled counterfactual checks, confidence rubric, and all 150 randomly sampled component annotations.

\aphighlight{\textbf{ICA recovers many semantically interpretable directions.}}
We randomly sample 150 components from the 50-iteration fits, with 50 from each model. Two interpretability experts independently annotate each component and assign confidence scores using the same 0--5 rubric. With scores of 4 or 5 counted as high confidence, the mean high-confidence rate across annotators is 85.7\% overall, with rates of 87\% for GPT-2 Small, 85\% for Gemma~2~2B, and 85\% for Qwen~3.5~9B Base. Both annotators rate 123 of the 150 components as high confidence. The labels cover lexical, syntactic, semantic, positional, and contextual patterns. Appendix~\ref{app:annotator-comparison} reports agreement and illustrates differences in labels and confidence ratings.

\aphighlight{\textbf{ICA semantics are often one-sided.}}
Although every ICA component has two signed tails, the human audit reveals that a coherent semantic pattern is often concentrated in one tail, while the opposite tail is diffuse rather than an opposing concept.
Consistent with this observation, strongly non-Gaussian components also tend to have more asymmetric score distributions. Across the 150 audited components, excess kurtosis is strongly correlated with absolute skewness ($\rho_s=0.92$).
This suggests that much of the heavy-tailed structure recovered by ICA is asymmetric rather than equally distributed across both tails.
We use skewness computed on the fitting activations to identify the tail that drives the distributional asymmetry, and use that tail for subsequent analysis and evaluation.
A positive skew selects the positive tail, while a negative skew selects the negative tail.
Appendix~\ref{app:tail-selection} details the skewness-based orientation and tie-breaking rules.

\subsubsection{Automated Interpretability Evaluation}

\begin{figure}[ht!]
  \centering
  \includegraphics[width=\textwidth]{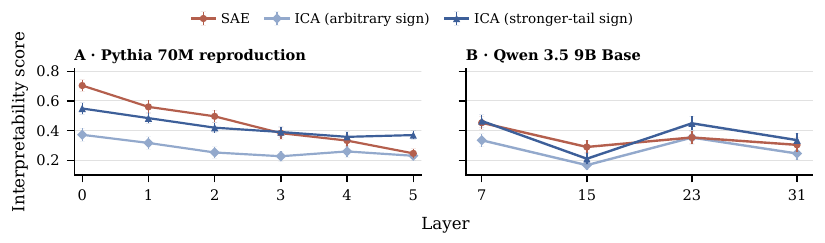}
  \caption{\textbf{ICA was underestimated by a sign-sensitive evaluation.}
    (A) On Pythia 70M, arbitrary-sign evaluation reproduces the reported gap to
    SAE; stronger-tail orientation recovers much of it.
    (B) On Qwen~3.5~9B Base, sign-corrected ICA is competitive with SAE.
  Points show means over 150 components.}
  \label{fig:historical-baseline}
\end{figure}

\aphighlight{\textbf{LLMs annotate components and score interpretability at scale.}}
Following the method and evaluation protocol of \citet{huben2024sparse}, we use an explanation--prediction pipeline to evaluate interpretability at scale.
An LLM annotator receives text fragments with their component scores and writes a short explanation of the activation pattern.
A separate LLM predictor then uses this explanation to predict token-level component scores on held-out fragments.
The automated interpretability score is the Pearson correlation between predicted and observed scores, so a high value indicates that the explanation generalizes to unseen examples.

\aphighlight{\textbf{A long-overlooked sign ambiguity systematically biases ICA evaluation downward.}}
The ICA--SAE comparison in \citet{huben2024sparse} evaluates only the positive tail of each ICA component. Because FastICA assigns signs arbitrarily, this protocol can overlook the tail containing coherent semantics and underestimate ICA interpretability.

Figure~\ref{fig:historical-baseline} isolates this effect. We first
reproduce the original Pythia 70M comparison using the released setup and
implementation,\footnote{We follow the released ICA baseline at
\url{https://github.com/HoagyC/sparse_coding/blob/main/autoencoders/}.}
recovering the previously reported gap between SAE and ICA
(\emph{ICA, arbitrary sign}). We then evaluate each component using the tail
selected by the procedure described above (\emph{ICA, stronger-tail sign}). With this correction, much of the gap disappears and ICA becomes competitive
with SAE. The same sign-corrected evaluation on Qwen~3.5~9B Base also finds ICA competitive with the public SAE baseline.
Appendix~\ref{app:auto-interpretability} specifies the explanation--prediction pipeline and matched feature-sampling rules, while Appendix~\ref{app:ica-sae-robustness} tests the comparison across ICA fitting budgets, training corpora, and evaluator models.

\subsubsection{Revisiting the Link Between Interpretability and Non-Gaussianity}
\label{sec:interpretability-non-gaussianity}

\begin{figure}[ht!]
  \centering

  \begin{subfigure}[t]{0.58\textwidth}
    \centering
    \includegraphics[width=\linewidth]{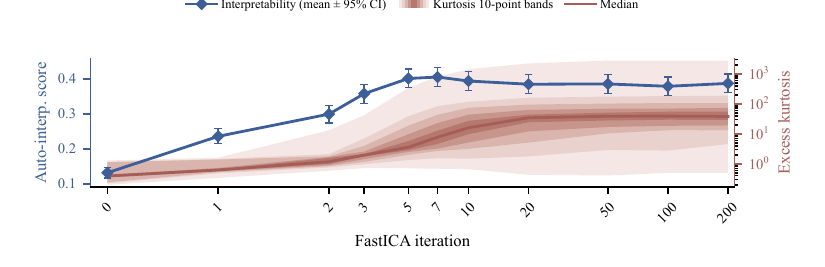}
    \caption{Across FastICA fitting iterations.}
    \label{fig:nonGaussian-interp-dynamics}
  \end{subfigure}
  \hfill
  \begin{subfigure}[t]{0.39\textwidth}
    \centering
    \includegraphics[width=\linewidth]{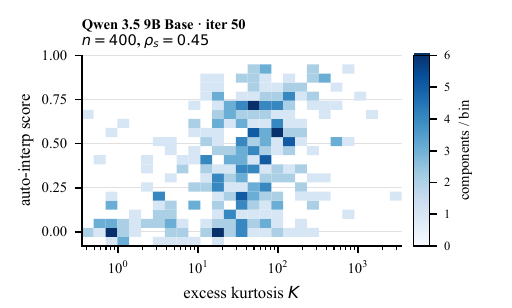}
    \caption{Across components at iteration 50.}
    \label{fig:nonGaussian-interp-components}
  \end{subfigure}

  \caption{\textbf{Interpretability rises with non-Gaussianity.}
    (A) Mean automated interpretability rises during FastICA fitting and then saturates as non-Gaussianity increases.
  (B) At iteration~50, excess kurtosis remains positively associated with automated interpretability across individual components.}
  \label{fig:nonGaussian-interp}
\end{figure}

We next examine whether stronger non-Gaussianity is associated with greater interpretability. We track 400 Qwen 3.5 9B Base components (100 each from layers 7, 15, 23, and 31) across FastICA checkpoints, recording their excess kurtosis and sign-corrected automated interpretability scores. Figure~\ref{fig:nonGaussian-interp} shows the relationship at two levels.
Across fitting iterations, mean automated interpretability rises as FastICA finds increasingly non-Gaussian directions and then saturates.
Across components at iteration~50, excess kurtosis also remains positively associated with automated interpretability, with a Spearman correlation of $\rho_s=0.45$. 
Thus, both the fitting dynamics and the component-level comparison show that stronger non-Gaussianity is associated with greater interpretability.
Appendix~\ref{app:interpretability-error-analysis} examines cases that deviate from this trend and identifies two failure modes, showing that some outliers reflect limitations of automated evaluation.

\begin{figure}[ht!]
  \centering
  \begin{subfigure}[t]{0.49\textwidth}
    \centering
    \includegraphics[width=\linewidth]{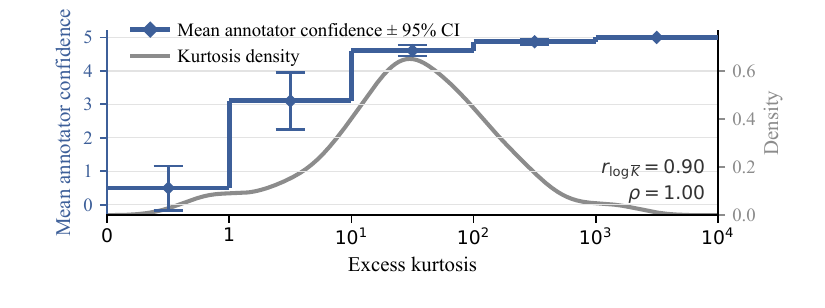}
    \caption{Human-label confidence.}
    \label{fig:kurtosis-binned-human-confidence}
  \end{subfigure}
  \hfill
  \begin{subfigure}[t]{0.49\textwidth}
    \centering
    \includegraphics[width=\linewidth]{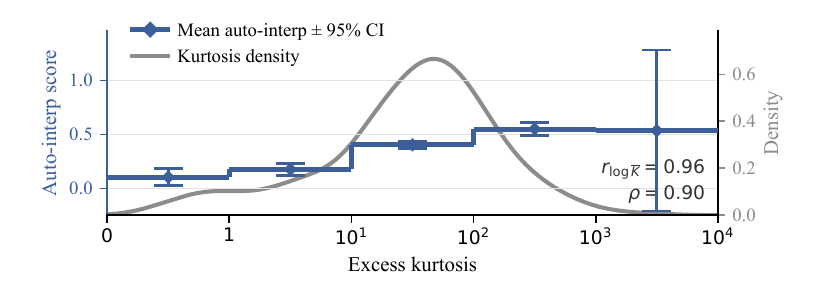}
    \caption{Automated interpretability.}
    \label{fig:kurtosis-binned-auto-interp}
  \end{subfigure}

  \caption{\textbf{Interpretability increases across excess-kurtosis bins.}
    Blue lines show bin means; gray curves show component density.
    (A) Human-label confidence for 150 audited components across three models.
  (B) Sign-corrected automated interpretability for 400 Qwen~3.5~9B Base components.}
  \label{fig:kurtosis-binned-interp}
\end{figure}

The relationship becomes clearer when components are compared across broad ranges of non-Gaussianity.
In practice, components with excess kurtosis above 1 are usually interpretable for human annotators, and those above 10 can generally be labeled with high confidence. This suggests that real interpretability changes across broad kurtosis ranges rather than point by point.
We therefore group components into excess-kurtosis bins.
Figure~\ref{fig:kurtosis-binned-human-confidence} reports the mean human-label confidence in each bin for the 150 audited components from all three models.
Figure~\ref{fig:kurtosis-binned-auto-interp} reports the mean sign-corrected automated interpretability score for the 400 sampled Qwen~3.5~9B Base components at iteration~50.
Despite using different components and evaluation methods, human confidence and automated interpretability score show the same threshold-like trend. Most ICA components lie in the 10--100 range, where both measures are higher than in the low-kurtosis bins.

\subsection{Non-Gaussian Directions Are Useful Representation Coordinates}
\label{sec:ica-evaluation}

We next test whether the recovered directions provide useful coordinates for reading and changing model representations. Four evaluations connect information preservation to behavioral control: sparse reconstruction measures how well a few coordinates retain activation information; sparse probing tests whether they support concept prediction; targeted probe perturbation (TPP) measures the selectivity of feature ablation on probe predictions; and RAVEL tests whether feature interventions change a target attribute in the model's generated answers while preserving other attributes. Probing and TPP use matched top-$k$ coordinate budgets, while RAVEL uses the attribute-specific masks learned under the SAEBench protocol. The reconstruction, probing, and TPP protocols are detailed in Appendix~\ref{app:evaluation-protocol}.
Appendix~\ref{app:using-directions} further illustrates case studies about how labeled ICA directions explain new activations across layers and support steering and clamping during generation.

\subsubsection{Sparse Reconstruction}

\begin{figure}[ht!]
  \centering
  \includegraphics[width=\textwidth]
  {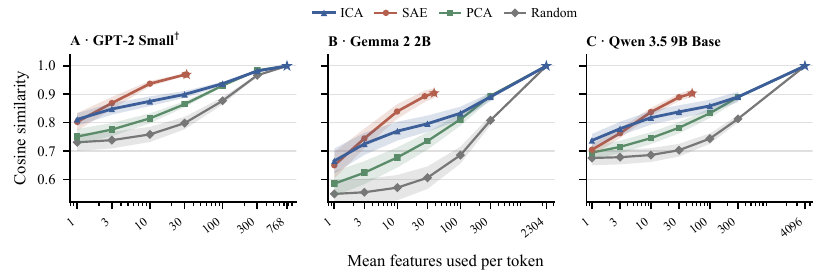}
  \caption{\textbf{Non-Gaussian directions support sparse reconstruction.} Top-$k$ reconstruction cosine on held-out activations. Lines show means $\pm1$ std across six datasets and all model layers.}
  \label{fig:reconstruction-cosine}
\end{figure}

Although ICA is not optimized for reconstruction, its largest-magnitude coordinates preserve substantial information in held-out activations. Across all three models, ICA outperforms PCA and random bases at sparse budgets and approaches the reconstruction cosine similarity of public SAEs (Figure~\ref{fig:reconstruction-cosine}). Appendix~\ref{app:reconstruction} provides results across six datasets and model depth, including normalized MSE as an alternative metric.

\subsubsection{Sparse Concept Readout}

\begin{figure}[ht!]
  \centering
  \includegraphics[width=\textwidth]
  {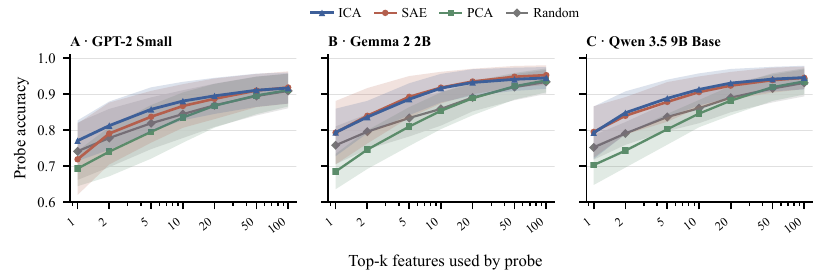}
  \caption{\textbf{Non-Gaussian directions support sparse concept readout.} SAEBench sparse-probing accuracy using the top-$k$ features.}
  \label{fig:sparse-probing}
\end{figure}

Sparse reconstruction measures how much information the coordinates retain, but not whether that information is useful for specific tasks. We therefore evaluate concept prediction using SAEBench sparse probing (Figure~\ref{fig:sparse-probing}). With only a few coordinates, ICA outperforms public SAEs on GPT-2 Small and remains competitive on Gemma~2~2B and Qwen~3.5~9B Base. On Gemma~2~2B layer~12, ICA also outperforms compact Matryoshka SAE variants~\citep{bussmann2025learning} and the training-light ITDA baseline~\citep{leask2025inference} across the tested feature budgets. Appendix~\ref{app:sparse-probing} provides the full protocol, additional baseline comparisons, and dataset-level results across all eight tasks.

\subsubsection{Selective Probe-Level Intervention}

\begin{figure}[ht!]
  \centering
  \includegraphics[width=\textwidth]
  {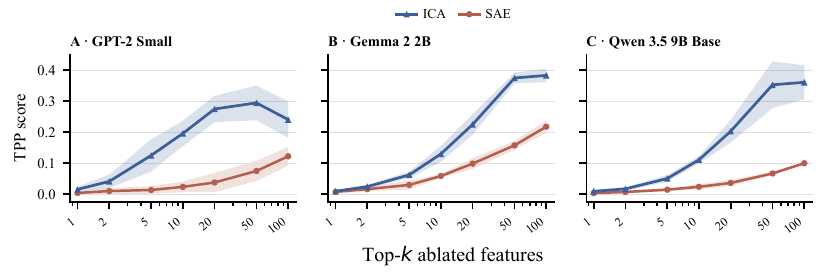}
  \caption{\textbf{ICA components support selective intervention.} TPP scores under top-$k$ feature ablation.}
  \label{fig:targeted-probe-perturbation}
\end{figure}

Sparse probing shows that concept information can be read from these coordinates; TPP tests whether it can also be selectively disrupted. We ablate the top-$k$ coordinates associated with a target probe and compare the resulting accuracy drop with the average drop on non-target probes. Across all three models and evaluated budgets, ICA consistently achieves higher TPP scores than public SAEs (Figure~\ref{fig:targeted-probe-perturbation}), indicating more selective effects on the target probe. This advantage is driven primarily by larger target-probe effects, while unintended effects remain comparatively small. Appendix~\ref{app:tpp} gives the full TPP definition and separates intended from unintended effects at the dataset level.

\subsubsection{Causal Intervention on Model Outputs}
\label{sec:ravel}

\noindent
\begin{minipage}[t]{0.54\linewidth}
\vspace{0pt}

We evaluate whether the coordinates can selectively control model-generated answers using RAVEL~\citep{huang2024ravelevaluatinginterpretabilitymethods}, which has been integrated into SAEBench~\citep{karvonen2025saebench} and MIB~\citep{mueller2025mibmechanisticinterpretabilitybenchmark} to evaluate whether changing one entity attribute preserves the model's knowledge of other attributes.
For example, editing the continent of Paris using Tokyo as a source should change the answer from Europe to Asia, while keeping its country as France. For each target attribute, we learn a binary mask over feature coordinates and replace the selected coordinates in the base representation with those from the source. We measure two success rates: \emph{Cause}, the fraction of target-attribute answers changed to the source value, and \emph{Isolation}, the fraction of other-attribute answers preserved. Their average gives the \emph{Disentangle} score, with higher values indicating more selective control. On Gemma~2~2B (layer~12), ICA achieves the highest Disentangle score (0.767), outperforming PCA (0.721) and Gemma Scope SAE (0.664) across six city and Nobel laureate attribute tasks (Figure~\ref{fig:ravel-overall}). These results show that ICA features support selective changes to generated answers, extending their utility beyond probe-level interventions. Appendix~\ref{app:ravel} provides the full protocol.
\end{minipage}\hfill
\begin{minipage}[t]{0.43\linewidth}
\vspace{-0.7em}
\centering
\includegraphics[width=\linewidth]{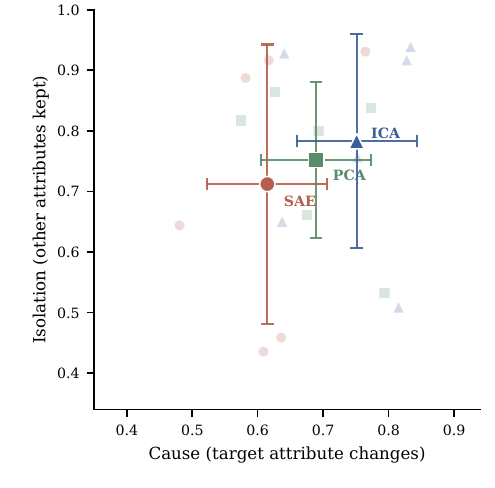}

\captionof{figure}{\textbf{Causal interventions on generated answers.} Higher is better on both axes. Evaluated on six RAVEL attribute tasks across two seeds. Large markers show means $\pm1$ std; small dots show each task.}

\label{fig:ravel-overall}
\end{minipage}

\section{Related Work}

\textbf{Sparse Feature Learning for LLM Interpretability.}
Sparse autoencoders (SAEs) have become a standard tool for decomposing language-model activations into interpretable feature coordinates, with work spanning monosemantic feature discovery, improved architectures and objectives, automated labeling, standardized evaluation, and model-wide SAE releases~\citep{huben2024sparse,bricken2023monosemanticity,templeton2024scaling,gao2025scaling,paulo2024automatically,karvonen2025saebench,makelov2025towards,rajamanoharan2024improving,rajamanoharan2024jumping,bussmann2024batchtopk,mudide2025efficient,bussmann2025learning,lieberum2024gemma,deng2026qwen}.
Despite these advances, feature discovery is still driven primarily by sparse reconstruction, while interpretability is assessed after the dictionary has been learned.
Our work asks an inverse scientific question: whether interpretable directions share a measurable property of their activation distributions that can itself serve as the discovery signal.
Our results identify non-Gaussianity of projection distributions as such a signal.

\textbf{Non-Gaussian Directions in Representation Space.}
Independent component analysis (ICA) is a classical method for recovering statistically independent, non-Gaussian directions in linear representation spaces~\citep{hyvarinen1999fast,hyvarinen2000independent}, and prior work has found interpretable ICA components in several representation spaces~\citep{bell1997independent,daubechies2009independent,yamagiwa2023discovering,musil-marecek-2024-exploring,musil2024transforminghiddenstatesbinary}.
Our work contributes a broader scientific framing for non-Gaussian feature discovery, a corrected evaluation protocol for ICA interpretability, a practical workflow for modern LLMs, and a systematic comparison with public SAEs.

\begin{table}[htbp]
  \centering
  \caption{Summary of prior gaps and our contributions.}
  \label{tab:ica-contributions}
  \small
  \renewcommand{\tabularxcolumn}[1]{m{#1}}
  \setlength{\tabcolsep}{6pt}
  \renewcommand{\arraystretch}{1.15}
  \begin{tabularx}{\textwidth}{
    @{}
    >{\raggedright\arraybackslash}X
    >{\raggedright\arraybackslash}X
    @{}
  }
    \toprule
    \textbf{Prior work and gaps}
    & \textbf{Our findings and contributions} \\
    \midrule

    SAEs discover interpretable directions through sparse reconstruction~\citep{huben2024sparse,gao2025scaling}, but it remains unclear what statistical properties distinguish interpretable directions and whether these properties can directly guide feature discovery.
    & \textbf{Discovery Principle.} We propose non-Gaussianity as an inductive bias for interpretable feature discovery, showing that interpretable directions can be recovered by directly searching for non-Gaussian projections in activation space. \\

    \addlinespace[2pt]
    An early ICA--SAE comparison reported low interpretability for ICA~\citep{huben2024sparse}. ICA's potential for LLM interpretability has long been underestimated. 
    & \textbf{Evaluation Correction.} We identify and correct a sign-sensitive evaluation bias, recovering much of the reported gap and making ICA competitive with the evaluated SAEs without changing the fitted directions. \\

    \addlinespace[2pt]
    Earlier work found only a weak positive correlation between the kurtosis of SAE latent activations and automated interpretability~\citep{huben2024sparse}, and did not examine this relationship for ICA directions.
    & \textbf{Interpretability Evidence.} We systematically examine the relationship between non-Gaussianity and interpretability across ICA fitting checkpoints and components, finding a clearer positive association. \\

    \addlinespace[2pt]
    Training separate SAE dictionaries across models and layers can be costly~\citep{lieberum2024gemma}; prior ICA studies~\citep{yamagiwa2023discovering,musil2024transforminghiddenstatesbinary} did not address practical challenges at LLM scale, like slow CPU-based fitting and convergence criteria not tailored to interp objective.
    & \textbf{Efficient Workflow.} We identify and address practical challenges in applying ICA to LLM interpretability and release ICA Lens, to our knowledge the first end-to-end workflow combining GPU-accelerated ICA fitting, improved convergence diagnostics, systematic annotation, and faithful evaluation. \\

    \addlinespace[2pt]
    ICA remains underrepresented in widely used SAE benchmarks~\citep{karvonen2025saebench} and feature-visualization platforms~\citep{neuronpedia}, leaving its practical potential insufficiently explored.
    & \textbf{Downstream Utility.} We compare ICA with public SAEs on interpretability, sparse reconstruction, sparse probing, probe perturbation, causal interventions, and illustrate its use for activation inspection and steering. \\

    \bottomrule
  \end{tabularx}
\end{table}

\section{Conclusion}
\label{sec:conclusion}

We show that non-Gaussianity can be used to discover interpretable directions directly from language-model activations. Across GPT-2 Small, Gemma~2~2B, and Qwen~3.5~9B Base, we find that these directions capture coherent, human-interpretable patterns, with stronger non-Gaussianity associated with greater interpretability. We also revisit an early ICA--SAE comparison and show that correcting for arbitrary component signs recovers much of the previously reported interpretability gap. 
Beyond interpretability, the recovered directions support new activation interpretation and generation-time steering, and perform competitively with public SAEs across sparse reconstruction, concept readout, and selective intervention, while fitting in minutes per layer and using substantially less parameter memory. We release \textbf{ICA Lens} as a practical workflow for discovering, annotating, and evaluating these directions. More broadly, our findings suggest that higher-order activation statistics offer a simple and efficient first lens for discovering interpretable structure in language models.

\nocite{liu2026agentic,han2026read}
\newpage
\bibliography{references}
\bibliographystyle{iclr2027_conference}

\clearpage
\appendix
\addtocontents{toc}{\protect\setcounter{tocdepth}{3}}
\newpage
\par\vspace*{-5em}
\tableofcontents

\section*{Appendix Roadmap}

The appendix provides additional evidence for the connection between non-Gaussianity and interpretability, details the ICA Lens workflow and evaluation protocols, and reports extended analyses of its efficiency and downstream utility. It is organized into six parts.

\begin{itemize}
  \item \textbf{Appendix~\ref{app:further-analysis}: Further Analysis and Discussion.}
    We extend the evidence beyond ICA components through concept-constructed directions and native MLP neurons, showing how non-Gaussianity relates to coherent token families and might help rank neurons for inspection. We then compare ICA and SAE representations through directional overlap and tokenwise responses, and discuss the scope, limitations, and future directions of non-Gaussian feature discovery.

     \item \textbf{Appendix~\ref{app:ica-implementation}: ICA Lens Implementation, Efficiency, and Convergence.}
    We describe activation collection, whitening, and symmetric FastICA, and report memory use, latency, and end-to-end fitting costs. We show that population-level non-Gaussianity stabilizes early, while individual directions become more reproducible across seeds with longer fitting. A sample-size ablation shows that small fitting sets can yield high kurtosis on fitting tokens but much lower kurtosis on held-out tokens.
    
    \item \textbf{Appendix~\ref{app:interpretability}: Labeling and Interpreting Non-Gaussian Directions.}
    We describe skewness-based tail selection and the human and automated evaluation protocols. We report both experts' independent annotations for the same 150 randomly sampled components, their agreement on confidence ratings, sentence-level traces, controlled counterfactual tests, and the annotation interface. We also reproduce the earlier ICA--SAE comparison, test the robustness of the sign-orientation effect across fitting budgets, corpora, and evaluators, and analyze errors in automated interpretability scores.

  \item \textbf{Appendix~\ref{app:using-directions}: Reading and Intervening with Non-Gaussian Directions.}
    We illustrate how labeled components can be used to inspect and modify model activations. Case studies analyze contextual representations of polysemous words, identify interpretable mixtures in token embeddings, and use selected coordinates for generation-time steering and clamping.

    \item \textbf{Appendix~\ref{app:quantitative-evaluation}: Full Quantitative Evaluation of Non-Gaussian Directions.}
    We report protocols and dataset-level results for sparse reconstruction, sparse probing, and targeted probe perturbation. We break down reconstruction by model depth, check the aggregate results with NMSE, compare sparse probing with compact Matryoshka SAEs and ITDA, and quantify intended and unintended probe effects. We also detail the learned masks and the Cause and Isolation metrics used to evaluate selective changes to generated answers in RAVEL.

  \item \textbf{Appendix~\ref{app:sae-baselines}: SAE Baselines and Checkpoints.}
    We record the exact public SAE releases and revisions used throughout the paper and specify the active-feature filtering used whenever individual SAE features are sampled for comparison.
\end{itemize}

\newpage

\section{Further Analysis and Discussion}
\label{app:further-analysis}

We first examine whether the connection between non-Gaussianity and interpretable structure extends beyond ICA components, using concept-constructed directions and native MLP neurons. We then compare ICA and SAE features through directional overlap and tokenwise responses. Finally, we discuss the scope of the current evidence, its limitations, and future directions for non-Gaussian feature discovery.

\subsection{Constructing Interpretable Non-Gaussian Directions from Concepts}
\label{app:concept-constructed-directions}

\begin{figure}[ht!]
  \centering
  \includegraphics[width=\textwidth]{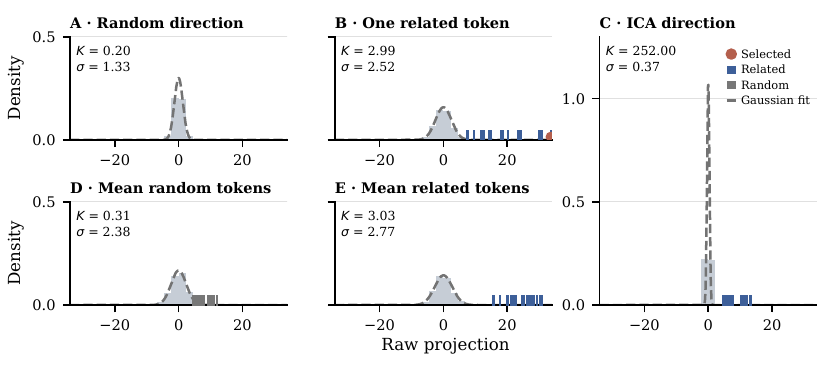}
  \caption{\textbf{Concept-constructed directions are non-Gaussian, while ICA finds a cleaner separation between the concept tail and background.}
    The panels show raw projections of all 50,256 token activations. $K$ denotes excess kurtosis and $\sigma$ denotes the standard deviation across all tokens.
    (A) A random direction produces an approximately Gaussian distribution.
    (B) The direction of one related token moves the concept family into the positive tail.
    (C) The ICA direction retains the same coherent tail while concentrating most background tokens near zero.
    (D) Averaging random-token directions does not produce a coherent tail.
  (E) Averaging the related-token directions again produces a non-Gaussian tail, but with a broader background than the ICA direction.}
  \label{fig:toy-directions}
\end{figure}

Our main experiments search for non-Gaussian directions and then examine their interpretability. Here we ask a complementary question: can simple directions constructed from a coherent token family also produce non-Gaussian projection distributions? This tests whether the observed geometry can arise without directly optimizing a non-Gaussianity objective for those directions.

\begin{table}[ht!]
  \centering
  \small
  \setlength{\tabcolsep}{8pt}
  \begin{tabular}{@{}llll@{}}
    \toprule
    \texttt{Neither}
    & \texttt{\textvisiblespace neither}
    & \texttt{\textvisiblespace Neither}
    & \texttt{\textvisiblespace either} \\

    \texttt{\textvisiblespace Either}
    & \texttt{\textvisiblespace Both}
    & \texttt{Both}
    & \texttt{Either} \\

    \texttt{\textvisiblespace both}
    & \texttt{either}
    & \texttt{both}
    & \texttt{OTH} \\

    \texttt{ither}
    & \texttt{between}
    & \texttt{\textvisiblespace Between}
    & \texttt{\textvisiblespace between} \\

    \texttt{Between}
    & \texttt{ween}
    & \texttt{\textvisiblespace latter}
    & \texttt{\textvisiblespace respectively} \\
    \bottomrule
  \end{tabular}
  \caption{\textbf{The 20 related tokens used in the concept-construction experiment.}
  \texttt{\textvisiblespace} denotes a leading space. The family is defined by the largest-magnitude projections onto the selected ICA component.}
  \label{tab:toy-related-tokens}
\end{table}

We study this question using the layer~0 activations of all 50,256 non-special GPT-2 tokens. We first select an ICA component with high excess kurtosis whose largest-magnitude tokens form a coherent family involving variants of \emph{either}, \emph{neither}, \emph{both}, \emph{between}, and related tokens. We use these tokens as a controlled concept family and ask whether simple directions constructed from them also produce non-Gaussian projection distributions. Table~\ref{tab:toy-related-tokens} lists the 20 tokens.

We compare five unit directions in Figure~\ref{fig:toy-directions}: a random direction, the activation direction of one related token (\emph{either}), the ICA direction, the mean direction of 20 random tokens, and the mean direction of the 20 related tokens. All projection statistics are computed over the same full vocabulary.

Simple concept-seeded directions already produce non-Gaussian structure. Using one related token as the direction moves the concept family into the positive tail and increases excess kurtosis from $K=0.20$ for a random direction to $K=2.99$ in Figure~\ref{fig:toy-directions}B. Averaging the 20 related-token directions gives a similar distribution with $K=3.03$ in Figure~\ref{fig:toy-directions}E. In contrast, averaging 20 random-token directions remains close to Gaussian ($K=0.31$). Thus, a direction constructed directly from a coherent concept can itself produce a non-Gaussian projection distribution.

The ICA direction shows a stronger version of the same geometry. The related-token family remains in the tail, while most other tokens are concentrated much closer to zero. The single-token and concept-mean directions have standard deviations of $\sigma=2.52$ and $\sigma=2.77$, respectively, while the ICA direction has $\sigma=0.37$ and much larger excess kurtosis ($K=252.00$). The fitted ICA direction therefore combines a coherent concept tail with a substantially more concentrated background. This separation is also useful in practice: when interpreting a direction from its top-activating examples, we want unrelated background tokens to remain well separated from the concept tokens in the tail.

\begin{keyinsight}[Key Insight]
  Non-Gaussian tails can arise from simple directions constructed from a coherent token family. In this example, ICA retains the same interpretable tail while concentrating the background more tightly. This illustrates the geometry that non-Gaussian feature discovery seeks: a coherent tail separated from most other activations.
\end{keyinsight}

\subsection{Non-Gaussianity in MLP Neurons}
\label{app:mlp-nongaussianity}

We next examine non-Gaussianity in native MLP neurons and its relationship to interpretable activation patterns. This extends our analysis from residual-stream directions to the model's existing coordinates.

We uniformly sample 100 of the 25,600 MLP neurons in layer 0 of Qwen3-32B without replacement (seed 0). We profile their signed post-SwiGLU activations, immediately before the MLP output projection, over all 10,000 Pile-10k documents (15,392,600 token positions per neuron). We compute excess kurtosis from the full activation distribution and inspect the 20 largest and 20 smallest activations.
For each neuron, we measure lexical concentration $C_j$ as the largest fraction of a single token ID among the 20 extreme positions on either side. Excess kurtosis is positively associated with this measure (Spearman $\rho_s=0.55$), and mean concentration increases from 45.8\% in the lowest kurtosis quartile to 80.8\% in the highest (Table~\ref{tab:mlp-kurtosis-quartiles}). Retaining only the most extreme available position per document in each tail and recomputing concentration yields a similar association ($\rho_s=0.53$). These results suggest that kurtosis may be useful for prioritizing neurons with concentrated lexical responses for inspection.

\begin{table}[ht!]
    \centering
    \caption{Higher-kurtosis MLP neurons show greater lexical concentration. Each quartile contains 25 neurons.}
    \label{tab:mlp-kurtosis-quartiles}
    \small
    \begin{tabular}{lrr}
        \toprule
        Kurtosis quartile & Excess-kurtosis range & Mean $C_j$ \\
        \midrule
        Q1 (lowest)  & 11.3--26.5         & 45.8\% \\
        Q2           & 27.8--146.2        & 53.2\% \\
        Q3           & 157.8--3,133.7     & 79.8\% \\
        Q4 (highest) & 3,941.2--39,454.8  & 80.8\% \\
        \bottomrule
    \end{tabular}
\end{table}

\begin{figure}[ht!]
    \centering
    \includegraphics[width=\linewidth]{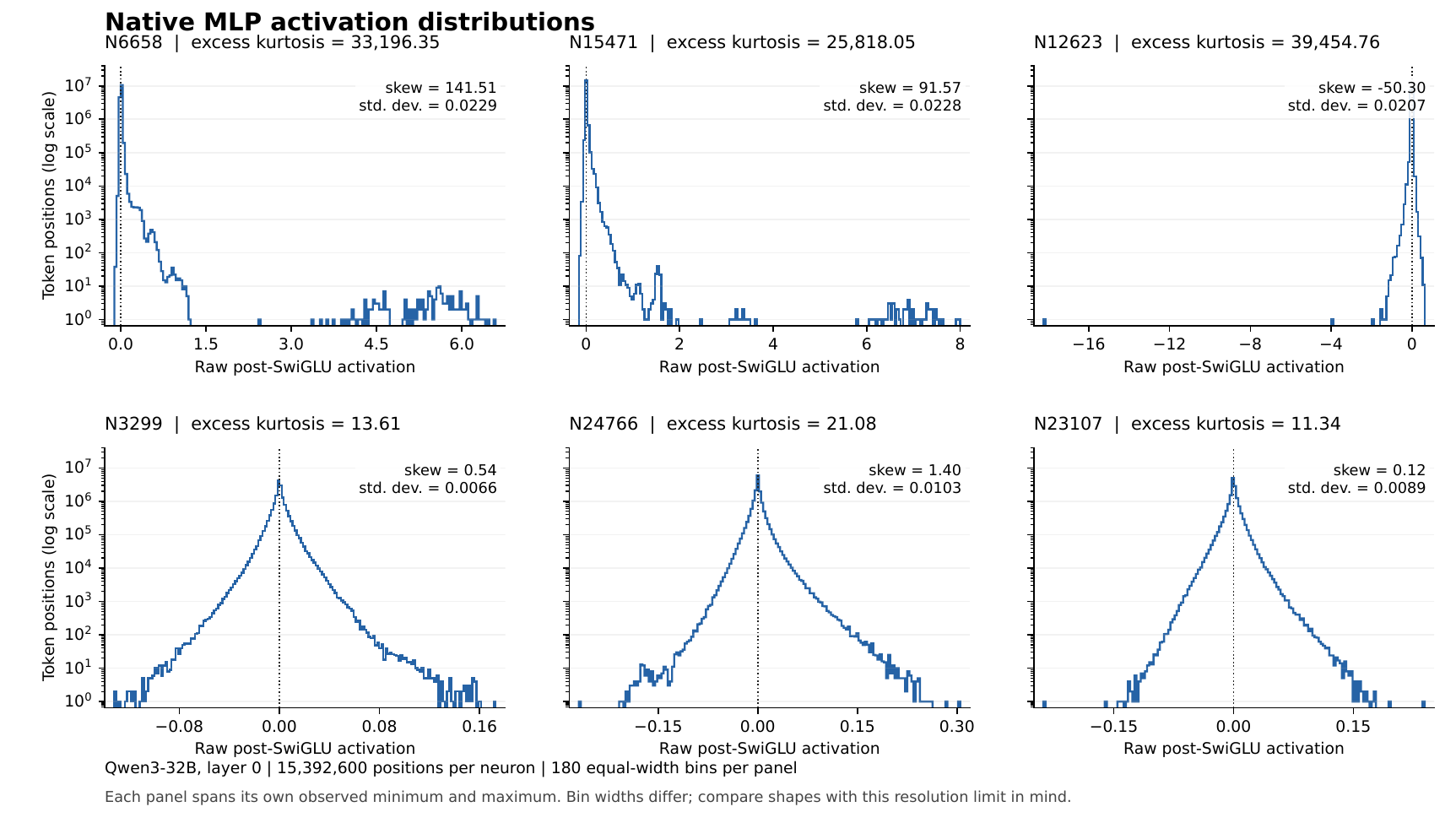}
    \caption{\textbf{Representative activation distributions of native MLP neurons.} Each panel uses 180 equal-width bins over the neuron's observed range, with signed activation on the horizontal axis and token counts on a logarithmic vertical axis. High-kurtosis examples in the upper row exhibit pronounced asymmetric tails; lower-kurtosis ones in the lower row are more balanced around zero.}
    \label{fig:mlp-neuron-distributions}
\end{figure}

Figure~\ref{fig:mlp-neuron-distributions} shows representative activation distributions, and Table~\ref{tab:mlp-neuron-examples} summarizes the observed patterns. We find that high-kurtosis neurons often exhibit recognizable lexical patterns. All 20 positive extremes of N6658 occur on the copyright-symbol token, and those of N15471 occur on \emph{COVID}. N3236 responds to exception-related identifiers, while all 20 negative extremes of N19765 occur on \emph{similar}.
It is worth noting that kurtosis can guide neuron selection, while tail inspection is necessary to distinguish recurring patterns from isolated extremes. For example, N12623 has the highest kurtosis in the sample, but a single position contributes 99.4\% of its fourth central moment. Its extreme contexts mix natural language and programming identifiers, making it difficult to assign a single coherent interpretation.

Another interesting finding is that the minimum excess kurtosis among the sampled MLP neurons is 11.34, whereas random residual-stream directions in our main experiments have values close to zero. Since MLP neurons form a privileged basis, this may reflect some inherent interpretable structure in these coordinates.

\begin{table}[ht!]
    \centering
    \caption{Illustrative MLP neurons, including exceptions to a strict ordering by kurtosis. Tail signs refer to the recorded activations, and descriptions summarize the observed extreme contexts.}
    \label{tab:mlp-neuron-examples}
    \small
    \setlength{\tabcolsep}{4pt}
    \renewcommand{\arraystretch}{1.12}
    \begin{tabularx}{\linewidth}{rrcX}
        \toprule
        Neuron & Excess kurtosis & Tail & Observed pattern \\
        \midrule
        6658  & 33,196.35 & $+$ & Copyright-symbol token (20/20 extremes). \\
        15471 & 25,818.05 & $+$ & \emph{COVID} token (20/20). \\
        3236  & 14,265.50 & $+$ & Exception-related identifiers. \\
        19765 & 4,505.21  & $-$ & \emph{similar} token (20/20). \\
        \midrule
        3299  & 13.61 & $+$ & \emph{.amazon} in URL contexts (17/20). \\
        24766 & 21.08 & $-$ & \emph{\_width} in Android layouts (18/20). \\
        23107 & 11.34 & Both & Mixed tokens and fragments; no clear single pattern. \\
        12623 & 39,454.76 & Both & Mixed contexts; fourth moment dominated by one position. \\
        \bottomrule
    \end{tabularx}
\end{table}

\begin{keyinsight}[Key Insight]
Higher excess kurtosis is associated with more concentrated lexical responses and may help prioritize MLP neurons for interpretation. This preliminary analysis motivates future work that directly quantifies neuron interpretability and examines its relationship with kurtosis.
\end{keyinsight}

\newpage
\subsection{Relationship Between Non-Gaussian Directions and SAE Features}
\label{app:ica-sae-relationship}

ICA and sparse autoencoders are fitted with very different objectives: ICA searches directly for non-Gaussian projections, while SAEs learn an overcomplete dictionary through sparse reconstruction. We therefore ask how much structure the two methods recover in common. We compare them from two complementary perspectives: geometric alignment between directions and tokenwise response patterns on actual sequences.

\subsubsection{Directional Overlap and Complementarity}
\label{app:ica-sae-overlap}

We first ask how closely ICA directions correspond to directions in the much larger SAE dictionaries. For each layer, we compare every ICA residual-space writing direction $d_j$ with every decoder direction in the corresponding released SAE. Both sets are $\ell_2$-normalized, and for each ICA direction we record its maximum absolute cosine similarity with any SAE decoder direction. The absolute value removes the arbitrary sign of an ICA component.

As a matched reference, we sample the same number of isotropic Gaussian directions directly in residual-stream space and compare them with the same SAE dictionary. Each layer therefore contains exactly $d$ ICA queries and $d$ random queries, where $d$ is the residual dimension. The random reference uses the same residual space, SAE dictionary, and number of queries as the ICA comparison.

\begin{figure}[ht!]
  \centering
  \includegraphics[width=\textwidth]{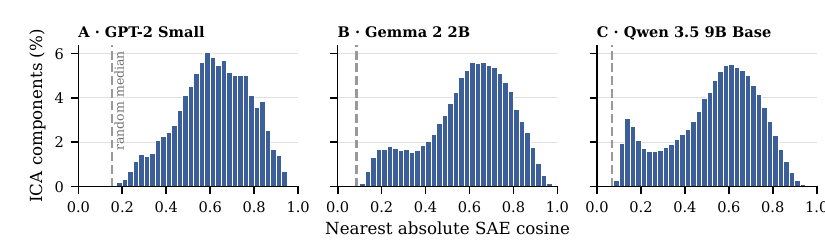}
  \caption{ICA directions show substantial but incomplete overlap with SAE features. Each panel pools all ICA components across the evaluated layers of one model and shows the maximum absolute cosine with any decoder direction from the corresponding same-layer SAE. Dashed gray lines mark the median obtained from the same number of isotropic random residual-space directions. ICA distributions are shifted well above the random references, while their broad spread also reveals directions with much weaker SAE alignment.}
  \label{fig:ica-sae-overlap-distribution}
\end{figure}

\begin{figure}[ht!]
  \centering
  \includegraphics[width=\textwidth]{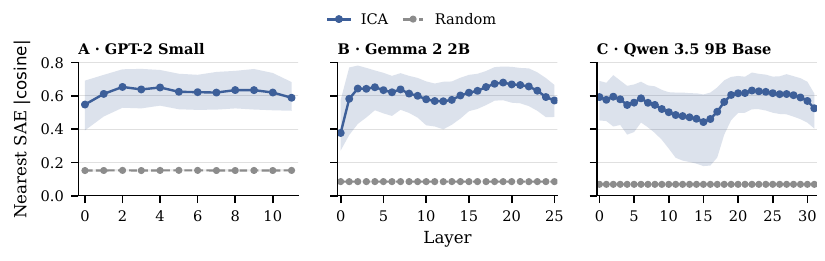}
  \caption{ICA-to-SAE directional overlap across model depth. Blue points show the median maximum absolute cosine between ICA writing directions and same-layer SAE decoder directions, with bands showing interquartile ranges. Gray points and bands report the same statistics for matched isotropic random directions in residual space. ICA remains clearly above the random reference throughout model depth, while the strength of alignment varies across layers.}
  \label{fig:ica-sae-overlap-layers}
\end{figure}

Figure~\ref{fig:ica-sae-overlap-distribution} shows the two methods recover related but not identical structure. Non-Gaussian search recovers many directions also represented by SAEs, while also exposing additional directions that are geometrically distinct from their nearest SAE features.
The same pattern persists across model depth. Figure~\ref{fig:ica-sae-overlap-layers} reports the median nearest-SAE similarity separately for every layer. ICA remains well above the matched random reference throughout all three models, showing that the overlap is not confined to a small subset of layers.

\begin{keyinsight}[Key Insight]
  ICA and SAEs recover overlapping but complementary structure. Many non-Gaussian directions have close SAE counterparts, while others remain only weakly aligned with the released SAE dictionaries. This supports direct non-Gaussian search as a useful first lens for model representations: it can quickly expose substantial interpretable structure before committing to a larger learned dictionary.
\end{keyinsight}

\subsubsection{Tokenwise Response Structure}
\label{app:activation-patterns}

Directional cosine measures geometric similarity, but it does not show how ICA and SAE features respond across an actual sequence. We therefore compare their tokenwise response patterns on the same sentence:
\begin{quote}
  \small
  Maya stopped at the bank before the trip, waiting in line to deposit a check and withdraw enough cash for the weekend.
\end{quote}

For each token, we identify the highest-response ICA component and the highest-response SAE feature. ICA selection uses the largest absolute component score, while SAE selection uses the largest native feature activation. We keep each selected component or feature once, even if it is the strongest response at multiple token positions, and plot its response across the full sentence. For visualization, ICA scores are shown in absolute value, SAE features use their native post-activations, and each row is normalized by its own maximum.

\begin{figure}[ht!]
  \centering
  \includegraphics[width=0.80\textwidth]{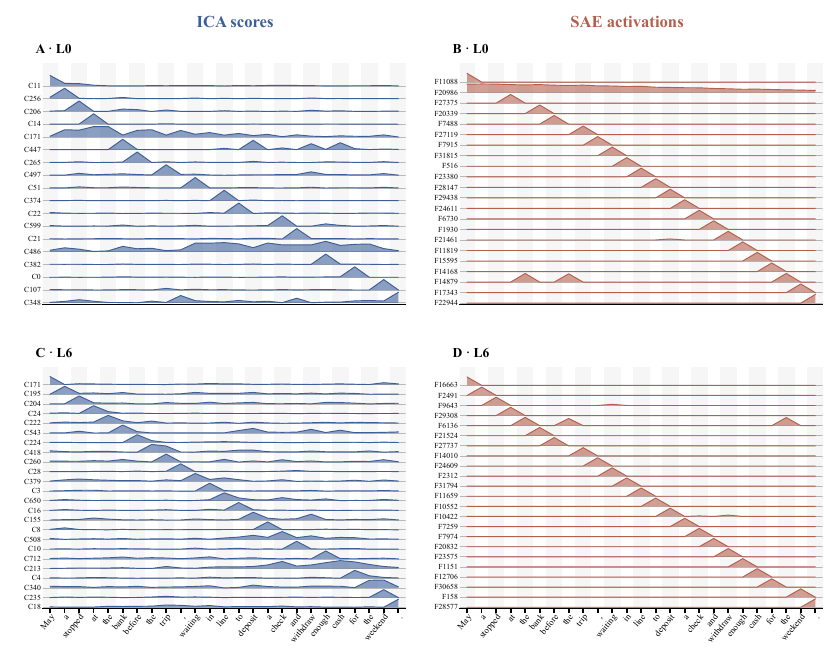}
  \caption{Tokenwise response patterns for GPT-2 Small at layers 0 and 6.}
  \label{fig:activation-pattern-gpt2}
\end{figure}

\begin{figure}[ht!]
  \centering
  \includegraphics[width=0.8\textwidth]{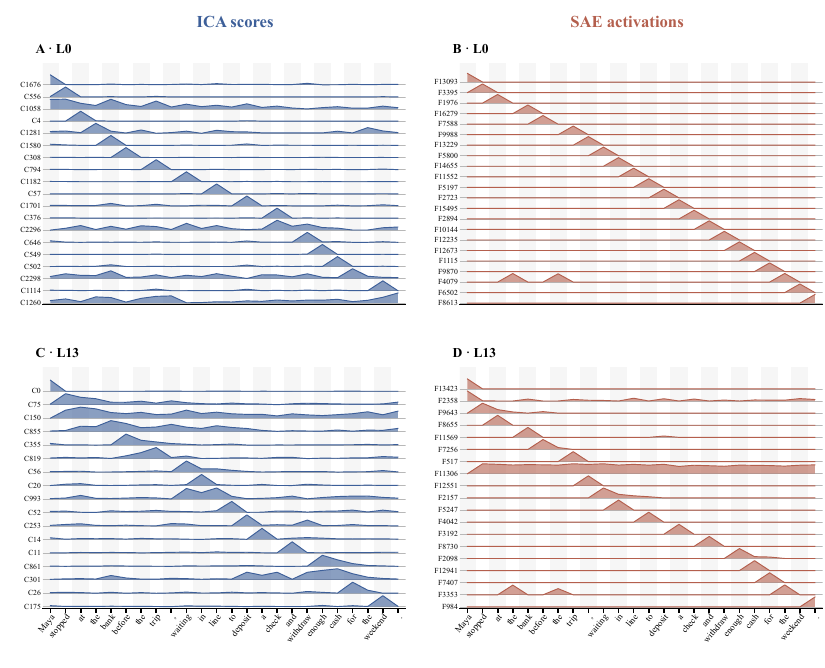}
  \caption{Tokenwise response patterns for Gemma~2~2B at layers 0 and 13.}
  \label{fig:activation-pattern-gemma2}
\end{figure}

\begin{figure}[ht!]
  \centering
  \includegraphics[width=0.8\textwidth]{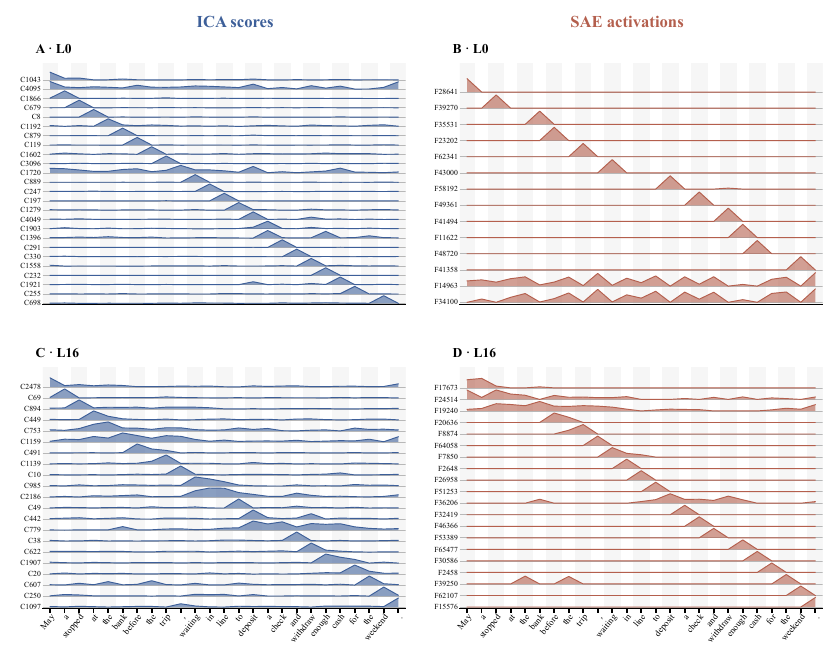}
  \caption{Tokenwise response patterns for Qwen 3.5 9B Base at layers 0 and 16.}
  \label{fig:activation-pattern-qwen9b-full}
\end{figure}

Figures~\ref{fig:activation-pattern-gpt2},~\ref{fig:activation-pattern-gemma2}, and~\ref{fig:activation-pattern-qwen9b-full} show a similar range of tokenwise response patterns across models and layers. Both ICA and SAE representations contain sharply localized responses concentrated around individual tokens, and broader responses that extend across several token positions. The relative mixture varies across models, layers, and selected components or features, but both types of structure appear consistently in the two representations.


\begin{keyinsight}[Key Insight]
  ICA and SAE representations show a similar range of tokenwise response patterns across models and layers. Both contain sharply localized responses as well as components or features whose activation extends across multiple tokens.
\end{keyinsight}

\subsection{Scope, Limitations, Future Work, and Impacts}
\label{app:scope-limitations}

\textbf{Understanding and Learning Interpretable Non-Gaussian Geometry.}
Our paper provides an empirical explanation for why non-Gaussianity aligns with interpretability. From the perspective of human annotation, we often try to explain a direction using its top-$k$ activating tokens. A nearly Gaussian direction is difficult to interpret from only its top-$k$ tokens, while a non-Gaussian direction has at least the potential to isolate some special tokens in its tail. In LLM residual-stream geometry, we find that the tails of many non-Gaussian directions correspond to interpretable concepts. This suggests a possible geometric picture in which tokens associated with the same concept form a cluster that is separated from unrelated tokens. An important question for future work is why language-model pretraining learns this structure, how it helps reduce the training loss, and whether we can use this structure during pretraining to make models more interpretable by design.

\textbf{Compact FastICA Basis as a First Lens.}
We use FastICA as a practical way to test our hypothesis that non-Gaussianity can serve as a feature-discovery signal. However, standard FastICA is compact: in a $d$-dimensional fitting space, it returns at most $d$ components. This makes the resulting basis efficient to fit and inspect, but it does not provide the large overcomplete feature vocabulary that modern SAEs are designed to learn. We therefore position ICA as a compact first lens for language-model activations, rather than as a replacement for sparse autoencoders. ICA Lens asks how much interpretable structure is already visible from activation geometry, helping analysts decide where heavier dictionary learning is worth the cost.

A natural extension is to explore overcomplete ICA and other ICA formulations that change the capacity, optimization procedure, or non-Gaussianity objective. Examples include overcomplete ICA, preconditioned or adaptive FastICA, deflationary ICA, JADE, Infomax, extended Infomax, and objectives designed for heavy-tailed distributions~\citep{podosinnikova2019overcomplete,ablin2018faster,miettinen2018fica,cardoso1993blind,bell1995information,lee1999independent,spurek2018ica}.
This perspective may also inspire new ways to design and improve SAE objectives.

\textbf{Beyond single-layer residual streams.}
Our experiments focus on token-level residual-stream activations analyzed one layer at a time, but ICA Lens is lightweight enough to support broader and more repeated analysis.
At a finer granularity, it could be applied to attention-head outputs, MLP outputs, or other submodules; across depth, normalized activations from multiple layers could be analyzed jointly to identify directions that emerge, persist, or transform.
A particularly promising use case is longitudinal feature discovery across training checkpoints.
Because ICA Lens can be refit quickly, the same analysis could be repeated throughout pretraining or post-training to track when a direction appears, strengthens, changes meaning, or disappears, without training a separate overcomplete dictionary at every checkpoint.
This could turn non-Gaussian feature discovery from a one-shot analysis into a practical tool for studying how internal structure emerges and changes during training.

The same idea extends beyond individual token states.
Pooling or segmenting activations could define representations at the sentence, reasoning-step, or trajectory level, allowing components to be tracked over a reasoning process.
Comparisons across base and instruction-tuned models, reinforcement-learning stages, or successive checkpoints could reveal which latent directions are preserved, created, or suppressed by training.
At the agent level, analogous analyses over observation--reasoning--action trajectories could search for directions associated with planning, uncertainty, recovery, or failure.
Longer term, joint analysis of language, vision, and action representations may reveal latent factors shared across modalities.

\newpage

\section{ICA Lens Implementation, Efficiency, and Convergence}
\label{app:ica-implementation}

This section describes the implementation of ICA Lens and studies its practical behavior at LLM scale. We first specify activation collection, whitening, and symmetric FastICA fitting, including the reading and writing directions used throughout the paper. We then measure the memory, latency, and end-to-end fitting cost of the resulting coordinate system. Finally, we examine fitting dynamics and stability across iterations and random initializations. Our results reveal an important distinction between population-level and component-level convergence. Population non-Gaussianity rises quickly and largely saturates early in fitting, while individual component identities continue to stabilize over longer runs.

\subsection{Activation Collection and FastICA Fitting}
\label{app:ica-fitting}

\paragraph{Activation collection.}

We collect fitting activations from the training split of \texttt{NeelNanda/pile-10k}. For each of its 10,000 documents, we tokenize the text with the corresponding model tokenizer and retain up to the first 1,024 tokens. Each document is processed as a full sequence, and we cache the residual-stream activation at every retained token position, so each activation contains the preceding document context available at that position. Unless stated otherwise, we uniformly sample one million token activations for each layer fit.

We fit a separate ICA basis to every evaluated residual-stream layer in GPT-2 Small, Gemma~2~2B, and Qwen~3.5~9B Base. For the projection-distribution analyses in the main text, we additionally sample a separate set of one million held-out activations per layer. These held-out activations are not used to fit ICA and are used to measure quantities such as excess kurtosis after fitting.

\paragraph{Centering and whitening.}

Let $X\in\mathbb{R}^{n\times d}$ denote the sampled activation matrix for one layer, with one residual-stream activation per row. We first subtract the empirical mean,
\begin{equation}
  \mu=\frac{1}{n}X^\top\mathbf{1},
  \qquad
  X_c=X-\mathbf{1}\mu^\top.
\end{equation}
We then compute the covariance eigendecomposition
\begin{equation}
  \frac{1}{n}X_c^\top X_c
  =
  U\Lambda U^\top
\end{equation}
and whiten the centered activations as
\begin{equation}
  Z=X_cU\Lambda^{-1/2}.
\end{equation}
This gives $\frac{1}{n}Z^\top Z=I_d$, removing the empirical mean and second-order correlations while preserving the full residual dimension.

\paragraph{Symmetric FastICA.}

After whitening, the remaining linear freedom is an orthogonal rotation. We recover this rotation with symmetric FastICA using the logcosh contrast~\citep{hyvarinen1999fast,hyvarinen2000independent}. Let $W\in\mathbb{R}^{d\times d}$ contain the current ICA axes as unit-norm rows and let
\begin{equation}
  Y=ZW^\top
\end{equation}
denote the corresponding component scores. Symmetric FastICA updates all $d$ components in parallel using a fixed-point step derived from the chosen non-Gaussianity contrast. For the logcosh contrast, the update uses $g(y)=\tanh(y)$ and its derivative $g'(y)=1-\tanh^2(y)$. In matrix form, the parallel update can be written as
\begin{equation}
  \widetilde{W}
  =
  \frac{1}{n}g(Y)^\top Z
  -
  \operatorname{diag}
  \left(
    \frac{1}{n}\sum_{i=1}^{n}g'(Y_{i,:})
  \right)W.
\end{equation}
The updated axes are then symmetrically decorrelated,
\begin{equation}
  W
  \leftarrow
  \left(
    \widetilde{W}\widetilde{W}^\top
  \right)^{-1/2}
  \widetilde{W},
\end{equation}
which keeps the recovered axes mutually orthogonal in whitened space while allowing all components to be optimized jointly. During this decorrelation step, eigenvalues of the component Gram matrix are floored at machine epsilon before taking the inverse square root for numerical stability.

All main experiments use symmetric FastICA for 50 iterations. Using the same fixed iteration budget across models and layers keeps the feature-discovery procedure consistent throughout the main evaluation. Longer runs of up to 200 or 500 iterations are used only in the convergence and stability analyses in Appendices~\ref{app:ica-stability} and~\ref{app:ica-convergence}.

\paragraph{Reading and writing directions.}

FastICA returns axes in whitened coordinates, but model inspection and intervention take place in the original residual-stream space. For an ICA axis $w_j$, the score of a centered residual activation $x_c$ can be written as
\begin{equation}
  y_j=x_c^\top r_j,
  \qquad
  r_j=U\Lambda^{-1/2}w_j,
\end{equation}
where $r_j$ is the residual-space \emph{reading direction}. Conversely, changing component coordinate $j$ by one unit maps back to residual space along the \emph{writing direction}
\begin{equation}
  d_j=U\Lambda^{1/2}w_j.
\end{equation}
The two sets form a dual pair,
\begin{equation}
  r_i^\top d_j=\delta_{ij},
\end{equation}
although neither set is generally orthogonal in the original residual-stream coordinates. We retain $\mu$, $U$, $\Lambda$, and the fitted ICA rotation so that component scoring, reconstruction, vocabulary readout, and intervention all use the same fitted coordinate system.

\subsection{Practical Efficiency}
\label{app:ica-efficiency}

ICA Lens produces a complete coordinate system with modest storage and application cost. Table~\ref{tab:ica-efficiency} compares the fitted ICA representation with the matched public SAE at one middle layer per model. Memory is the persistent float32 parameter and buffer storage required for one fitted layer. Latencies exclude language-model inference and report the median batch-amortized time per activation over 100 synchronized CUDA trials after 20 warm-up passes, using batches of 256 float32 residual vectors.

The ICA fitting time is the end-to-end wall-clock time required to fit one layer from one million sampled activations for 50 FastICA iterations. It includes covariance estimation, eigendecomposition, whitening, FastICA optimization, and construction of the final fitted transforms rather than only the FastICA iteration loop. We report the range over two complete fits on a single NVIDIA GB10. For the public SAE baselines, checkpoint-specific training times are not available, so the table compares their stored representation and application costs rather than training cost.

\begin{table}[!ht]
  \centering
  \caption{\textbf{Practical efficiency of ICA and the matched public SAE baselines.} Memory is persistent float32 parameter and buffer storage for one layer. Encode--decode latency includes mapping the representation back to residual space. Parentheses report the SAE-to-ICA ratio. ICA fitting time is end-to-end wall-clock time for a complete one-million-activation, 50-iteration fit.}
  \label{tab:ica-efficiency}
  \resizebox{\textwidth}{!}{%
    \begin{tabular}{@{}lccccc@{}}
      \toprule
      Model & Layer & Memory / layer (MiB) & Encode ($\mu$s) & Encode--decode ($\mu$s) & ICA fit (min) \\
      & & ICA / SAE & ICA / SAE & ICA / SAE & \\
      \midrule
      GPT-2 Small & 6 & 4.50 / 192.25 ($42.7\times$) & 0.206 / 11.616 ($56.4\times$) & 0.406 / 15.418 ($38.0\times$) & 0.92 \\
      Gemma 2 2B & 13 & 40.51 / 288.20 ($7.1\times$) & 0.998 / 6.748 ($6.8\times$) & 2.055 / 11.903 ($5.8\times$) & 5.51--5.57 \\
      Qwen 3.5 9B Base & 16 & 128.02 / 2048.52 ($16.0\times$) & 2.962 / 45.634 ($15.4\times$) & 5.839 / 75.975 ($13.0\times$) & 12.19--13.13 \\
      \bottomrule
    \end{tabular}%
  }
\end{table}

Across the three models, the fitted ICA representation uses $7.1$--$42.7\times$ less persistent memory than the matched public SAEs. Encoding is $6.8$--$56.4\times$ faster, and a full encode--decode pass is $5.8$--$38.0\times$ faster. End-to-end fitting takes less than one minute per layer for GPT-2 Small, about 5.5 minutes for Gemma~2~2B, and 12--13 minutes for Qwen~3.5~9B Base. The resulting coordinate system can therefore be fitted across many layers without a separate representation-learning training pipeline. ICA and the public SAEs have different capacities, since ICA contains exactly $d$ components while the SAEs are overcomplete; Appendix~\ref{app:scope-limitations} discusses how this difference affects their practical roles.

\FloatBarrier

\subsection{Fitting Dynamics}
\label{app:ica-fitting-dynamics}

\paragraph{Population non-Gaussianity is recovered quickly.}

\begin{figure}[ht]
  \centering
  \includegraphics[width=\textwidth]{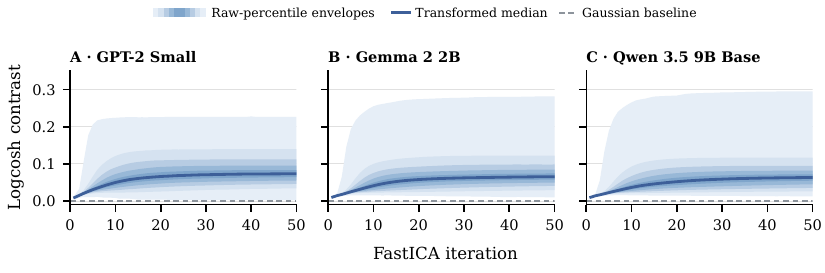}
  \caption{\textbf{FastICA rapidly recovers non-Gaussian structure across model scales.} Logcosh contrast is $C_{\mathrm{logcosh}}(y)=|\mathbb{E}[\log\cosh(y)]-\mathbb{E}[\log\cosh(\xi)]|$ for a standardized projection $y$ and $\xi\sim\mathcal{N}(0,1)$. The solid line is obtained by transforming the median raw logcosh statistic. From lightest to darkest, the nested bands transform the raw p0--p100, p10--p90, p20--p80, p30--p70, and p40--p60 intervals in the same way. Curves are averaged equally across fitted layers of each model, and the dashed line marks the Gaussian reference.}
  \label{fig:fitting-curves}
\end{figure}

Figure~\ref{fig:fitting-curves} tracks the FastICA logcosh contrast throughout the 50-iteration fitting budget used in the main experiments. Across GPT-2 Small, Gemma~2~2B, and Qwen~3.5~9B Base, the contrast increases most rapidly during the early iterations and changes much more slowly later in the run. The same pattern appears across substantially different residual dimensions, suggesting that the dominant non-Gaussian structure is found early in optimization.

The component population remains heterogeneous throughout fitting. Some axes become strongly non-Gaussian, while others stay much closer to the Gaussian reference. This broad spread is expected for a complete basis: FastICA must return $d$ directions, even though the strength of higher-order structure is not uniform across the residual space.

\FloatBarrier

\paragraph{The same fitting pattern appears across model depth.}
\label{app:fitting-curves-layerwise}

\begin{figure}[h]
  \centering
  \includegraphics[width=\textwidth]{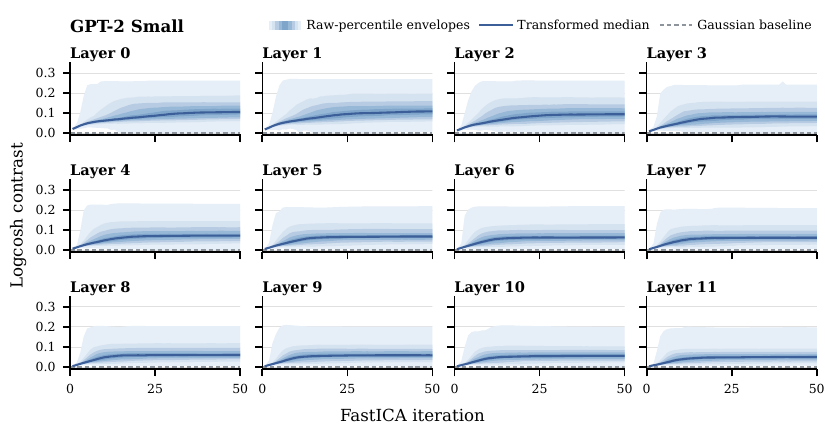}
  \caption{\textbf{Layer-wise FastICA fitting curves for GPT-2 Small.} Within each layer, the solid line and transformed percentile envelopes follow Figure~\ref{fig:fitting-curves}. A shared vertical scale makes differences in final contrast and component spread visible across depth.}
  \label{fig:fitting-curves-layerwise-gpt2}
\end{figure}

\begin{figure}[p]
  \centering
  \includegraphics[width=\textwidth]{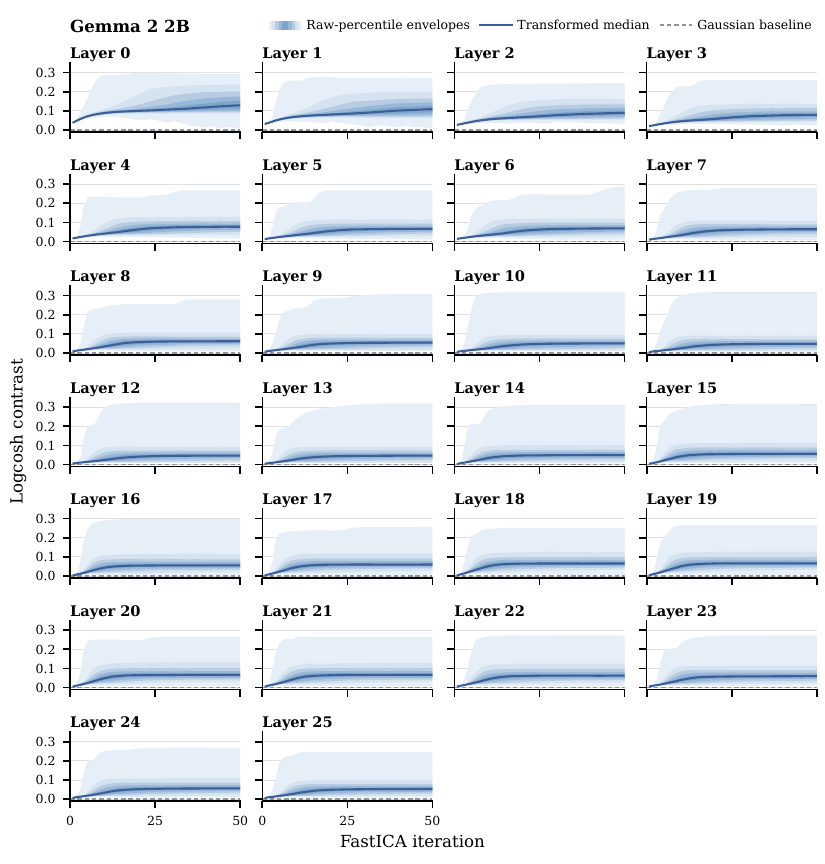}
  \caption{\textbf{Layer-wise FastICA fitting curves for Gemma~2~2B.} We use the same quantities, percentile envelopes, and axis limits as in Figure~\ref{fig:fitting-curves}.}
  \label{fig:fitting-curves-layerwise-gemma2}
\end{figure}

\begin{figure}[p]
  \centering
  \includegraphics[width=\textwidth]{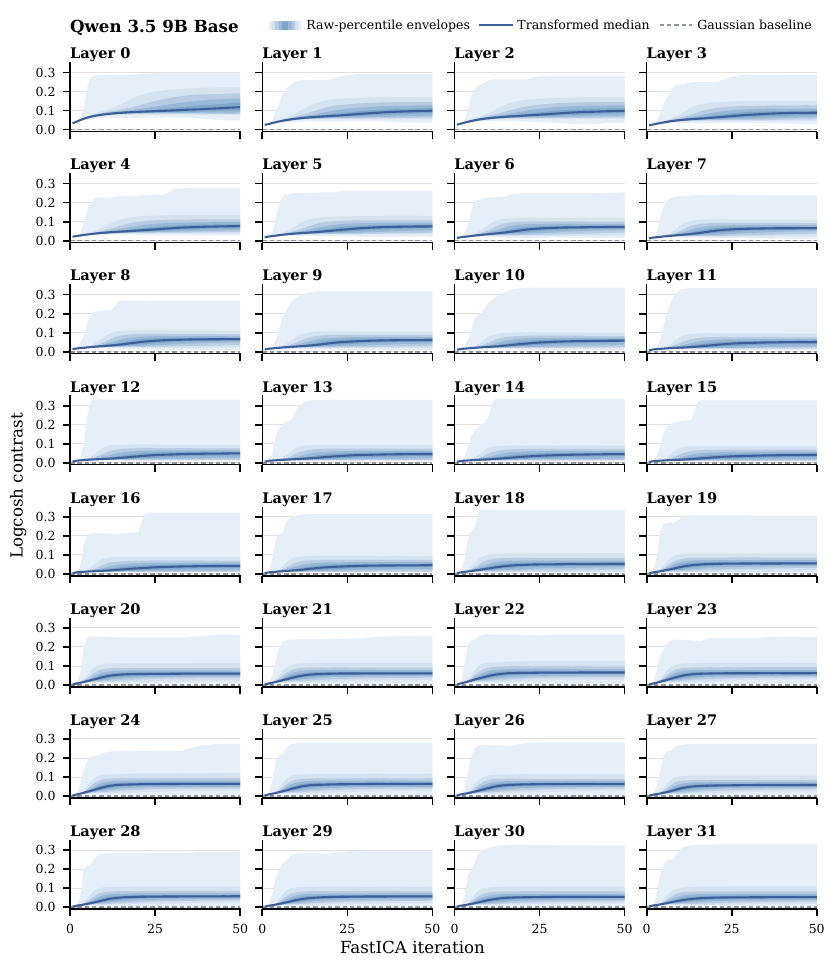}
  \caption{\textbf{Layer-wise FastICA fitting curves for Qwen~3.5~9B Base.} We use the same quantities, percentile envelopes, and axis limits as in Figure~\ref{fig:fitting-curves}.}
  \label{fig:fitting-curves-layerwise-qwen9b}
\end{figure}

Figures~\ref{fig:fitting-curves-layerwise-gpt2}--\ref{fig:fitting-curves-layerwise-qwen9b} show the corresponding trajectories separately for every layer. The rapid early increase is visible throughout model depth rather than being driven by a small subset of layers. Final contrast and component spread vary with depth, but the broad fitting dynamics remain consistent across layers and models.

\FloatBarrier

\subsection{Seed Stability and Iteration Budgets}
\label{app:ica-stability}

The fitting curves above describe the component population, but they do not show whether individual directions are reproducible across random initializations. We therefore measure component-level stability directly on GPT-2 Small layer~6 using five independent FastICA initializations and iteration budgets of 20, 50, 200, and 500. The one million fitting activations and all other fitting settings are held fixed across runs.

Before matching, we explicitly normalize every residual-space reading direction to unit $\ell_2$ norm. For each pair of seeds, we compute the absolute cosine-similarity matrix between the two fitted bases and use Hungarian matching to obtain the one-to-one assignment that maximizes total similarity across all 768 directions.

\begin{figure}[h]
  \centering
  \includegraphics[width=\textwidth]{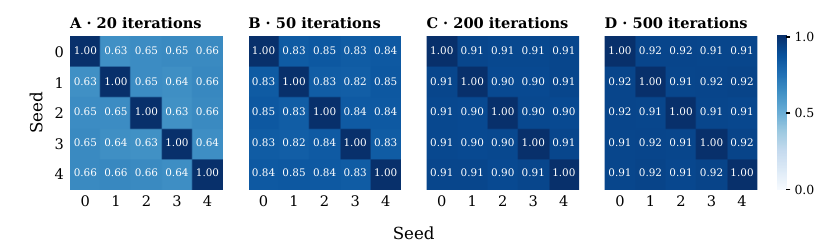}
  \caption{\textbf{Individual components become more reproducible with longer fitting.} Cross-seed stability for GPT-2 Small layer~6 at four FastICA iteration budgets. Each cell reports the median absolute cosine similarity across 768 unit-normalized reading directions after one-to-one Hungarian matching between two seeds. All fits use the same one million activation samples. Median agreement rises substantially from 20 to 50 and 200 iterations, with only a small additional increase from 200 to 500. Diagonal entries denote self-comparisons.}
  \label{fig:seed-stability-heatmap}
\end{figure}

Figure~\ref{fig:seed-stability-heatmap} shows that cross-seed agreement improves substantially with fitting time. At 20 iterations, the median matched cosine between different seeds is approximately $0.63$--$0.66$. It rises to approximately $0.82$--$0.85$ at 50 iterations and $0.90$--$0.91$ at 200 iterations. Increasing the budget from 200 to 500 iterations produces only a small additional gain, with medians around $0.91$--$0.92$. Most of the improvement in component identity therefore occurs before 200 iterations.

\begin{figure}[h]
  \centering
  \includegraphics[width=0.9\textwidth]{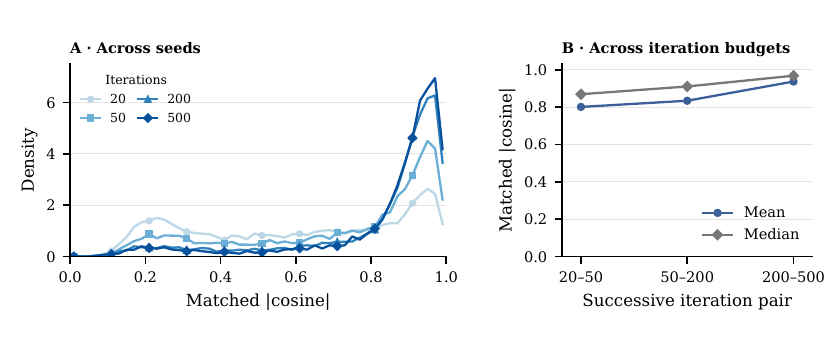}
  \caption{\textbf{Most component-level stabilization occurs by 200 iterations.} All comparisons use 768 unit-normalized GPT-2 Small layer~6 reading directions and one-to-one Hungarian matching. (A) Cross-seed distributions of matched absolute cosine similarity at each iteration budget, averaged equally across the ten distinct seed pairs. Curves connect the centers of 50 equal-width bins on $[0,1]$. (B) Within-seed agreement between successive iteration budgets. Mean and median matched similarities are computed and averaged across five seeds. Later iteration budgets produce progressively smaller changes to the recovered basis.}
  \label{fig:seed-stability-density}
\end{figure}

Figure~\ref{fig:seed-stability-density}A shows the full matched-similarity distributions underlying these medians. Longer runs increasingly concentrate components near high cosine similarity, while a smaller low-similarity tail remains. Figure~\ref{fig:seed-stability-density}B gives a complementary within-seed view by matching directions from successive iteration budgets. Agreement is highest between 200 and 500 iterations, showing that the recovered basis changes progressively less late in fitting.
The 50-iteration configuration used in the main experiments therefore recovers a largely shared basis across seeds, while longer fitting further improves the reproducibility of exact component identities.

\FloatBarrier

\subsection{Fitting Sample Size and Overfitting}
\label{app:ica-fitting-sample-size}

\begin{figure}[ht!]
  \centering
  \includegraphics[width=1\textwidth]{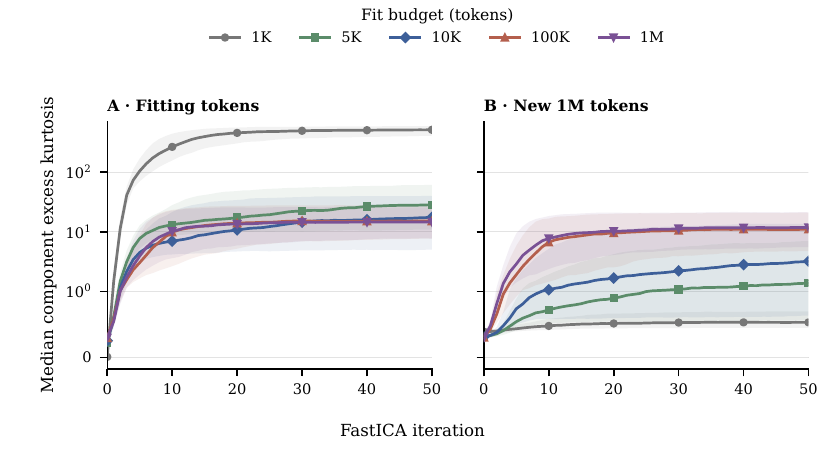}
  \caption{\textbf{FastICA can overfit when fitted on too few tokens.} Median excess kurtosis across 768 GPT-2 Small components over FastICA iterations, measured on (A) each fit's own fitting tokens and (B) a shared set of 1M held-out tokens. Shading denotes the interquartile range across components.}
  \label{fig:ica-fitting-sample-size}
\end{figure}

We examine how the number of fitting activations affects the non-Gaussian structure recovered by ICA. We fit all 768 components of GPT-2 Small's residual stream after block~11 using nested subsets of 1K, 5K, 10K, 100K, and 1M token activations drawn from the same Pile-10k pool. All fits use the same initialization seed and 50 symmetric FastICA updates. At initialization and after each update, we measure component excess kurtosis on the fitting subset and on a common set of one million additional token activations from disjoint Pile-10k documents. For held-out evaluation, we apply each fit's learned centering, whitening, and ICA rotation directly.

Figure~\ref{fig:ica-fitting-sample-size} shows that FastICA can overfit when the number of fitting tokens is small relative to the number of components. After 50 updates, the 1K-token fit reaches a median excess kurtosis of 514.94 on its fitting tokens and only 0.53 on held-out tokens. As the fitting sample grows, held-out kurtosis rises and the gap narrows, with held-out medians of 11.04 and 11.83 for the 100K- and 1M-token fits, respectively. These results suggest that ICA needs enough fitting activations to recover directions whose non-Gaussianity persists on unseen documents. One possible explanation is that larger samples cover more concepts and provide more examples of each concept across contexts, helping ICA recover recurring activation patterns.

\FloatBarrier

\subsection{Convergence Diagnostics: Non-Gaussianity vs.\ LIM}
\label{app:ica-convergence}

The results above point to two complementary notions of convergence. \emph{Statistical convergence} asks whether the recovered component population has already captured most of the non-Gaussian structure available to FastICA. \emph{Component convergence} asks whether individual directions have stabilized to nearly the same fixed points under additional optimization or different random initializations.

FastICA commonly tracks component convergence using the per-component limit statistic
\begin{equation}
  \mathrm{LIM}_j^{(t)}
  =
  \left|
  \left|
  (w_j^{(t)})^\top w_j^{(t-1)}
  \right|
  -1
  \right|,
\end{equation}
which measures how much a direction changes between successive updates, up to the arbitrary sign of an ICA component. Small LIM therefore indicates that an individual direction is approaching a fixed point. Maximum LIM is a particularly strict summary when thousands of components are optimized jointly because it is determined by the least stable direction.

\begin{figure}[ht!]
  \centering
  \includegraphics[width=\textwidth]{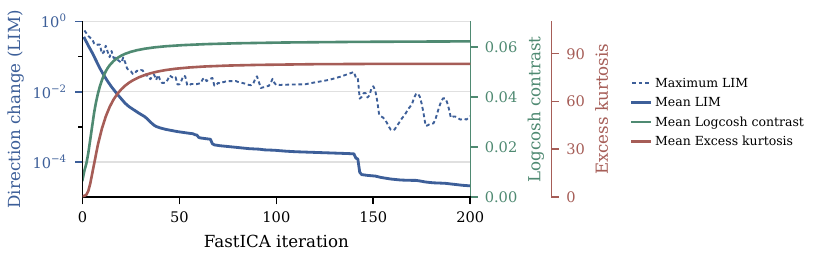}
  \caption{\textbf{Population non-Gaussianity stabilizes before individual directions.} We track all 4,096 components from each of Qwen~3.5~9B Base layers 7, 15, 23, and 31 over a 200-iteration FastICA fit, for 16,384 directions in total. Diagnostics are recomputed on the fitting activations at each checkpoint. Mean and maximum LIM measure changes between successive direction estimates, while mean logcosh contrast and mean excess kurtosis measure population non-Gaussianity. The non-Gaussianity statistics largely plateau by roughly 50 iterations, while directional changes continue to decrease over substantially longer fitting.}
  \label{fig:lim-diagnostic}
\end{figure}

Figure~\ref{fig:lim-diagnostic} compares directional change with population non-Gaussianity over a 200-iteration run. We track all 4,096 components from each of Qwen~3.5~9B Base layers 7, 15, 23, and 31, for 16,384 directions in total. At every checkpoint, LIM, logcosh contrast, and excess kurtosis are recomputed on the fitting activations.

Mean LIM decreases steadily throughout fitting, and maximum LIM remains noisier because it is driven by a small number of slowly stabilizing components. In contrast, mean logcosh contrast and mean excess kurtosis increase rapidly and largely plateau by roughly 50 iterations. FastICA directly optimizes logcosh rather than kurtosis, yet the two statistics follow closely aligned trajectories.

Together with the seed-stability analysis, this result explains the 50-iteration operating point used throughout the main experiments. By roughly 50 iterations, the population-level non-Gaussianity recovered by FastICA has largely saturated, while the full fitting procedure remains fast enough to apply across every layer of the evaluated models. Individual directions continue to become more reproducible beyond this point: on GPT-2 Small layer~6, the cross-seed median matched cosine rises from roughly $0.83$ at 50 iterations to roughly $0.91$ at 200 iterations. Longer runs therefore primarily improve the stability of exact component identities after most of the population-level statistical structure has already been recovered.

This distinction is also useful for interpreting the non-Gaussianity metric used in the main text. FastICA optimizes the logcosh contrast, while we report excess kurtosis because it has a simple Gaussian reference at zero and a wider numerical range that makes differences across directions, layers, and methods easier to visualize. Their closely aligned fitting trajectories show that both statistics track the same broad emergence and saturation of non-Gaussian structure in these fits.

\section{Labeling and Interpreting Non-Gaussian Directions}
\label{app:interpretability}

A useful feature representation should support interpretations that remain meaningful beyond a few selected examples.
We therefore treat component labeling as an iterative hypothesis-testing process rather than a one-shot description of the largest activations.
For each component, we combine evidence from both signed tails, high-scoring corpus contexts, Logit Lens and R-Lens vocabulary readouts, and controlled positive and negative examples.
We use the same component profiles for human annotation and for scalable automated interpretability evaluation.
This section describes the orientation rule, manual and automated evaluation protocols, robustness checks, and an error analysis of the automated metric.

\subsection{Selecting the Semantic Tail by Skewness}
\label{app:tail-selection}

FastICA identifies each component only up to sign: if $w_j$ is a valid component, then $-w_j$ represents the same ICA solution.
This matters for interpretation because a coherent semantic pattern is often concentrated in one tail, while the opposite tail is comparatively diffuse.
Inspecting only the arbitrary positive side can therefore miss the more informative tail.

We select the semantic tail using the skewness of the component scores.
Let $y_{ij}$ denote the score of activation $i$ on component $j$, consistent with the notation in Section~\ref{sec:ica-discovery}.
We compute
\begin{equation}
  \gamma_j =
  \frac{
    \frac{1}{n}\sum_{i=1}^{n}(y_{ij}-\bar y_j)^3
  }{
    \left[
      \frac{1}{n}\sum_{i=1}^{n}(y_{ij}-\bar y_j)^2
    \right]^{3/2}
  },
  \qquad
  \bar y_j=\frac{1}{n}\sum_{i=1}^{n}y_{ij}.
  \label{eq:component-skewness}
\end{equation}

In our main experiments, this statistic is computed on the one-million-token activation capture used to fit and profile each component.
Positive skew selects the positive tail and negative skew selects the negative tail.
If $\gamma_j=0$, we break the tie using the fraction of squared score mass on the positive side,
\begin{equation}
  q_j^{+}
  =
  \frac{
    \sum_{i=1}^{n}y_{ij}^{2}\mathbf{1}[y_{ij}>0]
  }{
    \sum_{i=1}^{n}y_{ij}^{2}
  }.
\end{equation}

We record the selected tail with a sign $s_j\in\{-1,+1\}$,
\begin{equation}
  s_j =
  \begin{cases}
    +1, & \gamma_j>0,\\
    -1, & \gamma_j<0,\\
    +1, & \gamma_j=0\ \text{and}\ q_j^{+}\geq 0.5,\\
    -1, & \text{otherwise},
  \end{cases}
\end{equation}
and define the semantic-tail score
\begin{equation}
  \widetilde{y}_{ij}=s_j y_{ij}.
\end{equation}

The fitted ICA basis itself is unchanged.
We store $s_j$ and apply it to component scores and the corresponding reading and writing directions when needed.
Thus, the selected semantic tail is always the positive side of $\widetilde{y}_{ij}$.

For procedures that require a one-sided feature, such as automated interpretability and targeted probe perturbation, we use
\begin{equation}
  y_{ij}^{+}=\max(0,\widetilde{y}_{ij}).
\end{equation}
Sparse probing instead uses the signed semantic-tail score $\widetilde{y}_{ij}$.
Example retrieval and manual labeling use high-scoring examples from the selected semantic tail.
Sign-invariant quantities such as excess kurtosis and reconstruction are unaffected by this choice.

\subsection{Manual Audit Protocol and Full Annotations}
\label{app:manual-audit}
\label{app:annotator-comparison}

We first use manual annotation to examine the structure of individual components in detail and to test whether their interpretations remain stable beyond the strongest corpus examples.
We uniformly sample 50 components from the 50-iteration fits without replacement over all available layer--component pairs from each model, giving 150 randomly selected components in total.
The sample therefore covers components across model depth.
Two interpretability experts independently annotate the same components without seeing each other's labels or confidence scores.

For each component, we first inspect both signed tails, the highest-scoring corpus occurrences, and the corresponding Logit Lens and R-Lens vocabulary readouts.
We use this evidence to form an initial concise hypothesis about the component.
We then test the hypothesis with controlled examples.
LLMs assist in generating positive examples that should express the proposed concept and negative or counterfactual examples that preserve nearby surface structure while removing it.
We evaluate the component on these examples, inspect mismatches, refine the hypothesis when needed, and repeat the process until the interpretation is stable.
This iterative procedure helps separate broad semantic or contextual patterns from narrower lexical, positional, or syntactic explanations that can fit the original top-activating examples.

Each annotator assigns a confidence score from 0 to 5 that summarizes the full annotation process, including how easily a coherent hypothesis emerges, how much refinement is required during contextual and counterfactual testing, and how consistently the final label explains the observed behavior.
Both annotators use the same rubric: 0 unknown, 1 weak, 2 partial, 3 plausible, 4 stable, and 5 very clear.
Scores 4--5 are treated as high-confidence labels.

\begin{figure}[ht!]
  \centering
  \includegraphics[width=\textwidth]{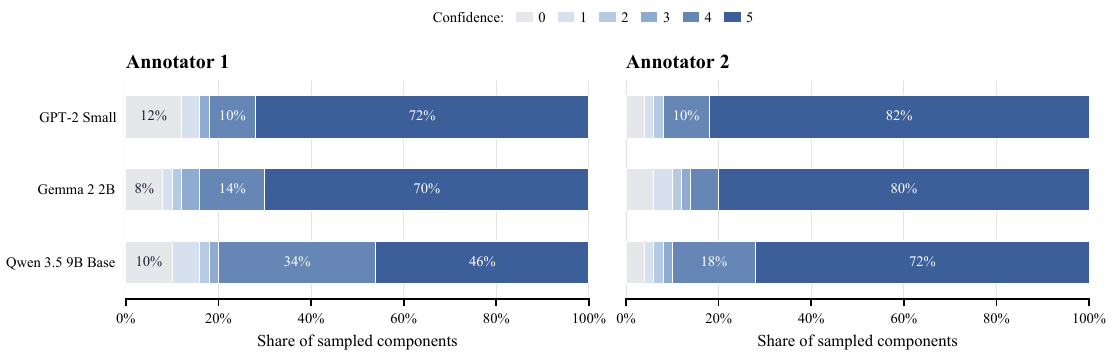}
  \caption{\textbf{Both annotators assign high confidence to most randomly sampled ICA components.} Bars show the fraction of components at each confidence score from 0 to 5, separately for each annotator and model. Each annotator labels the same 50 components per model.}
  \label{fig:annotation-confidence}
\end{figure}

Figure~\ref{fig:annotation-confidence} shows the confidence distributions for both annotators.
Table~\ref{tab:annotator-agreement} reports their high-confidence rates and agreement on confidence scores.
The mean high-confidence rate across annotators is 85.7\%, and both annotators rate 123 of the 150 components (82.0\%) as high confidence.

\begin{table}[ht!]
  \centering
  \small
  \setlength{\tabcolsep}{4pt}
  \caption{\textbf{Confidence ratings and inter-annotator agreement across three models.} Scores of 4 or 5 on the 0--5 scale count as high confidence. The 95\% confidence interval for overall $\kappa$ is computed from 20,000 paired bootstrap resamples stratified by model.}
  \label{tab:annotator-agreement}
  \begin{tabularx}{\linewidth}{@{}>{\raggedright\arraybackslash}Xcccc@{}}
    \toprule
    Metric & GPT-2 Small & Gemma 2 2B & \shortstack{Qwen 3.5\\9B Base} & Overall \\
    \midrule
    Components annotated & 50 & 50 & 50 & 150 \\
    High confidence, A1 & 41/50 (82\%) & 42/50 (84\%) & 40/50 (80\%) & 123/150 (82.0\%) \\
    High confidence, A2 & 46/50 (92\%) & 43/50 (86\%) & 45/50 (90\%) & 134/150 (89.3\%) \\
    \midrule
    Exact score agreement & 76\% & 82\% & 60\% & 72.7\% \\
    Scores within one point & 90\% & 100\% & 90\% & 93.3\% \\
    Agreement on high-confidence status & 90\% & 98\% & 90\% & 92.7\% \\
    Quadratically weighted Cohen's $\kappa$ & 0.667 & 0.958 & 0.702 &
    \shortstack{\textbf{0.772}\\{[0.635, 0.874]}} \\
    \bottomrule
  \end{tabularx}
\end{table}

\paragraph{Differences in labels and confidence.}
The annotators sometimes describe related patterns at different levels of specificity.
For Qwen~3.5~9B Base L4/C565, A1 labels the component as a week-or-month pattern, while A2 focuses on cross-lingual expressions for ``week.''
For Gemma~2~2B L14/C1937, A1 emphasizes conditional compilation, while A2 highlights line breaks before preprocessor directives.
Related labels can also receive different confidence scores.
For GPT-2 Small L7/C467, A1 assigns a score of 1 to a repeated-math-question interpretation, while A2 assigns a score of 4 to a quantitative-query interpretation.

Tables~\ref{tab:manual-annotations-gpt2}--\ref{tab:manual-annotations-qwen3.5-9b} report both annotators' labels and confidence scores for all 150 sampled components. The annotations include lexical, syntactic, semantic, positional, code-related, and longer-context patterns.

\begin{table}[ht!]
  \centering
  \caption{Manual component annotations for GPT-2 Small. A spaced slash separates A1 and A2 labels. Confidence scores are ordered A1/A2. Shared labels are shown once. Some labels are shortened for readability.}
  \label{tab:manual-annotations-gpt2}
  \scriptsize
  \renewcommand{\arraystretch}{0.86}
\setlength{\tabcolsep}{3pt}
\begin{tabularx}{\linewidth}{@{}rr>{\raggedright\arraybackslash}Xrrr@{}}
  \toprule
  Layer & Component & Annotation 1/2 & Skewness & \shortstack[r]{Excess\\kurtosis} & \shortstack[r]{Confidence\\1/2} \\
  \midrule
  0 & C421 & Key & 4.00 & 77.50 & 5/5 \\
  0 & C459 & Full & 3.55 & 70.78 & 5/5 \\
  0 & C524 & Personal & 1.35 & 16.82 & 5/5 \\
  0 & C642 & Light & -0.07 & 11.17 & 5/5 \\
  1 & C191 & 5 & 5.28 & 53.48 & 5/5 \\
  1 & C284 & Accented Latin characters & -5.95 & 93.59 & 4/5 \\
  2 & C46 & First token after \textless{}\textbar{}endoftext\textbar{}\textgreater{}\enspace/\allowbreak\enspace Document starts and title prediction & 11.69 & 199.3 & 5/5 \\
  3 & C61 & '\enspace/\allowbreak\enspace Contraction and possessive apostrophes & -7.94 & 106.1 & 5/4 \\
  3 & C393 & Contact relative clause/omitted ``that''\enspace/\allowbreak\enspace Subordinate relative clauses & 3.12 & 30.44 & 5/5 \\
  3 & C409 & Christ\enspace/\allowbreak\enspace Christianity and biblical terms & -2.37 & 30.23 & 5/5 \\
  3 & C455 & Item & -1.74 & 15.11 & 5/5 \\
  3 & C650 & Include\enspace/\allowbreak\enspace Inclusion and feature terms & 0.69 & 8.41 & 5/5 \\
  3 & C728 & Unknown\enspace/\allowbreak\enspace Western names (output) & 0.68 & 2.69 & 0/1 \\
  4 & C32 & Be & 8.19 & 100.9 & 5/5 \\
  4 & C103 & Enumerations/lists\enspace/\allowbreak\enspace List separation, continuation, or closure & -5.53 & 58.31 & 5/5 \\
  5 & C66 & He & 6.33 & 80.76 & 5/5 \\
  5 & C312 & Unknown\enspace/\allowbreak\enspace Entity completion and reflexive predicates & 2.25 & 12.87 & 0/4 \\
  5 & C333 & Studies & -2.83 & 32.28 & 5/5 \\
  5 & C482 & Post-\{ newline\enspace/\allowbreak\enspace Code/JSON block openings and closing braces & -1.69 & 12.21 & 5/5 \\
  5 & C598 & Pre\enspace/\allowbreak\enspace Pre- prefix and subword completion & -1.02 & 10.85 & 5/5 \\
  5 & C751 & Unknown\enspace/\allowbreak\enspace African related output & -0.01 & 0.91 & 0/0 \\
  6 & C159 & ://\enspace/\allowbreak\enspace URL scheme and hostname prediction & -7.00 & 156.4 & 5/5 \\
  6 & C174 & Unknown\enspace/\allowbreak\enspace Initial subwords and word completion & 2.83 & 16.06 & 0/5 \\
  7 & C215 & Combining, Interaction\enspace/\allowbreak\enspace Combination, reciprocity, and between/with & 2.86 & 22.25 & 5/5 \\
  7 & C222 & Been\enspace/\allowbreak\enspace Perfect aspect before temporal adverbs & 3.26 & 34.15 & 5/5 \\
  7 & C347 & Name & 2.42 & 26.41 & 5/5 \\
  7 & C380 & Post-colon newline\enspace/\allowbreak\enspace Post-colon newline in code or shell blocks & -2.56 & 25.65 & 5/5 \\
  7 & C395 & Sentence-ending period\enspace/\allowbreak\enspace Final period before a capitalized sentence start & -1.90 & 13.48 & 5/5 \\
  7 & C422 & Repeated variable in later polynomial term\enspace/\allowbreak\enspace Polynomial variables before exponentiation & -1.92 & 21.70 & 5/5 \\
  7 & C452 & Unknown\enspace/\allowbreak\enspace Entity subjects before definitional or relational verbs & -1.44 & 7.63 & 0/4 \\
  7 & C467 & Repeated math questions\enspace/\allowbreak\enspace Math queries about quantities & 1.42 & 10.62 & 1/4 \\
  7 & C625 & //\enspace/\allowbreak\enspace Code comment continuation (//) & -1.65 & 23.46 & 5/5 \\
  7 & C628 & Prefix non-\enspace/\allowbreak\enspace Non- prefix in compound completions & 1.42 & 17.89 & 5/5 \\
  7 & C757 & Post-URL for details or name a few\enspace/\allowbreak\enspace Reference and documentation pointers & 0.07 & 0.75 & 1/2 \\
  8 & C152 & Post-quote reporting verb\enspace/\allowbreak\enspace Speech attribution and reporting verbs after quotes & -4.34 & 52.80 & 5/5 \\
  8 & C396 & Programming/API specification language\enspace/\allowbreak\enspace API specifications and code constraints & -1.84 & 7.33 & 5/5 \\
  8 & C743 & Mr\enspace/\allowbreak\enspace Mr. before a surname & 0.24 & 2.92 & 5/5 \\
  9 & C56 & Do/does\enspace/\allowbreak\enspace Do/does in negation and object continuations & -7.09 & 122.1 & 5/5 \\
  9 & C119 & Japanese language\enspace/\allowbreak\enspace Japanese language continuation & 4.38 & 49.88 & 4/5 \\
  9 & C464 & Group/list\enspace/\allowbreak\enspace Aggregate nouns for group or list membership & -1.56 & 11.03 & 5/5 \\
  9 & C619 & Way/approach descriptor\enspace/\allowbreak\enspace Solution-evaluating adjectives before method nouns & 1.00 & 7.15 & 4/5 \\
  9 & C730 & Who & 0.25 & 2.21 & 5/5 \\
  10 & C294 & Prefix comp- & -2.21 & 17.61 & 5/5 \\
  10 & C307 & Opening parentheses in code\enspace/\allowbreak\enspace Function or constructor argument lists & 2.43 & 20.42 & 5/5 \\
  10 & C445 & Expecting a temporal-duration phrase\enspace/\allowbreak\enspace Temporal duration after perfect aspect & -1.67 & 13.63 & 4/4 \\
  10 & C446 & Opening quotation mark & 1.79 & 15.26 & 5/5 \\
  10 & C493 & Occupational title or official role\enspace/\allowbreak\enspace Job title and official role & -1.42 & 10.41 & 5/5 \\
  11 & C243 & Expecting letter D but need more conditions\enspace/\allowbreak\enspace Character-level d/D completion & 2.12 & 18.73 & 3/5 \\
  11 & C593 & Official agency or department\enspace/\allowbreak\enspace Government modifiers for institutions or officials & 0.83 & 5.68 & 4/5 \\
  11 & C752 & Unknown\enspace/\allowbreak\enspace noise feature & -0.02 & 0.62 & 0/0 \\
  \bottomrule
\end{tabularx}
\end{table}

\begin{table}[ht!]
  \centering
  \caption{Manual component annotations for Gemma~2~2B. A spaced slash separates A1 and A2 labels. Confidence scores are ordered A1/A2. Shared labels are shown once. Some labels are shortened for readability.}
  \label{tab:manual-annotations-gemma-2-2b}
  \scriptsize
  \renewcommand{\arraystretch}{0.86}
\setlength{\tabcolsep}{3pt}
\begin{tabularx}{\linewidth}{@{}rr>{\raggedright\arraybackslash}Xrrr@{}}
  \toprule
  Layer & Component & Annotation 1/2 & Skewness & \shortstack[r]{Excess\\kurtosis} & \shortstack[r]{Confidence\\1/2} \\
  \midrule
  0 & C603 & Pressure & 24.53 & 1317 & 5/5 \\
  0 & C761 & 3rd-person singular present verb & -6.38 & 68.75 & 5/5 \\
  0 & C1423 & Australian & -3.14 & 109.3 & 5/5 \\
  1 & C1290 & Words related to diagnosis\enspace/\allowbreak\enspace diagnosis & 5.88 & 220.2 & 5/5 \\
  1 & C2180 & Super & -0.20 & 14.64 & 5/5 \\
  2 & C1994 & Long underscore separator\enspace/\allowbreak\enspace Document divider and structural marker & -0.05 & 8.70 & 3/3 \\
  3 & C2037 & Substring tab or cat\enspace/\allowbreak\enspace Cat/tab subwords & -0.06 & 5.18 & 5/4 \\
  4 & C271 & Why & 9.26 & 251.6 & 5/5 \\
  4 & C494 & Programming errors/exceptions & 5.50 & 100.7 & 5/5 \\
  4 & C1229 & Letter Q & -2.36 & 29.08 & 5/5 \\
  4 & C1372 & Check & -3.42 & 74.02 & 5/5 \\
  4 & C2205 & Fast & -0.08 & 3.62 & 5/5 \\
  5 & C1839 & U.S. Supreme Court case citation\enspace/\allowbreak\enspace U.S. Supreme Court reporter citations & 3.46 & 181.3 & 4/5 \\
  6 & C1503 & Definitional statement\enspace/\allowbreak\enspace Formal definitions and concept explanations & -2.15 & 22.26 & 5/5 \\
  6 & C2082 & Unknown\enspace/\allowbreak\enspace noise feature & 0.06 & 2.45 & 0/0 \\
  7 & C39 & Post-Soviet context\enspace/\allowbreak\enspace Russian/Soviet entities and translated text & 6.79 & 84.98 & 5/5 \\
  7 & C878 & Code syntax class="\enspace/\allowbreak\enspace Code class=" & -5.40 & 143.5 & 5/5 \\
  7 & C1016 & Because of, a cause/reason expression\enspace/\allowbreak\enspace Reason expression & 2.61 & 28.14 & 5/5 \\
  8 & C138 & Previously introduced unfamiliar proper noun\enspace/\allowbreak\enspace Initial subwords of multi-token words & -3.96 & 29.91 & 4/5 \\
  10 & C834 & Website/E-mail domain name\enspace/\allowbreak\enspace URL and email address domain & -2.38 & 27.79 & 5/5 \\
  10 & C2197 & Unknown\enspace/\allowbreak\enspace Noise & 0.00 & 0.84 & 0/0 \\
  12 & C790 & Month-day in date\enspace/\allowbreak\enspace Final day-of-month digit in day-of-week contexts & -2.67 & 36.86 & 4/5 \\
  12 & C1615 & PHP web-development\enspace/\allowbreak\enspace PHP code and web development & 0.99 & 9.38 & 5/5 \\
  12 & C2205 & Digit in document-initial Federal Reporter citation\enspace/\allowbreak\enspace Digit in initial Federal Reporter (F.2d/F.3d) citation & -0.01 & 1.02 & 2/2 \\
  12 & C2262 & Long divider\enspace/\allowbreak\enspace Repeated ASCII punctuation and dividers & 0.06 & 0.92 & 1/1 \\
  13 & C210 & Toy/example code\enspace/\allowbreak\enspace Code snippets, placeholders, and framework templates & -3.97 & 45.84 & 5/5 \\
  13 & C528 & So & 3.43 & 53.65 & 5/5 \\
  14 & C211 & Second token position\enspace/\allowbreak\enspace Second Token after \textless{}bos\textgreater{} & 4.76 & 59.64 & 4/5 \\
  14 & C1109 & Data schema/field type definition\enspace/\allowbreak\enspace Database schema, DDL, and ORM field types & 1.94 & 24.06 & 5/5 \\
  14 & C1846 & Unknown\enspace/\allowbreak\enspace Archaic long-s vocabulary prediction & -0.54 & 3.13 & 0/1 \\
  14 & C1937 & Conditional compilation in C/C++ and other languages\enspace/\allowbreak\enspace Line break before C/C++ preprocessor directives & -0.40 & 5.83 & 3/4 \\
  15 & C1900 & Present & -0.89 & 8.93 & 5/5 \\
  18 & C1304 & Nuclear medicine imaging\enspace/\allowbreak\enspace Radiopharmaceuticals and radioisotope imaging & 3.27 & 60.50 & 5/5 \\
  18 & C1345 & Sentence-final additive/concessive adverb pattern, anyway\enspace/\allowbreak\enspace Clause-final adverbs and discourse particles & 2.27 & 22.12 & 5/5 \\
  20 & C86 & C\enspace/\allowbreak\enspace C-prefix & 12.08 & 302.8 & 5/5 \\
  20 & C736 & Banking/lending\enspace/\allowbreak\enspace Commercial banking, credit, and lending & 3.52 & 43.39 & 5/5 \\
  20 & C949 & Decorative aesthetics\enspace/\allowbreak\enspace Decorative crafts and interiors & -2.87 & 32.85 & 5/5 \\
  20 & C1154 & 1960s--1970s American culture/history\enspace/\allowbreak\enspace 1960s--1970s counterculture and the Vietnam War & 2.77 & 41.62 & 5/5 \\
  21 & C1514 & )\{ & -2.85 & 58.89 & 5/5 \\
  22 & C618 & After & 4.80 & 82.84 & 5/5 \\
  22 & C1380 & Digit in numeric identifier\enspace/\allowbreak\enspace Multi-digit sequence continuation & 2.11 & 22.71 & 4/5 \\
  22 & C1458 & Greek letters in scientific notation\enspace/\allowbreak\enspace Scientific/biomedical prefixes before Greek letters & 2.41 & 37.61 & 5/5 \\
  22 & C1852 & Am-, prefer capitalized, American\enspace/\allowbreak\enspace Capitalized ``Am-'' and ``American'' & -2.12 & 48.57 & 5/5 \\
  24 & C507 & 7 & 4.02 & 64.19 & 5/5 \\
  24 & C776 & Adverse health effects/medical complications\enspace/\allowbreak\enspace Clinical diseases, pathologies, and symptoms & 2.33 & 23.78 & 5/5 \\
  24 & C1827 & Expecting ``include''\enspace/\allowbreak\enspace Inclusion and exemplification before ``include'' or ``including'' & -1.06 & 7.87 & 5/5 \\
  24 & C2018 & JavaScript DOM property/method access\enspace/\allowbreak\enspace JavaScript DOM property & 0.07 & 5.61 & 4/4 \\
  25 & C1291 & Anticipated future change\enspace/\allowbreak\enspace Future projection, modals, and temporal or conditional cues & 1.49 & 10.05 & 4/5 \\
  25 & C1701 & Salesforce development/customization\enspace/\allowbreak\enspace Salesforce, Apex, and Aura development & 1.68 & 31.30 & 5/5 \\
  25 & C1952 & Unknown\enspace/\allowbreak\enspace Noise & -0.27 & 3.52 & 0/0 \\
  \bottomrule
\end{tabularx}
\end{table}

\begin{table}[ht!]
  \centering
  \caption{Manual component annotations for Qwen~3.5~9B Base. A spaced slash separates A1 and A2 labels. Confidence scores are ordered A1/A2. Shared labels are shown once. Some labels are shortened for readability.}
  \label{tab:manual-annotations-qwen3.5-9b}
  \scriptsize
  \renewcommand{\arraystretch}{0.86}
\setlength{\tabcolsep}{3pt}
\begin{tabularx}{\linewidth}{@{}rr>{\raggedright\arraybackslash}Xrrr@{}}
  \toprule
  Layer & Component & Annotation 1/2 & Skewness & \shortstack[r]{Excess\\kurtosis} & \shortstack[r]{Confidence\\1/2} \\
  \midrule
  0 & C609 & Store & -29.79 & 1594 & 5/5 \\
  0 & C3472 & Promise & -1.53 & 60.16 & 5/5 \\
  1 & C3036 & Do something/nothing\enspace/\allowbreak\enspace Do/does/did/done with nothing/anything/something & -0.79 & 41.92 & 5/5 \\
  1 & C3298 & spokesman & 0.50 & 53.94 & 5/5 \\
  2 & C231 & User & -18.70 & 666.4 & 5/5 \\
  2 & C2327 & -{}-{}-\enspace/\allowbreak\enspace Document-initial YAML delimiter (-{}-{}-) & 4.17 & 96.64 & 5/5 \\
  2 & C3057 & -y, -ie nickname\enspace/\allowbreak\enspace Nickname suffixes -ie, -y, and -nie & 0.30 & 25.22 & 1/2 \\
  4 & C565 & Week/month\enspace/\allowbreak\enspace Cross-lingual ``week'' & -9.49 & 276.6 & 4/5 \\
  4 & C1156 & Introduction\enspace/\allowbreak\enspace Introduction heading in academic or technical documents & -11.61 & 618.1 & 5/5 \\
  4 & C2219 & Measurement/unit-conversion math question\enspace/\allowbreak\enspace Elementary math and unit conversions & -2.51 & 34.17 & 4/4 \\
  5 & C3133 & Diabetes & -0.67 & 16.44 & 5/5 \\
  6 & C269 & Figurative expression/metaphor\enspace/\allowbreak\enspace Metaphors, similes, and idioms & -6.82 & 127.0 & 5/5 \\
  6 & C609 & Data & -6.85 & 146.0 & 5/5 \\
  6 & C992 & Notice/note as information\enspace/\allowbreak\enspace Note & 7.44 & 229.8 & 5/5 \\
  6 & C2199 & Counterfactual/hypothetical conditional\enspace/\allowbreak\enspace Subjunctive and counterfactual clauses & 4.88 & 160.7 & 5/5 \\
  7 & C2650 & /*\enspace/\allowbreak\enspace File-initial C-style block comment delimiter & -6.81 & 376.6 & 5/5 \\
  7 & C3045 & Helicopter/rotorcraft\enspace/\allowbreak\enspace Helicopters, rotorcraft aviation, and rotor aerodynamics & 2.31 & 77.51 & 5/5 \\
  7 & C3826 & Unknown\enspace/\allowbreak\enspace Noise & -0.02 & 1.74 & 0/0 \\
  8 & C935 & Repeated structured code/machine output\enspace/\allowbreak\enspace Repeated machine logs and code boilerplate & 3.07 & 27.33 & 4/4 \\
  9 & C2175 & Children's/preschool educational content\enspace/\allowbreak\enspace Children's picture books and preschool literature & -2.68 & 58.13 & 4/5 \\
  10 & C170 & Footnote/reference marker & 20.95 & 1463 & 5/5 \\
  10 & C4002 & Unknown\enspace/\allowbreak\enspace Noise & 0.03 & 1.38 & 0/0 \\
  13 & C637 & Online forum/discussion thread\enspace/\allowbreak\enspace Forum post boundaries and paragraph breaks & -4.52 & 87.23 & 4/5 \\
  13 & C889 & Commented-out code\enspace/\allowbreak\enspace Commented-out or disabled code & 5.52 & 175.4 & 4/5 \\
  15 & C1201 & Enumerating numbers\enspace/\allowbreak\enspace Intermediate elements in ordered sequences or scales & 2.25 & 27.98 & 1/4 \\
  16 & C61 & Wikipedia category/category-page\enspace/\allowbreak\enspace Wikipedia/MediaWiki ``Category:'' tag & 10.41 & 230.4 & 4/5 \\
  16 & C244 & Repeated previously mentioned named entity\enspace/\allowbreak\enspace Initial subwords of specialized terms or proper nouns & -3.25 & 22.96 & 4/4 \\
  16 & C1365 & Closing tag \textless{}/\enspace/\allowbreak\enspace HTML/XML closing tag delimiter (``\textless{}/'') & -2.80 & 45.07 & 4/5 \\
  16 & C3363 & Unknown\enspace/\allowbreak\enspace Preparation, upcoming events, or awaiting (write-dominant) & -0.00 & 0.85 & 0/1 \\
  17 & C530 & Country/state/region name\enspace/\allowbreak\enspace Geopolitical jurisdiction & -3.07 & 31.39 & 2/5 \\
  18 & C3329 & Mc- surname prefix & 1.32 & 34.35 & 5/5 \\
  19 & C146 & Italian/Italy-related context\enspace/\allowbreak\enspace Italian language, entities, and Italian-influenced English & 8.09 & 133.3 & 4/5 \\
  19 & C1472 & HTML DOM access/manipulation\enspace/\allowbreak\enspace HTML DOM manipulation and web markup & -2.44 & 23.88 & 4/4 \\
  19 & C2291 & Unknown\enspace/\allowbreak\enspace Uncapitalized proper nouns and brand names & 1.85 & 17.55 & 0/4 \\
  19 & C3174 & Unknown\enspace/\allowbreak\enspace First-person emotional reflection and confession & 0.08 & 4.05 & 0/3 \\
  20 & C352 & Both & -10.24 & 283.2 & 5/5 \\
  20 & C1145 & Related to letter F or expecting F\enspace/\allowbreak\enspace Next-token prediction of initial f/F & 2.76 & 23.25 & 1/4 \\
  20 & C2259 & Document/source attribution\enspace/\allowbreak\enspace Attribution to documents or written statements & -2.72 & 42.52 & 4/5 \\
  21 & C704 & Microsoft platform technical documentation\enspace/\allowbreak\enspace Microsoft ecosystem and Graph API documentation & -3.94 & 46.69 & 4/5 \\
  22 & C2808 & Urination/bladder function\enspace/\allowbreak\enspace Urination, bladder physiology, and incontinence & -2.18 & 28.84 & 4/5 \\
  22 & C3385 & Right-wing/conservative political context\enspace/\allowbreak\enspace Political conservatism and right-wing politics & -0.21 & 7.65 & 4/4 \\
  23 & C1583 & Limit\enspace/\allowbreak\enspace Limitation, restriction, and constraint & -4.81 & 111.5 & 5/5 \\
  25 & C602 & Lawsuit/civil litigation & -5.22 & 96.00 & 5/5 \\
  25 & C1260 & Maryland/Baltimore\enspace/\allowbreak\enspace Maryland and Baltimore geographic or legal context & 7.15 & 231.3 & 4/5 \\
  25 & C2695 & Tell & 3.42 & 76.78 & 5/5 \\
  25 & C3939 & Causal graphical-model reasoning\enspace/\allowbreak\enspace Causality, causal discovery, and DAGs & -0.70 & 18.01 & 3/4 \\
  26 & C68 & Prefix Sh- & -12.89 & 330.7 & 5/5 \\
  26 & C1210 & Deep-learning framework/model\enspace/\allowbreak\enspace Deep learning frameworks and neural network modeling & 4.81 & 95.20 & 5/5 \\
  28 & C3025 & Prefix Var-\enspace/\allowbreak\enspace Subword and suffix completion for var- & -1.94 & 33.56 & 4/5 \\
  30 & C2683 & Expecting ``see''\enspace/\allowbreak\enspace Prediction and promotion of ``see'' or ``seen'' & 1.54 & 13.75 & 5/5 \\
  \bottomrule
\end{tabularx}
\end{table}

\clearpage

\paragraph{Tracing component behavior across context.}
\label{app:component-traces}

Sentence-level traces provide one way to refine an initial component hypothesis.
A component whose largest activation occurs at one token may still represent a broader phrase or contextual state, so we inspect its behavior across complete sequences when useful.

\begin{quote}
  \small
  Sentence 1. She arrived at the library to study for her exam.

  Sentence 2. Maya stopped at the bank before the trip, waiting in line to deposit a check and withdraw enough cash for the weekend.
\end{quote}

\begin{figure}[ht!]
  \centering
  \includegraphics[width=\linewidth]{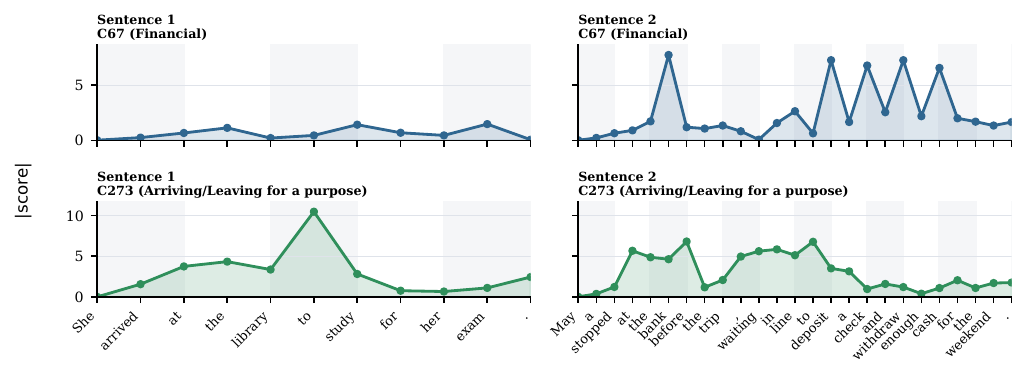}
  \caption{\textbf{Sentence-level traces help refine component interpretations.} Raw absolute scores are shown for GPT-2 Small layer~6 components C67 and C273. C67 follows financial context across several related tokens, while C273 responds to a broader arrival- and purpose-related construction.}
  \label{fig:gpt2-layer6-sentence-trace}
\end{figure}

Figure~\ref{fig:gpt2-layer6-sentence-trace} illustrates two cases where the trace provides more information than the top-activating token alone.
C67 is labeled as a financial component.
Its score stays near zero in Sentence 1, which contains no financial context, but rises around \emph{bank} in Sentence 2 and remains active near \emph{deposit}, \emph{check}, \emph{withdraw}, and \emph{cash}.
The trace therefore supports a broader financial-context interpretation rather than a detector for the word \emph{bank}.

C273 shows a more construction-level pattern.
In Sentence 1, it rises across ``arrived at the library to study'' and peaks near the \emph{to study} region, where motion, destination, and purpose come together.
It also responds to related movement and purpose structure in Sentence 2.
This behavior supports an arrival- or purpose-related interpretation rather than a label tied to a single token.

\begin{table}[h]
  \centering
  \caption{\textbf{Examples from the counterfactual stage of the manual annotation workflow.} Five Qwen~3.5~9B Base components are tested on two positive and two negative examples each.}
  \label{tab:counterfactual-label-tests}
  \scriptsize
\begin{tabular}{@{}llp{0.55\textwidth}rr@{}}
  \toprule
  Component & Expected & Prompt & Score & Rank \\
  \midrule
  \multicolumn{5}{@{}l}{\textbf{L1/C3036: Do something / nothing}} \\
  L1/C3036 & activate & The committee did \underline{nothing} to prevent the closure. & $-40.724$ & 1 \\
  L1/C3036 & activate & Please do \underline{something} to fix the broken lock. & $-24.828$ & 3 \\
  L1/C3036 & not activate & The committee \underline{knew} about the closure. & $+1.588$ & 581 \\
  L1/C3036 & not activate & Please use a \underline{wrench} to fix the broken lock. & $+1.123$ & 1678 \\
  \midrule
  \multicolumn{5}{@{}l}{\textbf{L9/C2175: Children’s / preschool educational content}} \\
  L9/C2175 & activate & Designed for preschoolers learning colors, this is a picture \underline{book}. & $-34.483$ & 1 \\
  L9/C2175 & activate & A cheerful alphabet song teaches young children their \underline{ABC}. & $-21.774$ & 2 \\
  L9/C2175 & not activate & Designed for graduate engineers studying thermodynamics, this is a reference \underline{book}. & $-0.334$ & 2991 \\
  L9/C2175 & not activate & The television affiliate announced a change to its \underline{ABC} schedule. & $+0.491$ & 2913 \\
  \midrule
  \multicolumn{5}{@{}l}{\textbf{L19/C146: Italian / Italy-related context}} \\
  L19/C146 & activate & Rome, Milan, and Naples are cities in \underline{Italy}. & $+10.361$ & 4 \\
  L19/C146 & activate & She ordered pasta and espresso while traveling through \underline{Italy}. & $+10.687$ & 3 \\
  L19/C146 & not activate & Beijing, Shanghai, and Guangzhou are cities in \underline{China}. & $+1.107$ & 898 \\
  L19/C146 & not activate & She ordered sushi and tea while traveling through \underline{Japan}. & $-0.132$ & 3582 \\
  \midrule
  \multicolumn{5}{@{}l}{\textbf{L25/C1260: Maryland / Baltimore}} \\
  L25/C1260 & activate & The court issued its ruling under the laws of \underline{Maryland}. & $+25.503$ & 1 \\
  L25/C1260 & activate & After leaving Annapolis, the train continued toward \underline{Baltimore}. & $+23.690$ & 1 \\
  L25/C1260 & not activate & The court issued its ruling under the laws of \underline{California}. & $+0.252$ & 2904 \\
  L25/C1260 & not activate & After leaving Providence, the train continued toward \underline{Boston}. & $-0.729$ & 1285 \\
  \midrule
  \multicolumn{5}{@{}l}{\textbf{L26/C1210: Deep-learning framework/model}} \\
  L26/C1210 & activate & Built with PyTorch, the classifier is a deep-learning \underline{model}. & $+17.809$ & 2 \\
  L26/C1210 & activate & Using TensorFlow, we trained a convolutional neural \underline{network}. & $+24.952$ & 1 \\
  L26/C1210 & not activate & Built from cardboard, the display is an architectural \underline{model}. & $+1.733$ & 259 \\
  L26/C1210 & not activate & Using fiber optics, the offices share a communication \underline{network}. & $+1.896$ & 94 \\
  \bottomrule
\end{tabular}
\end{table}

\paragraph{Example counterfactual checks.}
\label{app:counterfactual-label-tests}

Table~\ref{tab:counterfactual-label-tests} illustrates the counterfactual stage of the same annotation workflow for five Qwen~3.5~9B Base components drawn from the random audit.
For each component, we construct two examples expected to activate the proposed interpretation and two controls expected not to activate it.
The controls retain relevant surface form or sentence structure where practical while changing the hypothesized concept.

We report each component's signed ICA score and its one-based rank by absolute score among all 4,096 components at the underlined target token.
All ten predicted positive examples rank within the top four components at their target positions, whereas the ten controls rank between 94 and 3,582.
These examples illustrate how controlled tests help refine and verify working labels.

\begin{figure}[ht!]
  \centering
  \includegraphics[width=\textwidth]{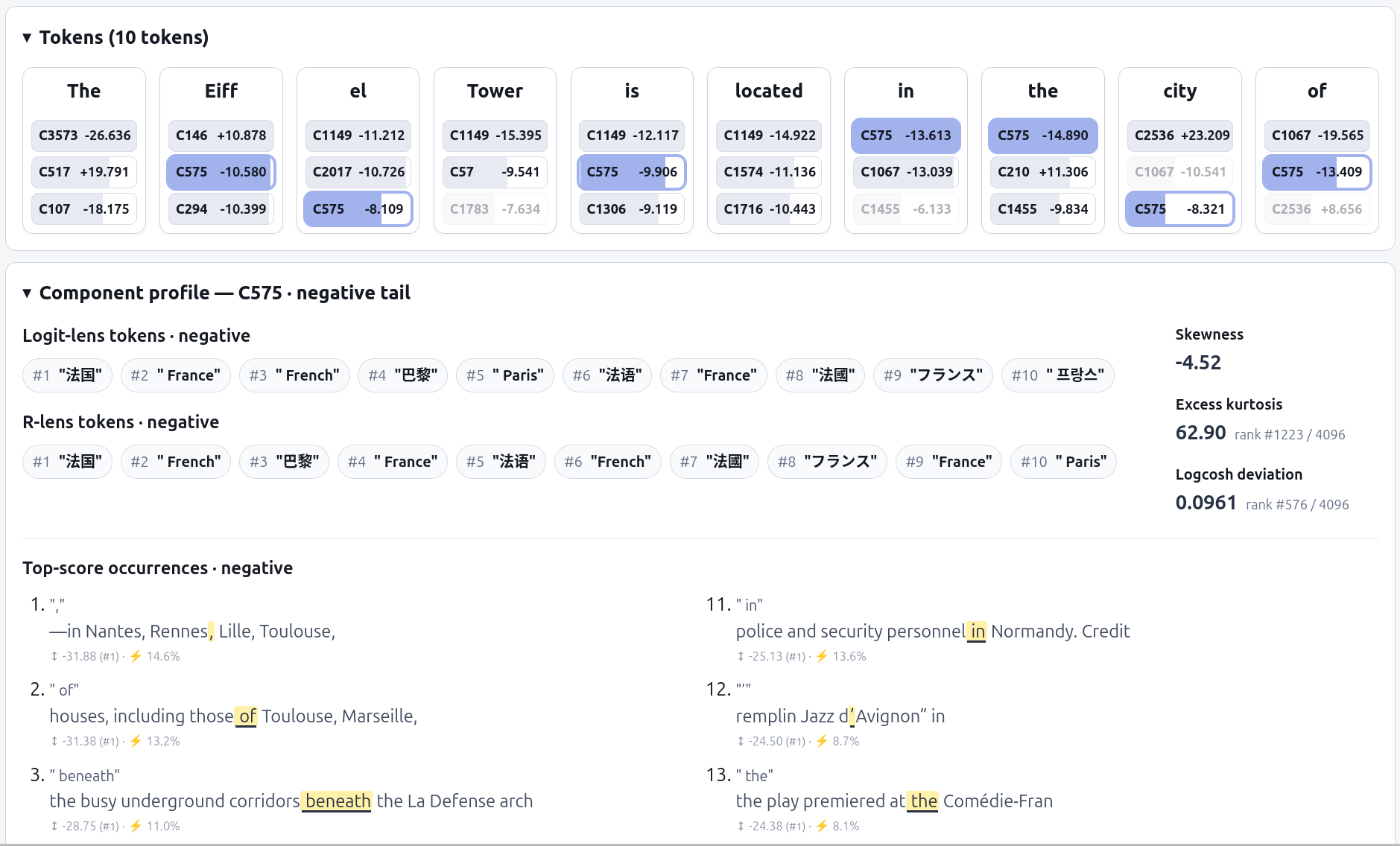}
  \caption{\textbf{ICA Lens interactive annotation interface.}
    For the prompt ``The Eiffel Tower is located in the city of'', the upper panel shows the highest-scoring ICA components at each token for Qwen~3.5~9B Base, layer~27.
    Selecting component C575's negative semantic tail opens its profile below, including Logit Lens and R-Lens token readouts, skewness, excess kurtosis, logcosh contrast, and representative high-scoring corpus occurrences.
  The interface is used to support systematic component inspection and annotation.}
  \label{fig:human-inspection-interface}
\end{figure}

\textbf{Interactive annotation interface.}
To support systematic human inspection, we built a dedicated ICA Lens interface for annotating ICA components.
For any supplied prompt, the interface shows the highest-scoring components at each token and links them directly to the evidence used for interpretation.
Selecting a component and semantic tail opens its profile, including Logit Lens and R-Lens token readouts, distribution statistics, and representative high-scoring corpus occurrences.
This interface allows annotators to move between model behavior, component scores, vocabulary readouts, and corpus evidence within a single workflow rather than inspecting each source separately.
Figure~\ref{fig:human-inspection-interface} shows the interface for Qwen~3.5~9B Base at layer~27.

\FloatBarrier

\newpage
\subsection{Automated Interpretability Protocol}
\label{app:auto-interpretability}

We next use explanation--prediction evaluation to extend interpretability measurement to a larger sample of components.
An explainer receives high-activating text fragments and their component scores and produces a concise description of the activation pattern.
A predictor then receives that explanation and predicts component activations on held-out fragments.
We report the Pearson correlation between predicted and observed activations, so a high score requires the explanation to generalize beyond the examples used to construct it.

For the historical Pythia~70M reproduction, we follow the released explanation--prediction protocol of \citet{huben2024sparse}.
GPT-4.1-mini is used for both explanation generation and held-out activation prediction.
For Qwen~3.5~9B Base, \texttt{thinkingmachines/Inkling} generates explanations and \texttt{Qwen3.8-27B} predicts held-out activations.
The arbitrary-sign and skewness-selected ICA conditions use exactly the same fitted axes and differ only in which signed tail is presented to the evaluator.

We compare usable ICA and SAE features under matched sampling rules.
For Pythia, we follow the released procedure by sampling 150 feature IDs uniformly from the full SAE dictionary and skipping dead features.
For Qwen, we first restrict each representation to features that are active on the shared fragment pool and then uniformly sample 150 features from the eligible set.
Eligibility requires at least 20 top-activation records and 20 nonzero random records for both ICA and SAE features.

For the fitting-trajectory analysis in Figure~\ref{fig:nonGaussian-interp}, we follow the same 100 Qwen components from each of layers 7, 15, 23, and 31 across FastICA checkpoints.
At every checkpoint, we retrieve that checkpoint's own top 25 occurrences from distinct documents.
Five fragments are used to generate an explanation and the remaining 20 are used for activation prediction.
This checkpoint-specific retrieval lets the evaluation follow the component as its direction changes during fitting.

\subsection{ICA--SAE Reproduction and Robustness Checks}
\label{app:ica-sae-robustness}

We next test whether the interpretability comparison between ICA and SAE features is stable across changes in ICA fitting budget, training corpus, and evaluator pipeline. These experiments extend the historical Pythia~70M reproduction and separate the effect of ICA sign orientation from other choices in the evaluation setup.

\paragraph{The sign effect is stable under a much smaller ICA fitting budget.}

\begin{figure}[ht]
  \centering
  \includegraphics[width=\textwidth]{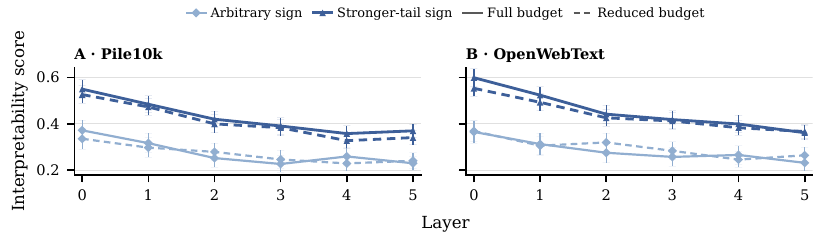}
  \caption{\textbf{The sign correction is stable under a substantially reduced ICA fitting budget.} Solid curves use the released-code fitting budget of 2,098,176 activations and up to 200 iterations, while dashed curves use 524,288 activations and 20 iterations. The two budgets produce similar interpretability curves on both Pile-10k and OpenWebText, while the gap between arbitrary-sign and skewness-selected ICA remains large.}
  \label{fig:ica-reduced-budget}
\end{figure}

The released Pythia reproduction uses 2,098,176 fitting activations and up to 200 FastICA iterations.
We compare it with a substantially smaller configuration using 524,288 activations and 20 iterations.
Figure~\ref{fig:ica-reduced-budget} shows that the two fitting budgets produce closely aligned interpretability curves on both Pile-10k and OpenWebText.
Reducing both the sample count and iteration budget changes the aggregate skewness-selected ICA score by only about 0.02 under either corpus, while the difference between arbitrary-sign and skewness-selected ICA remains much larger.

\paragraph{The same pattern persists across corpora and evaluator pipelines.}

\begin{figure}[ht]
  \centering
  \includegraphics[width=\textwidth]{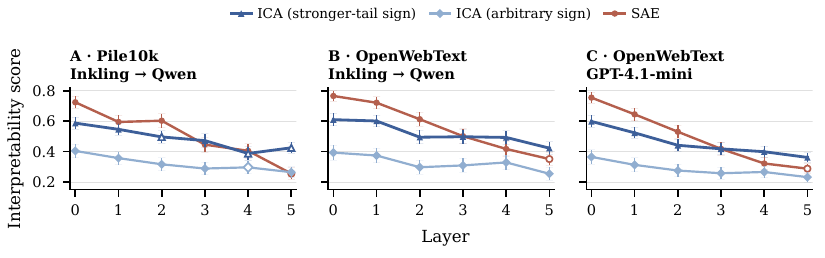}
  \caption{\textbf{The skewness-selected ICA comparison is stable across training corpora and evaluator pipelines.} (A--B) We hold the Inkling--Qwen3.8-27B evaluator fixed while changing the ICA and SAE training corpus from Pile-10k to OpenWebText. (B--C) We hold OpenWebText training fixed while changing the evaluator to GPT-4.1-mini. Absolute scores vary across settings, but resolving ICA's arbitrary sign produces the same qualitative improvement throughout.}
  \label{fig:ica-evaluator-robustness}
\end{figure}

Figure~\ref{fig:ica-evaluator-robustness} provides a triangular robustness comparison.
Panels A and B keep the Inkling--Qwen3.8-27B evaluator fixed while changing the ICA and SAE training corpus from Pile-10k to OpenWebText.
Panels B and C keep OpenWebText training fixed while changing the evaluator to GPT-4.1-mini.
Absolute interpretability scores move across these settings, but resolving the arbitrary ICA sign consistently recovers a large part of the measured gap to the SAE baseline and yields comparable scores at several layers.

\FloatBarrier

\subsection{Error Analysis of the Relationship Between Non-Gaussianity and Automated Interpretability}
\label{app:interpretability-error-analysis}

\begin{figure}[ht!]
  \centering
  \includegraphics[width=\textwidth]{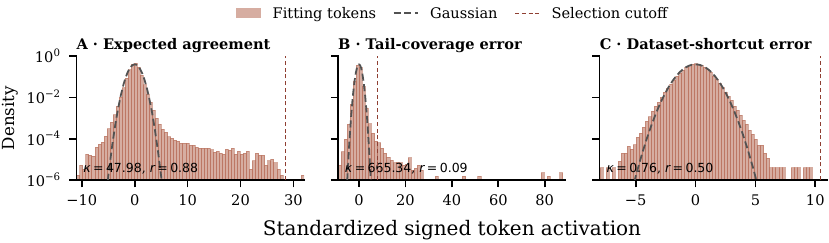}
  \caption{\textbf{Three regimes of agreement between non-Gaussianity and automated interpretability.} Panels show expected agreement (A), a sparse-tail coverage error (B), and a dataset-shortcut error (C).}
  \label{fig:auto-interp-error-modes}
\end{figure}

Automated interpretability and non-Gaussianity are positively associated overall, but individual components can deviate substantially from this trend.
To understand these outliers, we inspect components with high excess kurtosis but low automated interpretability, as well as components with low excess kurtosis but relatively high automated interpretability.
We identify two recurring failure modes in the automated evaluation: insufficient coverage of sparse semantic tails and dataset-specific shortcuts that make some components easier to predict.

Figure~\ref{fig:auto-interp-error-modes} compares three representative regimes at FastICA iteration 50.
Each histogram shows the signed component-score distribution over the one-million-token Pile-10k fitting population.
The dashed Gaussian curve provides a reference distribution.
The red selection cutoff marks the maximum token activation of the weakest among the 25 distinct-document OpenWebText fragments admitted to explanation and held-out evaluation, expressed on the standardized scale of the fitting population.
This view shows whether the evaluation examples actually reach the component tail that drives its non-Gaussianity.

\paragraph{Well-covered semantic tails produce agreement.}

Figure~\ref{fig:auto-interp-error-modes}A shows Qwen L7/C338.
Its selected examples consistently concern courts and judicial proceedings, and the evaluation fragments reach the same high-activation tail visible in the fitting distribution.
The resulting explanation therefore predicts held-out activations well, giving both high non-Gaussianity and high automated interpretability.

\paragraph{Sparse semantic tails can be missed by the evaluation corpus.}

Figure~\ref{fig:auto-interp-error-modes}B shows Qwen L7/C2035.
The component has fitting-population excess kurtosis $\kappa=665.34$ and reaches a standardized score of 88.03.
In a systematic 100,000-token diagnostic sample, all seven scores above 20 occur in C/C++ preprocessor contexts, primarily header guards and conditional compilation.
The component therefore contains an extremely sparse but coherent tail.
The 50,000-fragment OpenWebText candidate pool does not reach comparable examples: the weakest selected top fragment has a standardized score of only 8.02, and the explanation examples instead contain more moderate activations from unrelated contexts.
The automated evaluator consequently sees little of the rare structure responsible for the component's high kurtosis.

\paragraph{Dataset regularities can make lower-kurtosis components predictable.}

Figure~\ref{fig:auto-interp-error-modes}C shows Qwen L15/C1503.
The explainer initially proposes a lexical pattern involving words such as \emph{Human}, \emph{What}, and \emph{How}, which appear at position 0 in the selected examples and receive raw activations of 11.82, 11.75, and 11.74.
Controlled examples reveal a broader positional pattern: the same words activate near zero when placed later in the sequence, while unrelated position-0 tokens such as \emph{The}, \emph{I}, and \emph{will} still receive activations of 8.99, 8.68, and 7.59.
The evaluation examples therefore contain a regularity that makes the component predictable even though the initial lexical explanation does not capture the underlying positional structure.

These cases suggest that automated interpretability can underestimate the relationship between non-Gaussianity and human interpretability. Sparse but coherent semantic tails may be missed by the sampled evaluation contexts, while dataset-specific regularities can inflate scores for less meaningful patterns. The observed correlation with automated interpretability may therefore be a conservative estimate of the underlying relationship.

These failure modes also suggest a concrete direction for improved evaluation.
A coverage-aware protocol could guarantee examples from the selected semantic tail across multiple activation quantiles and distinct documents, evaluate the resulting explanation on a second corpus to reduce dataset-specific shortcuts, and validate a subset with blinded human judgments.

\FloatBarrier

\section{Reading and Intervening with Non-Gaussian Directions}
\label{app:using-directions}

We can decompose new activations into labeled component mixtures to inspect how information is represented, or modify selected coordinates to test their effect on model behavior.
We demonstrate these two uses through contextual decomposition, embedding analysis, and generation-time interventions.

\subsection{Reading Component Mixtures}
\label{app:component-mixtures}

\subsubsection{Contextual Decomposition of Polysemy}
\label{app:polysemy}

Polysemous words provide a useful setting for studying contextual decomposition because the surface token stays fixed while the surrounding evidence changes.
We use a paragraph containing four occurrences of \emph{bank}: two referring to a financial institution, denoted F1 and F2, and two referring to a river edge, R1 and R2.

\begin{quote}
  \small
  Maya stopped at the bank [F1] before the trip, waiting in line to deposit a check and withdraw enough cash for the weekend.
  The teller asked about her travel plans, and Maya said she was driving north to visit the old village by the river.
  By sunset she was sitting on the grassy bank [R1] with her shoes beside her, watching the water curl around stones and reeds.
  It amused her that the same word had followed her all day: first a bank [F2] with counters, accounts, and vaults, then a bank [R2] of earth holding the river in place.
\end{quote}

\begin{figure}[ht!]
  \centering
  \includegraphics[width=\linewidth]{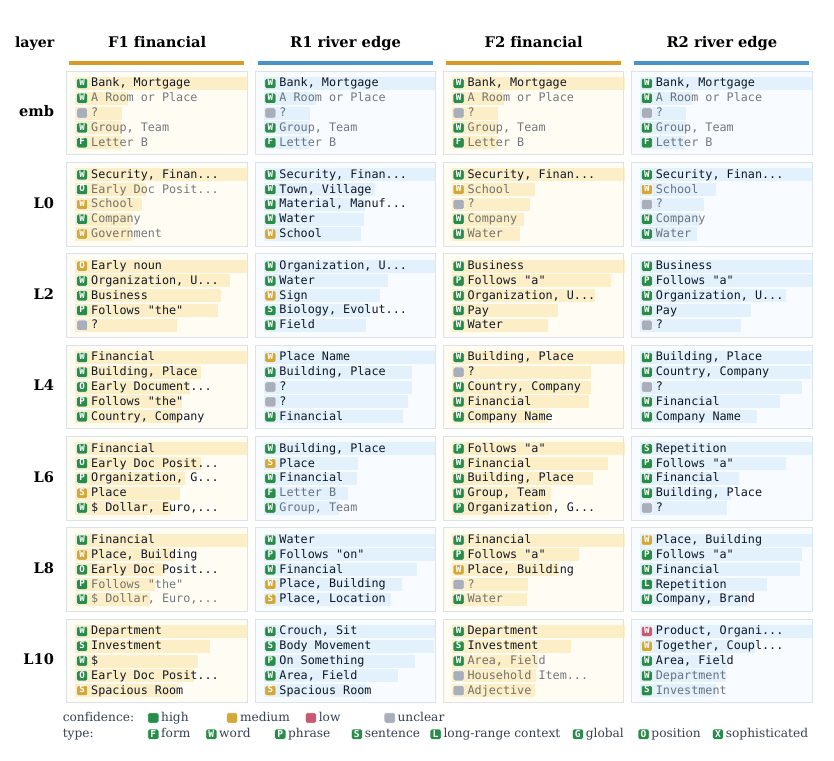}
  \caption{\textbf{Contextual decomposition of a polysemous word in GPT-2 Small.} We probe two financial uses of \emph{bank} (F1, F2) and two river-edge uses (R1, R2). Each cell lists the five largest-absolute-score ICA components at the target token. The embedding representation is identical across occurrences, while later layers separate into different mixtures of lexical, syntactic, semantic, positional, and longer-context components.}
  \label{fig:gpt2-bank-case-study}
\end{figure}

Figure~\ref{fig:gpt2-bank-case-study} decomposes the four \emph{bank} tokens into their largest-absolute-score ICA components at the embedding layer and several GPT-2 Small residual-stream layers.
Each cell lists the top five components at that token position, with bar length showing relative score magnitude within the cell and the annotation badge showing component type and confidence.

The embedding row provides a simple control.
All four columns are identical because the input embedding depends only on the token identity \emph{bank}.
Its strongest components include directions labeled \emph{Bank, Mortgage}, \emph{A Room or Place}, \emph{Group, Team}, and \emph{Letter B}.
Before contextual processing begins, the four occurrences therefore share the same lexical mixture.

The residual-stream representations separate as context accumulates.
F1 appears after the prefix ``Maya stopped at the,'' which already supports a financial reading before later phrases such as ``deposit a check'' and ``withdraw enough cash'' become available.
Later layers contain components related to finance, organizations, companies, investment, and departments.
R1 instead follows a context involving a village and river, with the immediate prefix ``sitting on the grassy.''
Its mixture shifts toward directions related to water, place, posture, movement, and outdoor space.

F2 returns to the financial sense but appears after both meanings of \emph{bank} have already been introduced.
Its representation combines financial components such as \emph{Business}, \emph{Organization}, \emph{Pay}, and \emph{Investment} with local phrase-level and discourse-related components.
The same token can therefore combine semantic content with syntax and discourse state rather than mapping to a single sense feature.

R2 also highlights the importance of autoregressive context.
A human reader identifies the phrase as a river edge after seeing ``bank of earth holding the river,'' but at the target token the model has only seen the prefix ending in ``then a.''
That prefix still contains the earlier financial use.
Finance-related components at R2 are therefore consistent with the information available to the model at that position.

Overall, ICA does not reduce polysemy to one financial-versus-river axis.
The same surface token is represented by a changing mixture of lexical, syntactic, semantic, positional, and longer-context directions, and that mixture evolves with the information available at each token position and layer.

\subsubsection{Lexical Structure in Token Embeddings}
\label{app:embedding-structure}

The previous analyses study contextual residual-stream representations.
We now remove context and apply the same component view to token embeddings, where each token has a fixed representation before any transformer layer is applied.

As an additional test on a newer model outside the main three-model evaluation, we fit ICA to the complete token-embedding matrix of Qwen~3.5~2B Base.
No vocabulary sampling is used.
This setting asks whether the same non-Gaussian decomposition can expose readable lexical structure directly in the embedding space.

Classical word-embedding analyses often summarize lexical relations through vector offsets, such as \emph{king} $-$ \emph{man} $+$ \emph{woman} $\approx$ \emph{queen}.
ICA provides a complementary view by decomposing each token into multiple readable directions rather than summarizing the relation with one displacement.

\begin{figure}[ht!]
  \centering
  \includegraphics[width=\linewidth]{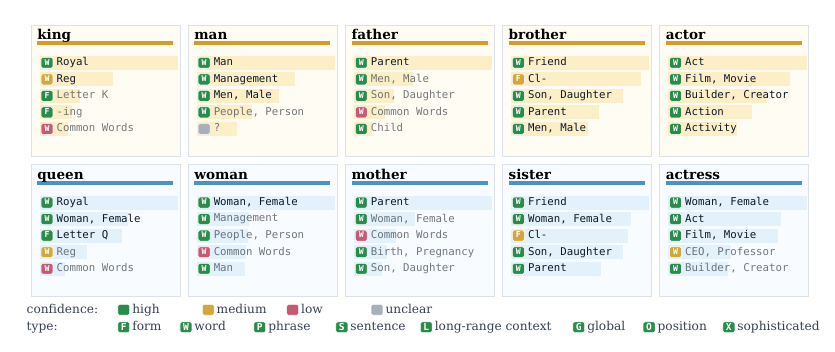}
  \caption{\textbf{Embedding-layer ICA decomposition of familiar lexical relations in Qwen~3.5~2B Base.} Each box lists the five largest-absolute-score ICA components for one token embedding. Related tokens share interpretable directions while differing along additional semantic, gender-associated, morphological, and surface-form components.}
  \label{fig:embedding-analogy-ica-examples}
\end{figure}

Figure~\ref{fig:embedding-analogy-ica-examples} shows several familiar lexical pairs.
For \emph{king} and \emph{queen}, both embeddings contain a strong \emph{Royal} component, while the remaining mixture separates them through directions such as \emph{Woman, Female}, letter identity, morphology, and common-token structure.
The pair is therefore represented as shared structure together with additional token-specific factors.

The same pattern appears across other lexical families.
\emph{Father} and \emph{mother} share a strong \emph{Parent} component and related family directions, while gender-associated and birth-related components distinguish them.
\emph{Brother} and \emph{sister} share several social and family-relation components.
\emph{Actor} and \emph{actress} share acting-, film-, and creation-related directions, while \emph{actress} additionally activates a strong \emph{Woman, Female} component.

The component view therefore exposes both shared structure and asymmetry.
Familiar lexical relations are represented by mixtures of semantic, morphological, surface-form, and other token-specific directions rather than by a single isolated axis.
This extends the same reading framework used for contextual residual streams to the model's static lexical representation.

\FloatBarrier

\subsection{Illustrative Generation Interventions}
\label{app:generation-interventions}

The analyses above use labeled ICA coordinates to read model activations.
We next use the same coordinates for generation-time intervention.
We consider two operations: additive steering along an ICA writing direction and coordinate clamping in the fitted ICA basis.

\paragraph{Selecting intervention components.}

The GPT-2 example uses layer~7 component C317, selected from the labeled components for its weekend-activity pattern.
For Gemma and Qwen, we select language-related components from 1,000 seeded English--target-language sentence pairs from ManyThings.
For each sentence, we extract the residual activation at the final non-padding text token, excluding EOS, and transform it into ICA coordinates.

For component $c$, we compute
\begin{equation}
  \Delta(c)
  =
  \overline{y}_{\mathrm{target}}(c)
  -
  \overline{y}_{\mathrm{English}}(c),
\end{equation}
where $\overline{y}_{\mathrm{target}}(c)$ and $\overline{y}_{\mathrm{English}}(c)$ are the mean ICA scores over the 1,000 target-language and English sentences.
We rank components by $|\Delta(c)|$, so the selected component is the one that most strongly separates the two languages within the searched layers.
The sign of $\Delta(c)$ indicates which side of the ICA axis is associated with the target language.
Searching Gemma layers 18--22 selects layer~20 component C105 for Chinese versus English, while searching Qwen layers 23--27 selects layer~26 component C9 for French versus English.

\paragraph{Additive steering.}

Let $d_j$ be the residual-space writing direction of component $j$, and let $\alpha$ denote the desired change in its ICA coordinate.
Without a cap, additive steering is
\begin{equation}
  x' = x + \alpha d_j.
\end{equation}
The intervention is applied at the current generation position: the final prompt token during prefill and, when continued steering is enabled, each newly generated token during decoding.
For GPT-2 C317, we use $\alpha=-20$ at the final prompt token and no further steering.
For Gemma C105, we use $\alpha=+35$ at the final prompt token.
For Qwen C9, we use an initial $\alpha=-40$ followed by $\alpha=-20$ during decoding.

\paragraph{Coordinate clamping.}

Additive steering moves the activation along a selected writing direction, while clamping directly removes that coordinate from the current representation.
Given a residual activation, we transform it into ICA coordinates, set the selected component score to zero, map the modified coordinates back to residual space, and restore the original residual norm.
For the Gemma and Qwen language examples, this clamp is applied throughout generation.

\begin{table}[ht]
  \centering
  \caption{\textbf{Generation interventions with individual ICA components.} GPT-2 C317 is a labeled weekend-activity component. Gemma C105 and Qwen C9 are selected from 1,000 English--target-language pairs as the components with the largest absolute difference in mean ICA score. Additive steering changes the selected coordinate along its writing direction, while clamping sets that coordinate to zero. We use greedy decoding.}
  \label{tab:language-steering-showcase}
  \scriptsize
  \begin{CJK*}{UTF8}{gbsn}
{\setlength{\tabcolsep}{2.5pt}
\begin{tabular}{@{}p{50pt}p{110pt}p{103pt}p{114pt}@{}}
  \toprule
  Model & Prompt & Baseline continuation & Intervened continuation \\
  \midrule
  \raggedright GPT-2 Small\newline L7/C317 & \raggedright \textit{Prompt:} ``My favorite thing to do on the weekend is to'' & \raggedright \textit{Base:} ``go to a local bar and have a drink.'' & \raggedright \textit{Steer:} ``play with the kids.'' \tabularnewline
  \midrule
  \raggedright Gemma~2~2B\newline L20/C105 & \raggedright \textit{EN:} ``Hangzhou is a historic city known for''\newline \textit{ZH:} ``杭州是一座历史悠久的城市，它以'' & \raggedright \textit{EN:} ``its beautiful scenery and rich culture.''\newline \textit{ZH:} ``其悠久的历史和丰富的文化遗产而闻名。'' & \raggedright \textit{EN to ZH, steer:} ``其历史悠久的文化和美丽的风景。''\newline \textit{ZH to EN, ablate:} ``its rich history and culture as well as its beautiful scenery.'' \tabularnewline
  \midrule
  \raggedright Qwen~3.5~9B Base\newline L26/C9 & \raggedright \textit{EN:} ``Deep learning is a machine learning technique that''\newline \textit{FR:} ``L’apprentissage profond est une technique d’apprentissage automatique qui'' & \raggedright \textit{EN:} ``uses artificial neural networks to learn from data.''\newline \textit{FR:} ``utilise des réseaux de neurones artificiels pour apprendre à partir de données.'' & \raggedright \textit{EN to FR, steer:} ``utilise des réseaux de neurones artificiels pour apprendre à partir de données.''\newline \textit{FR to EN, ablate:} ``uses neural networks to learn from data.'' \tabularnewline
  \bottomrule
\end{tabular}}
  \end{CJK*}
\end{table}

The GPT-2 example generates up to 10 new tokens, while the Gemma and Qwen examples generate up to 24.
In these examples, steering moves generation toward the behavior associated with the selected component, while clamping the language-related coordinate shifts target-language prompts toward English.

\FloatBarrier

\section{Full Quantitative Evaluation of Non-Gaussian Directions}
\label{app:quantitative-evaluation}

The main text summarizes reconstruction, sparse probing, and targeted probe perturbation (TPP) after averaging across datasets and evaluated layers.
Here we give the full evaluation protocol and disaggregate the results by dataset, model depth, and metric.
The breakdowns show that the main trends are not driven by a small number of evaluation settings.

\subsection{Evaluation Protocol and Aggregation}
\label{app:evaluation-protocol}

Table~\ref{tab:representation-coordinate-evaluation-setup} summarizes the layer coverage for the three evaluations.
Reconstruction is evaluated at every model layer, while sparse probing and TPP use two representative layers per model.
All downstream evaluation datasets are separate from the Pile-10k activations used to fit ICA.

At each top-$k$ budget, we first average the evaluated layers within each dataset.
The main-text curves then average equally across datasets, with shaded regions showing $\pm1$ standard deviation across datasets.
The figures below report the corresponding dataset-level results, and reconstruction is additionally disaggregated across model depth.

\begin{table}[ht!]
  \centering
  \small
  \setlength{\tabcolsep}{5pt}
  \begin{tabular}{@{}lccc@{}}
    \toprule
    \textbf{Evaluation}
    & \textbf{GPT-2 Small}
    & \textbf{Gemma~2~2B}
    & \textbf{Qwen~3.5~9B Base} \\
    \midrule
    Reconstruction & All layers & All layers & All layers \\
    Sparse probing & 6, 10 & 12, 20 & 12, 20 \\
    TPP & 6, 10 & 12, 20 & 12, 20 \\
    \bottomrule
  \end{tabular}
  \caption{\textbf{Layer coverage for representation-coordinate evaluations.} Reconstruction uses all layers, while sparse probing and TPP use two representative layers per model.}
  \label{tab:representation-coordinate-evaluation-setup}
\end{table}

\paragraph{Reconstruction datasets.}

We evaluate reconstruction on six datasets spanning news, general text, code, multilingual text, and dialogue:
\begin{itemize}
  \item \nolinkurl{fancyzhx/ag_news};
  \item \nolinkurl{Salesforce/wikitext} (\texttt{wikitext-103-raw-v1});
  \item \nolinkurl{codeparrot/github-code};
  \item \nolinkurl{wikimedia/wikipedia} (\texttt{20231101.es}; Spanish Wikipedia);
  \item \nolinkurl{wikimedia/wikipedia} (\texttt{20231101.zh}; Chinese Wikipedia); and
  \item \nolinkurl{HuggingFaceH4/ultrachat_200k}.
\end{itemize}

\paragraph{Sparse-probing datasets.}

We use eight SAEBench sparse-probing datasets:
\begin{itemize}
  \item \nolinkurl{LabHC/bias_in_bios_class_set1};
  \item \nolinkurl{LabHC/bias_in_bios_class_set2};
  \item \nolinkurl{LabHC/bias_in_bios_class_set3};
  \item \nolinkurl{canrager/amazon_reviews_mcauley_1and5};
  \item \nolinkurl{canrager/amazon_reviews_mcauley_1and5_sentiment};
  \item \nolinkurl{codeparrot/github-code};
  \item \nolinkurl{fancyzhx/ag_news}; and
  \item \nolinkurl{Helsinki-NLP/europarl}.
\end{itemize}

\paragraph{TPP datasets.}

We evaluate TPP on \nolinkurl{LabHC/bias_in_bios_class_set1} and \nolinkurl{canrager/amazon_reviews_mcauley_1and5}.

\subsection{Sparse Reconstruction}
\label{app:reconstruction}

Sparse reconstruction asks how much of a held-out residual activation can be recovered from only a small number of representation coordinates.
For a linear dictionary such as ICA, PCA, or a random basis, let $y_j$ be the signed score for coordinate $j$ and $d_j$ its residual-space writing direction.
We rank coordinates by their contribution magnitude,
\begin{equation}
  \mathrm{contrib}_j = |y_j|\,\lVert d_j\rVert_2,
\end{equation}
retain the original signed scores for the top-$k$ coordinates, set the rest to zero, and reconstruct
\begin{equation}
  \hat{x}
  =
  \mu
  +
  \sum_{j\in S_k}y_j d_j,
\end{equation}
where $\mu$ is the fitted center and $S_k$ is the selected coordinate set.
We also evaluate the complete-basis reconstruction at $k=d$.

For SAE baselines, we first encode with the published checkpoint using its native nonlinearity or TopK mechanism.
We then rank the resulting non-negative latent codes by the same contribution criterion, retain the top-$k$ active codes, and decode with the published SAE decoder.
Thus, top-$k$ truncation is applied after the SAE's native encoder sparsity rather than replacing it.

The reconstruction plots use the mean number of nonzero retained features per token on the horizontal axis.
For the dense linear bases, this equals the requested $k$.
For an SAE, the effective feature count can be smaller than the requested budget when the native encoder already produces fewer nonzero latents.
The endpoint stars show complete-basis reconstruction for ICA, PCA, and random directions, and the native sparse encode--decode reconstruction for each SAE.

\subsubsection{Across Datasets}
\label{app:reconstruction-datasets}

\begin{figure}[ht]
  \centering
  \includegraphics[width=\textwidth]{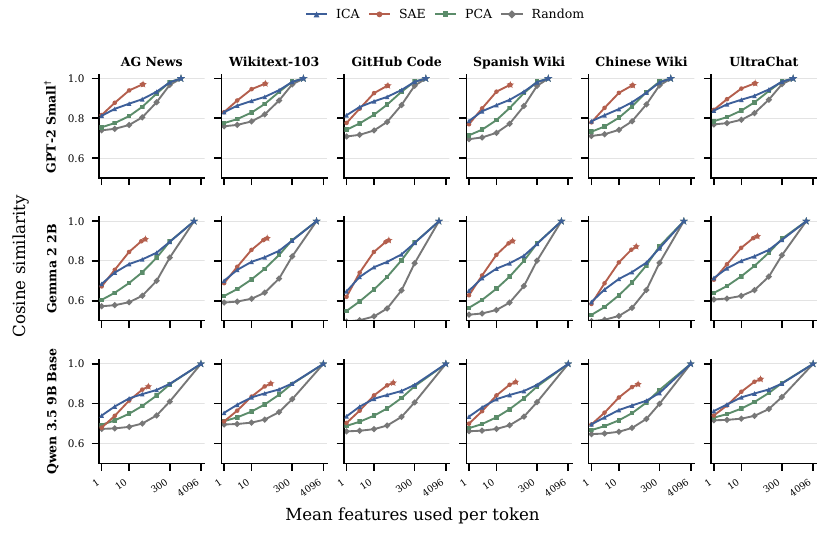}
  \caption{\textbf{Sparse reconstruction across evaluation domains.} Cosine similarity is shown separately for every model--dataset pair. ICA consistently outperforms PCA and random bases at sparse budgets and remains close to the public SAE baselines across six evaluation domains.}
  \label{fig:reconstruction-datasets}
\end{figure}

Figure~\ref{fig:reconstruction-datasets} reports cosine-similarity reconstruction separately for every model--dataset pair.
The same broad ordering appears across news, general text, code, multilingual Wikipedia, and dialogue.
At small coordinate budgets, ICA consistently reconstructs better than PCA and random bases across all three models.
The public SAE baselines are often strongest at sparse budgets, but ICA remains substantially closer to SAE than to the generic linear baselines across domains.

This consistency is notable because ICA is fitted once per model layer and is not adapted to any of these downstream datasets.
The recovered non-Gaussian coordinates therefore preserve useful activation structure across a wide range of held-out text distributions.

\subsubsection{Across Layers}
\label{app:reconstruction-layers}

\begin{figure}[ht]
  \centering
  \includegraphics[width=\textwidth]{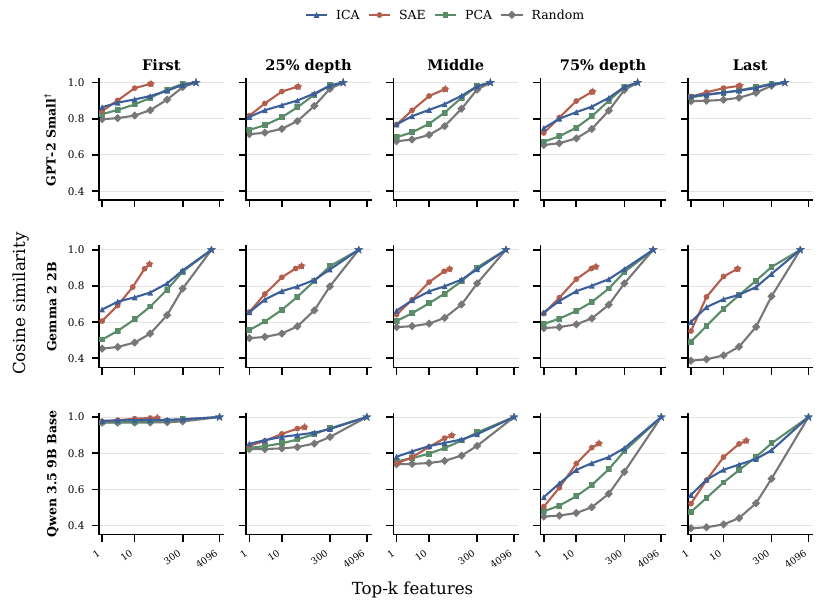}
  \caption{\textbf{Sparse reconstruction across model depth.} Cosine similarity is averaged across evaluation datasets at five normalized depth positions. ICA retains a strong sparse-reconstruction advantage over generic linear baselines across depth, while its gap to SAE varies by layer and model.}
  \label{fig:reconstruction-layers}
\end{figure}

Figure~\ref{fig:reconstruction-layers} disaggregates reconstruction across model depth.
For the five panels we use residual-stream layers $\{0,3,5,8,11\}$ for GPT-2 Small, $\{0,6,12,19,25\}$ for Gemma~2~2B, and $\{0,8,16,23,31\}$ for Qwen~3.5~9B Base, corresponding to the layers nearest 0\%, 25\%, 50\%, 75\%, and 100\% of each stack.

ICA remains a strong sparse reconstruction basis throughout depth.
Its advantage over random directions is large across nearly all layers, and it also generally improves over PCA at small budgets.

\FloatBarrier

\subsubsection{Alternative Metric}
\label{app:reconstruction-alternative-metric}

\begin{figure}[ht!]
  \centering
  \includegraphics[width=\textwidth]{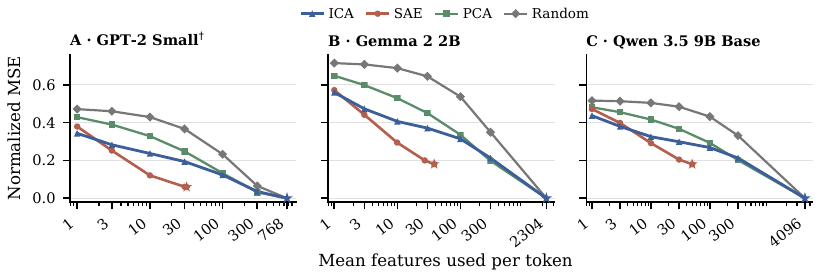}
  \caption{\textbf{Sparse reconstruction under an alternative metric.} Held-out top-$k$ reconstruction measured by NMSE, averaged equally across evaluated layers and datasets. Lower is better. The comparison gives the same broad ordering as cosine similarity.}
  \label{fig:reconstruction-nmse}
\end{figure}

Figure~\ref{fig:reconstruction-nmse} repeats the aggregate comparison using normalized mean squared error (NMSE), where lower values are better.
The qualitative ordering agrees with cosine similarity.
At sparse budgets, ICA generally has lower reconstruction error than PCA and random bases across all three models, while SAE is often strongest at intermediate sparse budgets.
The agreement across metrics shows that the reconstruction result is not specific to cosine similarity.

\FloatBarrier

\subsection{Sparse Probing}
\label{app:sparse-probing}

Sparse probing asks whether concept-relevant information can be read from a small number of representation coordinates.
We use the SAEBench sparse-probing evaluator with feature budgets
\begin{equation}
  k\in\{1,2,5,10,20,50,100\},
\end{equation}
a 4,000-example training split, a 1,000-example test split, and context length 128.
The reported sparse probes are L2-regularized logistic-regression classifiers.

Feature selection follows the SAEBench protocol.
Let $\ell\in\{+,-\}$ denote the binary class label. For each coordinate $j$, we compute the absolute difference in its mean activation between the positive and negative classes,
\begin{equation}
  \mathrm{score}_j
  =
  \left|
  \mathbb{E}[a_j\mid \ell=+]
  -
  \mathbb{E}[a_j\mid \ell=-]
  \right|,
\end{equation}
and retain the top-$k$ coordinates before fitting the probe.
The same selector is used for ICA, SAE, PCA, and random features.

For sparse probing, ICA exposes signed, skewness-oriented coordinates.
If $y_j$ is the raw ICA score, $s_j\in\{-1,+1\}$ is the orientation selected from the fitting distribution, and $d_j$ is the corresponding writing direction, the feature passed to SAEBench is
\begin{equation}
  a_j = (s_j y_j)\,\lVert d_j\rVert_2.
\end{equation}

\begin{figure}[ht!]
  \centering
  \includegraphics[width=0.72\textwidth]{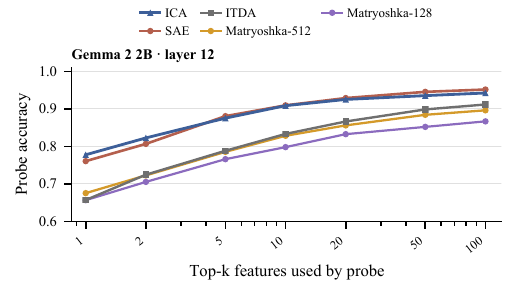}
  \caption{\textbf{Sparse probing with additional baselines on Gemma~2~2B layer~12.}
    We compare ICA with the public SAE, compact Matryoshka SAE variants, and ITDA.
  ICA and the full SAE achieve the strongest sparse-readout accuracy across feature budgets, with ICA slightly stronger at the smallest budgets and the SAE slightly stronger as the budget increases.}
  \label{fig:sparse-probing-extended-baselines}
\end{figure}

\textbf{Additional baselines.}
We further broaden the comparison on Gemma~2~2B layer~12.
\textbf{Matryoshka SAEs} test whether smaller effective SAE dictionaries improve sparse readout under compact feature budgets~\citep{bussmann2025learning}, while \textbf{ITDA} provides another training-light, non-SAE activation-dictionary baseline~\citep{leask2025inference}.
Figure~\ref{fig:sparse-probing-extended-baselines} shows that ICA and the full SAE form the strongest pair across the tested budgets: ICA has a small advantage at the smallest budgets, while the full SAE becomes slightly stronger as $k$ increases.
Both remain clearly above the compact Matryoshka variants and ITDA in this setting.
This comparison broadens the sparse-probing result beyond a single SAE baseline and shows that non-Gaussian coordinates provide strong concept readout relative to both compact learned dictionaries and another training-light alternative.

\begin{figure}[ht!]
  \centering
  \includegraphics[width=\textwidth]{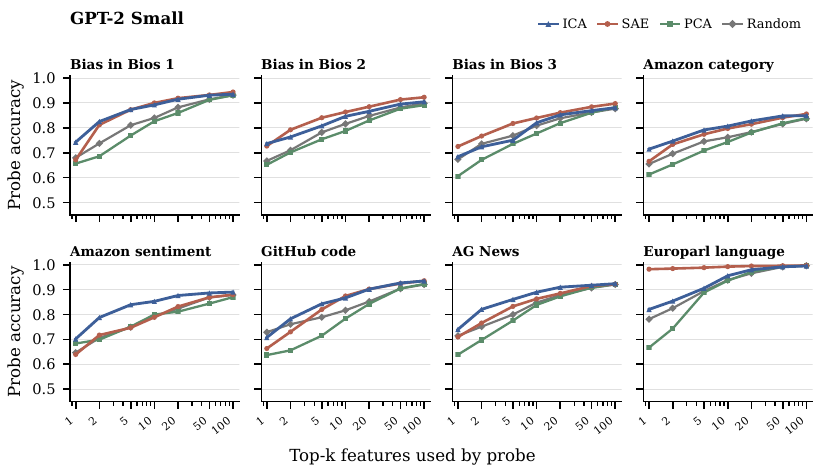}
  \caption{\textbf{Dataset-level sparse probing for GPT-2 Small.} Each panel reports one SAEBench dataset and averages equally over layers 6 and 10.}
  \label{fig:sparse-probing-datasets-gpt2}
\end{figure}

\begin{figure}[ht!]
  \centering
  \includegraphics[width=\textwidth]{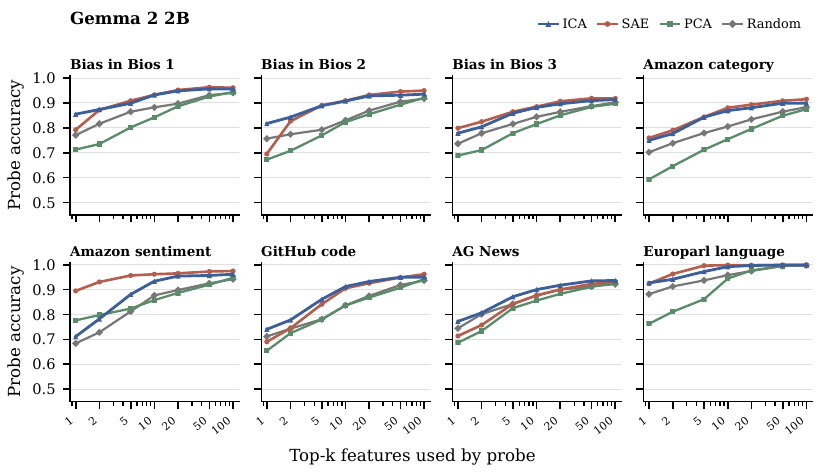}
  \caption{\textbf{Dataset-level sparse probing for Gemma~2~2B.} Each panel averages equally over layers 12 and 20. ICA and SAE remain close across most tasks, with the relative advantage depending on the dataset and feature budget.}
  \label{fig:sparse-probing-datasets-gemma2}
\end{figure}

\begin{figure}[ht!]
  \centering
  \includegraphics[width=\textwidth]{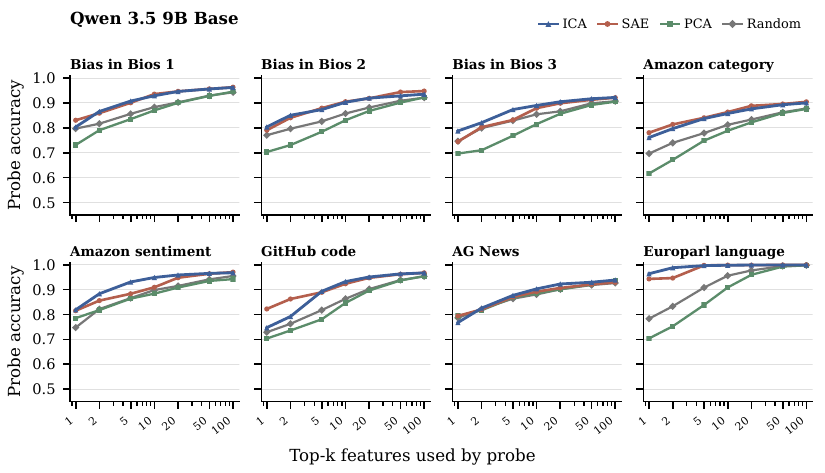}
  \caption{\textbf{Dataset-level sparse probing for Qwen~3.5~9B Base.} Each panel averages equally over layers 12 and 20. ICA and SAE are broadly competitive across datasets, particularly once more than a few coordinates are available.}
  \label{fig:sparse-probing-datasets-qwen}
\end{figure}

\textbf{Dataset-level results.}
Figures~\ref{fig:sparse-probing-datasets-gpt2}--\ref{fig:sparse-probing-datasets-qwen} disaggregate the main-text result by dataset.
ICA and SAE are close on many tasks, with the relative advantage depending on the concept and feature budget.
For GPT-2 Small, ICA is particularly strong at small budgets on Amazon sentiment, GitHub code, AG News, Amazon category, and several Bias in Bios tasks, while SAE has a clear advantage on Europarl language identification.
Gemma~2~2B and Qwen~3.5~9B Base show a more mixed pattern, with small advantages alternating across tasks.

Across models, a small number of ICA coordinates often support strong concept prediction.
As $k$ increases, the methods generally approach similar probe accuracy, suggesting that much of the concept-relevant information is concentrated in a small subset of non-Gaussian directions.

\FloatBarrier

\subsection{Targeted Probe Perturbation}
\label{app:tpp}

Sparse probing measures whether a small number of coordinates can predict a concept.
Targeted probe perturbation asks whether those coordinates can also be removed selectively.
For a target class $c$ and an ablation budget $k$, SAEBench defines the intended effect as the drop in the target probe's accuracy,
\begin{equation}
  \Delta_{\mathrm{intended}}(c,k)
  =
  \mathrm{Acc}_{\mathrm{clean}}(c)
  -
  \mathrm{Acc}_{\mathrm{ablated}\rightarrow c}(c),
\end{equation}
and the unintended effect as the mean accuracy drop over all non-target probes,
\begin{equation}
  \Delta_{\mathrm{unintended}}(c,k)
  =
  \frac{1}{|C\setminus\{c\}|}
  \sum_{c'\neq c}
  \left[
    \mathrm{Acc}_{\mathrm{clean}}(c')
    -
    \mathrm{Acc}_{\mathrm{ablated}\rightarrow c}(c')
  \right].
\end{equation}
Here, $C$ is the set of evaluated classes, and $\mathrm{Acc}_{\mathrm{ablated}\rightarrow c}(c')$ is probe $c'$'s accuracy after ablating the coordinates selected for target class $c$.
The TPP score is
\begin{equation}
  \mathrm{TPP}(c,k)
  =
  \Delta_{\mathrm{intended}}(c,k)
  -
  \Delta_{\mathrm{unintended}}(c,k).
\end{equation}
A high score therefore requires a large effect on the intended concept together with small collateral effects on unrelated concepts.

SAEBench ranks candidate features separately for each target class.
For feature $j$, the ranking combines its class-conditional activation difference with the corresponding probe readout,
\begin{equation}
  \mathrm{effect}_j
  =
  \max\!\left(
    \bar a_j^{+}-\bar a_j^{-},
    0
  \right)
  \left(w_{\mathrm{probe}}^\top d_j\right),
\end{equation}
and the top-$k$ nonzero effects are selected for ablation.
Here, $\bar a_j^{+}$ and $\bar a_j^{-}$ are the mean feature activations for the positive and negative examples of target class $c$, $w_{\mathrm{probe}}$ is the target probe's residual-space readout, and $d_j$ is the feature's decoder direction.

For TPP, each oriented ICA component is exposed as a one-sided non-negative feature,
\begin{equation}
  a_j
  =
  \max(s_j y_j,0)\,\lVert d_j\rVert_2,
\end{equation}
with decoder direction
\begin{equation}
  \widetilde{d}_j
  =
  \frac{s_j d_j}{\lVert d_j\rVert_2}.
\end{equation}
This representation uses the positive half-wave of the skewness-oriented ICA component as the feature passed to SAEBench.

ICA and SAE use the same reconstruction-error-preserving intervention:
\begin{equation}
  \tilde x
  =
  \mathrm{decode}
  \bigl(
    \mathrm{ablate}(\mathrm{encode}(x))
  \bigr)
  +
  x
  -
  \mathrm{decode}(\mathrm{encode}(x)).
\end{equation}
Ablation sets the selected features to zero while adding back the original representation's reconstruction error.
For ICA, this removes the selected oriented positive half-wave; for SAE, it zeroes the selected checkpoint latents.

\begin{figure}[ht!]
  \centering
  \includegraphics[width=\textwidth]{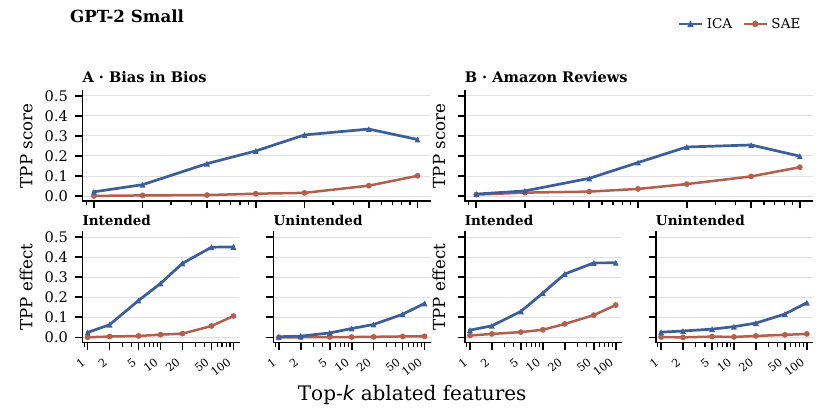}
  \caption{\textbf{Dataset-level targeted probe perturbation for GPT-2 Small.} Results are averaged equally over layers 6 and 10. ICA produces a substantially larger intended effect while unintended effects remain comparatively small, yielding higher TPP scores on both datasets.}
  \label{fig:tpp-datasets-gpt2}
\end{figure}

\begin{figure}[ht!]
  \centering
  \includegraphics[width=\textwidth]{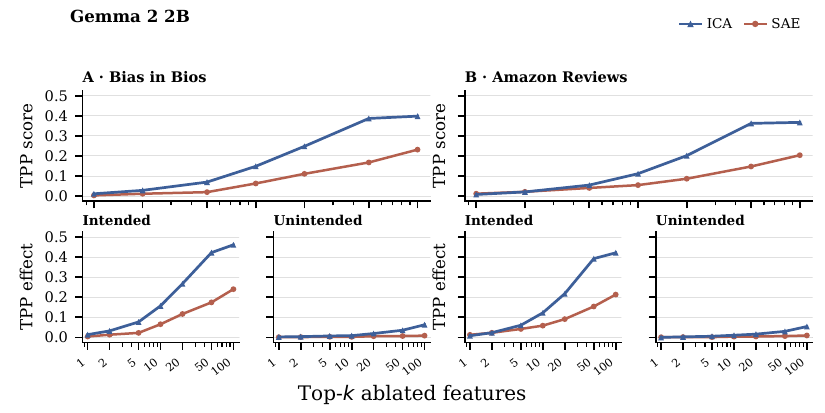}
  \caption{\textbf{Dataset-level targeted probe perturbation for Gemma~2~2B.} Results are averaged equally over layers 12 and 20. ICA again produces a larger target-probe effect than SAE, yielding a consistent advantage in total TPP score.}
  \label{fig:tpp-datasets-gemma2}
\end{figure}

\begin{figure}[ht!]
  \centering
  \includegraphics[width=\textwidth]{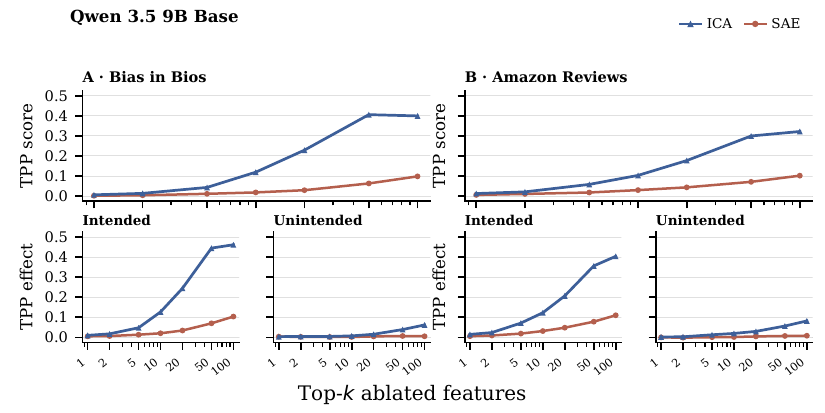}
  \caption{\textbf{Dataset-level targeted probe perturbation for Qwen~3.5~9B Base.} Results are averaged equally over layers 12 and 20. The same selectivity pattern persists at larger model scale: ICA produces substantially larger intended effects while unintended effects remain comparatively small.}
  \label{fig:tpp-datasets-qwen}
\end{figure}

Figures~\ref{fig:tpp-datasets-gpt2}--\ref{fig:tpp-datasets-qwen} separate the aggregate TPP score into its intended and unintended effects for both evaluation datasets.
The same pattern appears across GPT-2 Small, Gemma~2~2B, and Qwen~3.5~9B Base.
ICA obtains higher TPP scores than SAE across the useful sparse-ablation range on both Bias in Bios and Amazon Reviews.

The decomposition shows why.
Ablating ICA features produces a substantially larger drop in the target probe than ablating the corresponding number of SAE features.
Unintended effects also increase with larger ablation budgets, but remain much smaller than the intended effect.
The resulting separation between target and non-target behavior is therefore consistently larger for ICA across models and datasets.

These results complement sparse probing.
The same non-Gaussian coordinates that concentrate enough concept information to support accurate sparse probes can also be selected and removed in a way that disproportionately affects the intended concept.

\newpage

\newpage
\subsection{RAVEL Evaluation Protocol}
\label{app:ravel}

We follow the SAEBench implementation of RAVEL~\citep{huang2024ravelevaluatinginterpretabilitymethods,karvonen2025saebench} to measure selective changes in generated answers. We evaluate Gemma~2~2B at the layer-12 residual stream on six attributes: country, continent, and language for cities, and country of birth, field, and gender for Nobel laureates. Following the default protocol, we filter prompts that the model answers correctly and retain up to 500 entities and 90 prompt templates. For each target attribute, we sample up to 5,000 Cause pairs and an equal number of Isolation pairs. Cause pairs require the edited answer to match the source entity's target-attribute value, while Isolation pairs require answers about other attributes to retain their base-entity values.

For each attribute, we train a binary mask over representation coordinates using both pair types, with a 70/30 train/test split. At the final entity token, we replace masked coordinates of the base representation with those of the source, decode the edited representation, and generate up to six tokens with greedy decoding. We compare ICA, PCA, and the Gemma Scope 16k SAE (\texttt{average\_l0\_41}); ICA and PCA use sign-split coordinates. Masks are trained for two epochs with a learning rate of $10^{-3}$. We evaluate two seeds (0 and 42) using the same filtered dataset. Cause and Isolation are the respective success rates on held-out pairs, and Disentangle is their average. In Figure~\ref{fig:ravel-overall}, markers show means across the six attributes and two seeds, while whiskers indicate $\pm1$ standard deviation across attributes after averaging seeds.


\newpage
\section{SAE Baselines and Checkpoints}
\label{app:sae-baselines}

We compare ICA against public SAE releases matched to the same models and post-block residual-stream locations used in our experiments.
Whenever an evaluation samples individual SAE features, we exclude dead features before sampling and draw only from features that are active under that evaluation protocol.

\begin{table}[h]
  \centering
  \caption{\textbf{Public SAE baselines used throughout our comparisons.}}
  \label{tab:sae-baselines}
  \small
  \begin{tabular}{@{}llll@{}}
    \toprule
    Model & SAE release & Width & Native activation \\
    \midrule
    GPT-2 Small & OAI v5 32k & 32,768 & TopK-32 \\
    Gemma~2~2B & Gemma Scope residual 16k & 16,384 & JumpReLU \\
    Qwen~3.5~9B Base & Qwen Scope residual 64k & 65,536 & TopK-50 \\
    \bottomrule
  \end{tabular}
\end{table}

The exact checkpoints are:
\begin{itemize}
  \item \textbf{GPT-2 Small:} \url{jbloom/GPT2-Small-OAI-v5-32k-resid-post-SAEs}, \newline revision \url{21df74db7b9f690cd2403a20485cf75f46330d0d}.
  \item \textbf{Gemma~2~2B:} \url{google/gemma-scope-2b-pt-res}, \newline revision \url{fd571b47c1c64851e9b1989792367b9babb4af63}. 
  \item \textbf{Qwen~3.5~9B Base:} \url{Qwen/SAE-Res-Qwen3.5-9B-Base-W64K-L0_50}, \newline revision \url{7bb2370d360e566b2d8acb8b3d15a0b2b88f52a8}.
\end{itemize}

\end{document}